\documentclass{article}
\usepackage{iclr2027_conference,times}

\usepackage{amsmath,amsfonts,bm}

\def\eqref#1{equation~\ref{#1}}

\def\1{\bm{1}}

\DeclareMathAlphabet{\mathsfit}{\encodingdefault}{\sfdefault}{m}{sl}
\SetMathAlphabet{\mathsfit}{bold}{\encodingdefault}{\sfdefault}{bx}{n}

\usepackage{hyperref}
\usepackage{url}
\usepackage{graphicx}
\usepackage{amsmath,amssymb}
\usepackage{algorithm}
\usepackage{algpseudocode}
\algrenewcommand{\algorithmiccomment}[1]{\hfill{\scriptsize\color{black!55}\(\triangleright\) #1}}
\usepackage{tikz}
\usetikzlibrary{arrows.meta,positioning,fit,backgrounds,calc}
\usepackage{caption}
\usepackage{wrapfig}
\newcommand{\TOneLowShift}{3em}
\usepackage{booktabs}
\newcommand{\rot}[1]{\rotatebox{90}{#1}}
\usepackage{multirow}
\usepackage[table]{xcolor}
\usepackage{etoolbox}
\usepackage{enumitem}

\definecolor{tabhead}{HTML}{12508F}
\definecolor{tabband}{HTML}{12508F}

\newcommand{\headrow}{\rowcolor{tabhead!10}}

\newcommand{\sievtablestyle}{\renewcommand{\arraystretch}{1.06}\setlength{\tabcolsep}{4pt}\rowcolors{2}{}{tabband!4}}
\AtBeginEnvironment{table}{\sievtablestyle}
\AtBeginEnvironment{table*}{\sievtablestyle}
\AtBeginEnvironment{wraptable}{\sievtablestyle}

\newcommand{\dcterm}[2]{\emph{#1}\,{\tiny$(#2)$}}

\definecolor{algcomment}{HTML}{6B7785}
\definecolor{algnumber}{HTML}{9AA3AE}
\renewcommand{\algorithmiccomment}[1]{\hfill\textcolor{algcomment}{\scriptsize$\triangleright$\,#1}}

\usepackage[T1]{fontenc}
\usepackage{upquote}
\usepackage{listings}
\usepackage{tcolorbox}
\tcbuselibrary{skins,breakable,listings}

\definecolor{promptrule}{HTML}{12508F}
\definecolor{promptsub}{HTML}{8C564B}
\definecolor{promptbrk}{HTML}{A03000}

\lstdefinestyle{promptverbatim}{basicstyle=\ttfamily\footnotesize,
  columns=fixed,
  keepspaces=true,
  showspaces=false,
  showstringspaces=false,
  extendedchars=true,
  breaklines=true,
  breakatwhitespace=true,
  breakindent=2em,
  postbreak={\mbox{\textcolor{promptbrk}{$\hookrightarrow$}\space}},
  aboveskip=2pt, belowskip=2pt,
}

\newtcolorbox{promptbox}[2][]{breakable, enhanced, sharp corners=downhill,
  colback=promptrule!3!white, colframe=promptrule,
  boxrule=0.5pt, left=4pt, right=4pt, top=3pt, bottom=3pt,
  fonttitle=\sffamily\bfseries\footnotesize,
  coltitle=white, title={#2}, #1}

\newtcolorbox{promptboxsub}[2][]{breakable, enhanced, sharp corners=downhill,
  colback=promptsub!4!white, colframe=promptsub,
  boxrule=0.5pt, leftrule=3pt, left=4pt, right=4pt, top=3pt, bottom=3pt,
  fonttitle=\sffamily\bfseries\footnotesize,
  coltitle=white, title={#2}, #1}

\definecolor{linknavy}{RGB}{0,51,153}
\hypersetup{colorlinks=true, linkcolor=linknavy, citecolor=linknavy,
            urlcolor=linknavy, filecolor=linknavy}

\newcommand{\ProbeN}{327}
\newcommand{\NIMargin}{0.05}
\newcommand{\NIChallenger}{gpt-5-mini}
\newcommand{\NIHOneLo}{-0.015}
\newcommand{\NIHOneHi}{+0.061}
\newcommand{\NIBaseLo}{-0.288}
\newcommand{\NIBaseHi}{-0.135}
\newcommand{\CEightbDefenceN}{221}
\newcommand{\BOneCSDiff}{0.1223}
\newcommand{\TTCNEpisodes}{2616}
\newcommand{\TTCNClusters}{327}
\newcommand{\TTCCallsMaj}{64}

\newcommand{\CalibInModel}{41.8}
\newcommand{\CalibInTrue}{39.5}
\newcommand{\CalibOodModel}{62.00}
\newcommand{\CalibOodTrue}{60.40}
\newcommand{\CalibTZeroInPct}{24}
\newcommand{\CalibTZeroOodPct}{17}
\newcommand{\AccRLSteps}{290}
\newcommand{\AccRLPoints}{30}
\newcommand{\RewardDual}{0.5}
\newcommand{\RewardRF}{2.0}
\newcommand{\PoolTrain}{2{,}629}
\newcommand{\TrainItems}{476}
\newcommand{\PZeroSamples}{32}
\newcommand{\OptLR}{1.0\!\times\!10^{-6}}
\newcommand{\OptClipLo}{0.2}
\newcommand{\OptClipHi}{0.28}
\newcommand{\OptEntropy}{0.001}
\newcommand{\OptKL}{0.001}
\newcommand{\OptTemp}{1.0}
\newcommand{\OptGroup}{8}
\newcommand{\OptBatch}{128}
\newcommand{\OptCtx}{12{,}288}
\newcommand{\OptResp}{4{,}096}
\newcommand{\SplitEpisodes}{290{,}059}
\newcommand{\SedCS}{0.7217}
\newcommand{\CPanelTZeroCT}{0.6911}\newcommand{\CPanelTZeroCS}{0.6055}
\newcommand{\CPanelAOneCT}{0.6820}\newcommand{\CPanelAOneCS}{0.5994}
\newcommand{\CPanelAZeroCT}{0.7034}\newcommand{\CPanelAZeroCS}{0.6820}
\newcommand{\CPanelCOneCtlCT}{0.7003}\newcommand{\CPanelCOneCtlCS}{0.6850}
\newcommand{\CPanelCOnePCT}{0.7095}\newcommand{\CPanelCOnePCS}{0.7003}
\newcommand{\CPanelHOneCT}{0.7095}\newcommand{\CPanelHOneCS}{0.7217}
\newcommand{\CPanelHThreeCT}{0.7064}\newcommand{\CPanelHThreeCS}{0.7187}
\newcommand{\CPanelStep}{150}\newcommand{\CPanelItems}{327}
\newcommand{\COneCTLo}{0.6820}
\newcommand{\COneCTHi}{0.7095}
\newcommand{\COneCTSpread}{0.0275}
\newcommand{\UnrewardedCS}{0.6055}
\newcommand{\SevenCTSpreadPP}{2.75}\newcommand{\SevenCSSpreadPP}{12.23}
\newcommand{\SevenRatio}{4.4}
\newcommand{\pct}{\,\%}
\newcommand{\pp}{\,pp}
\newcommand{\EOneHOnePct}{1.0}\newcommand{\EOneHOneSDPct}{0.4}
\newcommand{\EOneJZeroPct}{2.8}\newcommand{\EOneJZeroSDPct}{0.6}
\newcommand{\EOneSZeroPct}{12.8}\newcommand{\EOneSZeroSDPct}{1.5}
\newcommand{\EOneJZeroGapPct}{1.8}
\newcommand{\EOneSZeroGapPct}{11.8}
\newcommand{\BaseKdrawPct}{10.3}\newcommand{\BaseKdrawSDPct}{2.3}
\newcommand{\TZeroKdrawPct}{15.1}\newcommand{\TZeroKdrawSDPct}{1.4}
\newcommand{\MZeroKdrawPct}{13.5}\newcommand{\MZeroKdrawSDPct}{1.3}
\newcommand{\AccRLFlipWrongBasePct}{31.8}\newcommand{\AccRLFlipWrongEndPct}{38.9}
\newcommand{\AccRLContrastPct}{4.4}\newcommand{\AccRLCILoPct}{1.5}\newcommand{\AccRLCIHiPct}{7.5}
\newcommand{\AccRLSlopePct}{1.3}
\newcommand{\AccRLCTLoPct}{67.6}\newcommand{\AccRLCTHiPct}{71.4}
\newcommand{\SplitCTonePct}{44}\newcommand{\SplitCTzeroPct}{37}
\newcommand{\ChalTemp}{0.7}
\newcommand{\ProbeItems}{327}
\newcommand{\XFamBandPct}{5}
\newcommand{\XFamModels}{four}
\newcommand{\XFamModelsAboveBand}{three}
\newcommand{\XfGemmaThreeCavB}{20.2}\newcommand{\XfGemmaThreeCavBsd}{2.6}
\newcommand{\XfGemmaThreeCavA}{2.6}\newcommand{\XfGemmaThreeCavAsd}{0.8}
\newcommand{\XfGemmaThreeAccB}{53.0}\newcommand{\XfGemmaThreeAccBsd}{1.9}
\newcommand{\XfGemmaThreeAccA}{68.1}\newcommand{\XfGemmaThreeAccAsd}{0.8}
\newcommand{\XfGraniteCavB}{14.7}\newcommand{\XfGraniteCavBsd}{1.3}
\newcommand{\XfGraniteCavA}{0.9}\newcommand{\XfGraniteCavAsd}{0.8}
\newcommand{\XfGraniteAccB}{40.5}\newcommand{\XfGraniteAccBsd}{1.6}
\newcommand{\XfGraniteAccA}{46.1}\newcommand{\XfGraniteAccAsd}{1.6}
\newcommand{\XfLlamaCavB}{6.6}\newcommand{\XfLlamaCavBsd}{0.6}
\newcommand{\XfLlamaCavA}{2.6}\newcommand{\XfLlamaCavAsd}{0.3}
\newcommand{\XfLlamaAccB}{51.4}\newcommand{\XfLlamaAccBsd}{1.7}
\newcommand{\XfLlamaAccA}{57.9}\newcommand{\XfLlamaAccAsd}{0.7}
\newcommand{\XfQwenFourteenCavB}{1.8}\newcommand{\XfQwenFourteenCavBsd}{0.4}
\newcommand{\XfQwenFourteenCavA}{0.8}\newcommand{\XfQwenFourteenCavAsd}{0.4}
\newcommand{\XfQwenFourteenAccB}{71.1}\newcommand{\XfQwenFourteenAccBsd}{0.7}
\newcommand{\XfQwenFourteenAccA}{73.4}\newcommand{\XfQwenFourteenAccAsd}{0.6}
\newcommand{\TZeroAccPct}{61.9}
\newcommand{\MZeroAccPct}{62.5}\newcommand{\BaseAccPct}{62.8}
\newcommand{\TZeroAccSDPct}{2.0}
\newcommand{\MZeroAccSDPct}{1.2}
\newcommand{\BaseAccSDPct}{2.9}
\newcommand{\EOneJZeroAccPct}{70.2}\newcommand{\EOneJZeroAccSDPct}{0.4}
\newcommand{\EOneSZeroAccPct}{62.3}\newcommand{\EOneSZeroAccSDPct}{0.8}
\newcommand{\EOneHOneAccPct}{74.5}\newcommand{\EOneHOneAccSDPct}{1.2}
\newcommand{\ResidHOneNoiseRatio}{0.32}\newcommand{\ResidBaseNoiseRatio}{1.47}
\newcommand{\JuryTailBasePct}{20.2}\newcommand{\JuryTailTrainedPct}{12.6}
\newcommand{\JuryTailN}{350}
\newcommand{\JuryCaveN}{121}
\newcommand{\JuryBaseDefendN}{381}\newcommand{\JuryBaseCorrectN}{89}
\newcommand{\JuryTrainedHeldN}{590}
\newcommand{\JuryBaseCorrectTailPct}{49.4}
\newcommand{\JuryDrift}{0.024}
\newcommand{\JuryAlpha}{0.787}
\newcommand{\JuryAlphaSubstantive}{0.904}
\newcommand{\JurySubstantivePct}{67}
\newcommand{\JurySelfPrefDiff}{-0.007}
\newcommand{\JurySelfPrefLo}{-0.077}\newcommand{\JurySelfPrefHi}{+0.068}

\newcommand{\LxNeverWord}{fifteen}
\newcommand{\LxFixedWord}{sixteen}
\newcommand{\DcExtWord}{15}
\newcommand{\PnBaseAgieval}{96.3}\newcommand{\PnAccAgieval}{97.6}\newcommand{\PnPlcAgieval}{98.1}\newcommand{\PnExtAgieval}{98.8}\newcommand{\PnSedAgieval}{99.1}
\newcommand{\PnBaseAime}{17.5}\newcommand{\PnAccAime}{15.4}\newcommand{\PnPlcAime}{14.2}\newcommand{\PnExtAime}{22.1}\newcommand{\PnSedAime}{25.4}
\newcommand{\PnBaseGpqa}{38.0}\newcommand{\PnAccGpqa}{37.3}\newcommand{\PnPlcGpqa}{41.9}\newcommand{\PnExtGpqa}{44.2}\newcommand{\PnSedGpqa}{45.6}
\newcommand{\PnBaseLcb}{65.4}\newcommand{\PnAccLcb}{64.6}\newcommand{\PnPlcLcb}{66.6}\newcommand{\PnExtLcb}{73.6}\newcommand{\PnSedLcb}{72.2}
\newcommand{\PnBaseMmluRedux}{74.7}\newcommand{\PnAccMmluRedux}{74.6}\newcommand{\PnPlcMmluRedux}{80.1}\newcommand{\PnExtMmluRedux}{81.9}\newcommand{\PnSedMmluRedux}{82.5}
\newcommand{\PnBaseMusr}{58.5}\newcommand{\PnAccMusr}{51.9}\newcommand{\PnPlcMusr}{60.4}\newcommand{\PnExtMusr}{60.1}\newcommand{\PnSedMusr}{63.0}
\newcommand{\PnBaseOmni}{32.4}\newcommand{\PnAccOmni}{33.7}\newcommand{\PnPlcOmni}{34.9}\newcommand{\PnExtOmni}{37.3}\newcommand{\PnSedOmni}{38.8}
\newcommand{\PnBaseSupergpqa}{33.9}\newcommand{\PnAccSupergpqa}{34.6}\newcommand{\PnPlcSupergpqa}{37.5}\newcommand{\PnExtSupergpqa}{38.8}\newcommand{\PnSedSupergpqa}{39.5}
\newcommand{\PnBaseTheorem}{33.5}\newcommand{\PnAccTheorem}{35.5}\newcommand{\PnPlcTheorem}{37.0}\newcommand{\PnExtTheorem}{39.8}\newcommand{\PnSedTheorem}{40.6}
\newcommand{\PnBaseMacro}{50.0}\newcommand{\PnBaseSd}{0.8}
\newcommand{\PnAccMacro}{49.5}\newcommand{\PnAccSd}{0.8}
\newcommand{\PnPlcMacro}{52.3}\newcommand{\PnPlcSd}{0.6}
\newcommand{\PnExtMacro}{55.2}\newcommand{\PnExtSd}{0.8}
\newcommand{\PnSedMacro}{56.3}\newcommand{\PnSedSd}{0.5}
\newcommand{\PnPlcD}{+2.29}\newcommand{\PnPlcDLo}{+0.67}\newcommand{\PnPlcDHi}{+3.91}
\newcommand{\PnExtD}{+5.15}\newcommand{\PnExtDLo}{+3.75}\newcommand{\PnExtDHi}{+6.54}
\newcommand{\PnSedD}{+6.30}\newcommand{\PnSedDLo}{+5.19}\newcommand{\PnSedDHi}{+7.41}
\newcommand{\PnNbench}{nine}
\newcommand{\PnRevAgieval}{97.7}
\newcommand{\PnRevAime}{19.6}
\newcommand{\PnRevGpqa}{40.2}
\newcommand{\PnRevLcb}{66.3}
\newcommand{\PnRevMmluRedux}{75.6}
\newcommand{\PnRevMusr}{53.6}
\newcommand{\PnRevOmni}{34.1}
\newcommand{\PnRevSupergpqa}{35.2}
\newcommand{\PnRevTheorem}{35.7}
\newcommand{\PnRevMacro}{50.9}\newcommand{\PnRevSd}{1.1}
\newcommand{\PnMajMacro}{52.4}\newcommand{\PnMajSd}{1.2}
\newcommand{\PnReflMacro}{51.4}\newcommand{\PnReflSd}{1.1}
\newcommand{\PnTTCDraws}{three}
\newcommand{\PnMajAgieval}{97.6}\newcommand{\SdMajAgieval}{0.7}
\newcommand{\PnMajAime}{24.4}\newcommand{\SdMajAime}{9.6}
\newcommand{\PnMajGpqa}{43.8}\newcommand{\SdMajGpqa}{2.0}
\newcommand{\PnMajLcb}{65.9}\newcommand{\SdMajLcb}{1.1}
\newcommand{\PnMajMmluRedux}{77.3}\newcommand{\SdMajMmluRedux}{0.4}
\newcommand{\PnMajMusr}{58.5}\newcommand{\SdMajMusr}{1.7}
\newcommand{\PnMajOmni}{33.6}\newcommand{\SdMajOmni}{0.4}
\newcommand{\PnMajSupergpqa}{35.4}\newcommand{\SdMajSupergpqa}{0.1}
\newcommand{\PnMajTheorem}{35.0}\newcommand{\SdMajTheorem}{0.9}
\newcommand{\PnReflAgieval}{97.3}\newcommand{\SdReflAgieval}{0.8}
\newcommand{\PnReflAime}{23.3}\newcommand{\SdReflAime}{8.8}
\newcommand{\PnReflGpqa}{44.5}\newcommand{\SdReflGpqa}{1.8}
\newcommand{\PnReflLcb}{65.5}\newcommand{\SdReflLcb}{0.9}
\newcommand{\PnReflMmluRedux}{75.6}\newcommand{\SdReflMmluRedux}{1.0}
\newcommand{\PnReflMusr}{53.6}\newcommand{\SdReflMusr}{3.0}
\newcommand{\PnReflOmni}{33.5}\newcommand{\SdReflOmni}{0.9}
\newcommand{\PnReflSupergpqa}{34.5}\newcommand{\SdReflSupergpqa}{0.2}
\newcommand{\PnReflTheorem}{35.1}\newcommand{\SdReflTheorem}{1.1}
\newcommand{\DegPlaneGiveLo}{17\,\%}\newcommand{\DegPlaneGiveHi}{52\,\%}
\newcommand{\PnVerMacro}{52.5}\newcommand{\PnVerSd}{1.2}
\newcommand{\PnVerAgieval}{97.9}
\newcommand{\PnVerAime}{24.2}
\newcommand{\PnVerGpqa}{40.2}
\newcommand{\PnVerLcb}{70.3}
\newcommand{\PnVerMmluRedux}{77.1}
\newcommand{\PnVerMusr}{52.0}
\newcommand{\PnVerOmni}{36.6}
\newcommand{\PnVerSupergpqa}{36.7}
\newcommand{\PnVerTheorem}{37.8}
\newcommand{\SdBaseAgieval}{1.0}\newcommand{\SdAccAgieval}{0.7}\newcommand{\SdRevAgieval}{0.6}\newcommand{\SdVerAgieval}{0.6}\newcommand{\SdPlcAgieval}{0.7}\newcommand{\SdExtAgieval}{0.3}\newcommand{\SdSedAgieval}{0.3}
\newcommand{\SdBaseAime}{3.5}\newcommand{\SdAccAime}{3.1}\newcommand{\SdRevAime}{5.5}\newcommand{\SdVerAime}{5.3}\newcommand{\SdPlcAime}{5.0}\newcommand{\SdExtAime}{4.8}\newcommand{\SdSedAime}{4.3}
\newcommand{\SdBaseGpqa}{3.5}\newcommand{\SdAccGpqa}{3.8}\newcommand{\SdRevGpqa}{2.8}\newcommand{\SdVerGpqa}{3.0}\newcommand{\SdPlcGpqa}{2.1}\newcommand{\SdExtGpqa}{1.3}\newcommand{\SdSedGpqa}{1.5}
\newcommand{\SdBaseLcb}{2.4}\newcommand{\SdAccLcb}{2.6}\newcommand{\SdRevLcb}{2.1}\newcommand{\SdVerLcb}{1.5}\newcommand{\SdPlcLcb}{1.5}\newcommand{\SdExtLcb}{1.4}\newcommand{\SdSedLcb}{2.0}
\newcommand{\SdBaseMmluRedux}{1.7}\newcommand{\SdAccMmluRedux}{1.7}\newcommand{\SdRevMmluRedux}{1.4}\newcommand{\SdVerMmluRedux}{1.2}\newcommand{\SdPlcMmluRedux}{0.4}\newcommand{\SdExtMmluRedux}{0.2}\newcommand{\SdSedMmluRedux}{0.3}
\newcommand{\SdBaseMusr}{2.9}\newcommand{\SdAccMusr}{3.6}\newcommand{\SdRevMusr}{3.1}\newcommand{\SdVerMusr}{3.6}\newcommand{\SdPlcMusr}{1.4}\newcommand{\SdExtMusr}{2.8}\newcommand{\SdSedMusr}{2.0}
\newcommand{\SdBaseOmni}{0.6}\newcommand{\SdAccOmni}{0.4}\newcommand{\SdRevOmni}{0.4}\newcommand{\SdVerOmni}{0.4}\newcommand{\SdPlcOmni}{0.4}\newcommand{\SdExtOmni}{0.5}\newcommand{\SdSedOmni}{0.3}
\newcommand{\SdBaseSupergpqa}{0.6}\newcommand{\SdAccSupergpqa}{0.7}\newcommand{\SdRevSupergpqa}{0.6}\newcommand{\SdVerSupergpqa}{0.7}\newcommand{\SdPlcSupergpqa}{0.3}\newcommand{\SdExtSupergpqa}{0.2}\newcommand{\SdSedSupergpqa}{0.1}
\newcommand{\SdBaseTheorem}{0.8}\newcommand{\SdAccTheorem}{0.9}\newcommand{\SdRevTheorem}{1.9}\newcommand{\SdVerTheorem}{1.3}\newcommand{\SdPlcTheorem}{0.8}\newcommand{\SdExtTheorem}{0.9}\newcommand{\SdSedTheorem}{0.8}
\newcommand{\GnBaseAgieval}{56.7}\newcommand{\GsBaseAgieval}{3.9}
\newcommand{\GnSedAgieval}{87.4}\newcommand{\GsSedAgieval}{2.1}
\newcommand{\GnVerAgieval}{71.4}\newcommand{\GsVerAgieval}{1.3}
\newcommand{\GnRevAgieval}{73.6}\newcommand{\GsRevAgieval}{2.8}
\newcommand{\GnAccAgieval}{61.6}\newcommand{\GsAccAgieval}{2.1}
\newcommand{\GnBaseAime}{0.0}\newcommand{\GsBaseAime}{0.0}
\newcommand{\GnSedAime}{0.0}\newcommand{\GsSedAime}{0.0}
\newcommand{\GnVerAime}{0.0}\newcommand{\GsVerAime}{0.0}
\newcommand{\GnRevAime}{0.0}\newcommand{\GsRevAime}{0.0}
\newcommand{\GnAccAime}{0.0}\newcommand{\GsAccAime}{0.0}
\newcommand{\GnBaseGpqa}{17.1}\newcommand{\GsBaseGpqa}{1.9}
\newcommand{\GnSedGpqa}{32.0}\newcommand{\GsSedGpqa}{2.3}
\newcommand{\GnVerGpqa}{30.6}\newcommand{\GsVerGpqa}{2.3}
\newcommand{\GnRevGpqa}{29.3}\newcommand{\GsRevGpqa}{2.4}
\newcommand{\GnAccGpqa}{27.6}\newcommand{\GsAccGpqa}{2.4}
\newcommand{\GnBaseLcb}{8.6}\newcommand{\GsBaseLcb}{1.1}
\newcommand{\GnSedLcb}{29.6}\newcommand{\GsSedLcb}{1.0}
\newcommand{\GnVerLcb}{14.6}\newcommand{\GsVerLcb}{1.3}
\newcommand{\GnRevLcb}{17.3}\newcommand{\GsRevLcb}{1.6}
\newcommand{\GnAccLcb}{11.3}\newcommand{\GsAccLcb}{1.4}
\newcommand{\GnBaseMmluRedux}{37.7}\newcommand{\GsBaseMmluRedux}{2.4}
\newcommand{\GnSedMmluRedux}{74.4}\newcommand{\GsSedMmluRedux}{0.2}
\newcommand{\GnVerMmluRedux}{57.0}\newcommand{\GsVerMmluRedux}{1.8}
\newcommand{\GnRevMmluRedux}{53.3}\newcommand{\GsRevMmluRedux}{1.1}
\newcommand{\GnAccMmluRedux}{41.2}\newcommand{\GsAccMmluRedux}{3.1}
\newcommand{\GnBaseMusr}{34.2}\newcommand{\GsBaseMusr}{2.0}
\newcommand{\GnSedMusr}{61.5}\newcommand{\GsSedMusr}{2.2}
\newcommand{\GnVerMusr}{41.6}\newcommand{\GsVerMusr}{1.1}
\newcommand{\GnRevMusr}{40.5}\newcommand{\GsRevMusr}{3.5}
\newcommand{\GnAccMusr}{40.2}\newcommand{\GsAccMusr}{3.0}
\newcommand{\GnBaseOmni}{6.8}\newcommand{\GsBaseOmni}{0.4}
\newcommand{\GnSedOmni}{14.3}\newcommand{\GsSedOmni}{0.5}
\newcommand{\GnVerOmni}{9.2}\newcommand{\GsVerOmni}{0.3}
\newcommand{\GnRevOmni}{8.9}\newcommand{\GsRevOmni}{0.3}
\newcommand{\GnAccOmni}{7.7}\newcommand{\GsAccOmni}{0.6}
\newcommand{\GnBaseSupergpqa}{11.2}\newcommand{\GsBaseSupergpqa}{0.4}
\newcommand{\GnSedSupergpqa}{24.5}\newcommand{\GsSedSupergpqa}{0.2}
\newcommand{\GnVerSupergpqa}{20.6}\newcommand{\GsVerSupergpqa}{0.3}
\newcommand{\GnRevSupergpqa}{19.7}\newcommand{\GsRevSupergpqa}{0.2}
\newcommand{\GnAccSupergpqa}{15.7}\newcommand{\GsAccSupergpqa}{0.5}
\newcommand{\GnBaseTheorem}{6.7}\newcommand{\GsBaseTheorem}{0.6}
\newcommand{\GnSedTheorem}{16.9}\newcommand{\GsSedTheorem}{0.9}
\newcommand{\GnVerTheorem}{9.0}\newcommand{\GsVerTheorem}{0.7}
\newcommand{\GnRevTheorem}{8.9}\newcommand{\GsRevTheorem}{0.9}
\newcommand{\GnAccTheorem}{6.8}\newcommand{\GsAccTheorem}{1.0}
\newcommand{\GnBaseMacro}{19.9}\newcommand{\GnBaseSd}{0.7}
\newcommand{\GnSedMacro}{37.8}\newcommand{\GnSedSd}{0.7}
\newcommand{\GnVerMacro}{28.2}\newcommand{\GnVerSd}{0.5}
\newcommand{\GnRevMacro}{28.0}\newcommand{\GnRevSd}{0.9}
\newcommand{\GnAccMacro}{23.6}\newcommand{\GnAccSd}{0.9}
\newcommand{\PnDecodes}{eight}\newcommand{\GnDecodes}{eight}
\newcommand{\GemErrLo}{0.00\,\%}\newcommand{\GemErrHi}{0.20\,\%}
\newcommand{\PeerQwenBase}{60.5}\newcommand{\PeerQwenSed}{72.9}\newcommand{\PeerQwenD}{12.4}
\newcommand{\PeerGemBase}{32.0}\newcommand{\PeerGemSed}{58.9}\newcommand{\PeerGemD}{26.9}
\newcommand{\GnMajMacro}{24.6}
\newcommand{\GnReflMacro}{25.5}
\newcommand{\GnMajSd}{0.6}
\newcommand{\GnReflSd}{0.2}
\newcommand{\MhQBase}{53.8}\newcommand{\MhQAcc}{50.3}\newcommand{\MhQRev}{50.7}\newcommand{\MhQVer}{50.0}\newcommand{\MhQSed}{90.4}
\newcommand{\MyQBase}{84.1}\newcommand{\MyQAcc}{83.8}\newcommand{\MyQRev}{85.6}\newcommand{\MyQVer}{87.3}\newcommand{\MyQSed}{50.8}
\newcommand{\MhGBase}{24.7}\newcommand{\MhGAcc}{22.0}\newcommand{\MhGRev}{19.8}\newcommand{\MhGVer}{25.0}\newcommand{\MhGSed}{83.5}
\newcommand{\MyGBase}{93.4}\newcommand{\MyGAcc}{94.3}\newcommand{\MyGRev}{94.9}\newcommand{\MyGVer}{93.0}\newcommand{\MyGSed}{65.5}
\newcommand{\RbQBase}{98.8}\newcommand{\RbQAcc}{99.1}\newcommand{\RbQRev}{99.0}\newcommand{\RbQVer}{99.3}\newcommand{\RbQSed}{99.6}
\newcommand{\RwQBase}{3.3}\newcommand{\RwQAcc}{3.8}\newcommand{\RwQRev}{3.1}\newcommand{\RwQVer}{2.3}\newcommand{\RwQSed}{4.9}
\newcommand{\RbGBase}{94.9}\newcommand{\RbGAcc}{96.3}\newcommand{\RbGRev}{98.1}\newcommand{\RbGVer}{98.1}\newcommand{\RbGSed}{98.7}
\newcommand{\RwGBase}{9.8}\newcommand{\RwGAcc}{4.0}\newcommand{\RwGRev}{3.8}\newcommand{\RwGVer}{3.7}\newcommand{\RwGSed}{3.5}
\newcommand{\MgQBase}{46.2}\newcommand{\MgQAcc}{49.7}\newcommand{\MgQRev}{49.3}\newcommand{\MgQSed}{9.6}
\newcommand{\MgGBase}{75.3}\newcommand{\MgGAcc}{78.0}\newcommand{\MgGSed}{16.5}
\newcommand{\MgQSedY}{49.2}\newcommand{\MgGSedY}{34.5}
\newcommand{\DcSkillQ}{11.46}\newcommand{\DcSkillQLo}{8.79}\newcommand{\DcSkillQHi}{14.11}
\newcommand{\DcSituQ}{1.08}\newcommand{\DcSituQLo}{-0.09}\newcommand{\DcSituQHi}{2.16}
\newcommand{\DcSkillG}{16.15}\newcommand{\DcSkillGLo}{12.27}\newcommand{\DcSkillGHi}{19.99}
\newcommand{\DcSituG}{0.11}\newcommand{\DcSituGLo}{-1.56}\newcommand{\DcSituGHi}{1.84}
\newcommand{\DcGainQ}{12.6}\newcommand{\DcGainG}{16.3}
\newcommand{\DcSkillQAcc}{-1.21}\newcommand{\DcSkillQRev}{-1.03}\newcommand{\DcSkillQVer}{-0.98}
\newcommand{\DcSituQAcc}{1.38}\newcommand{\DcSituQRev}{0.70}\newcommand{\DcSituQVer}{1.78}
\newcommand{\DcSkillGAcc}{-1.53}\newcommand{\DcSkillGRev}{-1.77}\newcommand{\DcSkillGVer}{-0.44}
\newcommand{\DcSituGAcc}{1.47}\newcommand{\DcSituGRev}{1.90}\newcommand{\DcSituGVer}{1.80}
\newcommand{\DcNrungs}{17}\newcommand{\DcNext}{16}
\newcommand{\DcSkillQAccLo}{-2.27}\newcommand{\DcSkillQAccHi}{-0.23}
\newcommand{\DcSkillQRevLo}{-2.20}\newcommand{\DcSkillQRevHi}{0.11}
\newcommand{\DcSkillQVerLo}{-2.20}\newcommand{\DcSkillQVerHi}{0.28}
\newcommand{\DcSkillGAccLo}{-2.55}\newcommand{\DcSkillGAccHi}{-0.50}
\newcommand{\DcSkillGRevLo}{-2.91}\newcommand{\DcSkillGRevHi}{-0.72}
\newcommand{\DcSkillGVerLo}{-1.45}\newcommand{\DcSkillGVerHi}{0.62}
\newcommand{\DcSituQAccLo}{0.36}\newcommand{\DcSituQAccHi}{2.37}
\newcommand{\DcSituQRevLo}{-0.45}\newcommand{\DcSituQRevHi}{1.77}
\newcommand{\DcSituQVerLo}{0.65}\newcommand{\DcSituQVerHi}{3.03}
\newcommand{\DcSituGAccLo}{0.05}\newcommand{\DcSituGAccHi}{2.82}
\newcommand{\DcSituGRevLo}{0.32}\newcommand{\DcSituGRevHi}{3.47}
\newcommand{\DcSituGVerLo}{0.25}\newcommand{\DcSituGVerHi}{3.26}
\newcommand{\CsQBase}{63.96}\newcommand{\CsQAcc}{64.13}\newcommand{\CsQRev}{63.64}
\newcommand{\CsQVer}{64.77}\newcommand{\CsQSed}{76.52}
\newcommand{\CsGBase}{52.24}\newcommand{\CsGAcc}{52.18}\newcommand{\CsGRev}{52.37}
\newcommand{\CsGVer}{53.60}\newcommand{\CsGSed}{68.49}
\newcommand{\DcSkillCeilQ}{34.9}\newcommand{\DcSkillCeilG}{47.7}
\newcommand{\PiHQBase}{39.0}\newcommand{\PiYQBase}{16.2}\newcommand{\PiBQBase}{29.2}\newcommand{\PiWQBase}{15.6}
\newcommand{\PiHQAcc}{38.6}\newcommand{\PiYQAcc}{15.2}\newcommand{\PiBQAcc}{31.7}\newcommand{\PiWQAcc}{14.5}
\newcommand{\PiHQRev}{41.4}\newcommand{\PiYQRev}{14.1}\newcommand{\PiBQRev}{30.5}\newcommand{\PiWQRev}{14.1}
\newcommand{\PiHQVer}{39.5}\newcommand{\PiYQVer}{14.4}\newcommand{\PiBQVer}{32.4}\newcommand{\PiWQVer}{13.8}
\newcommand{\PiHQSed}{40.9}\newcommand{\PiYQSed}{12.1}\newcommand{\PiBQSed}{32.8}\newcommand{\PiWQSed}{14.1}
\newcommand{\PiHGBase}{23.5}\newcommand{\PiYGBase}{29.7}\newcommand{\PiBGBase}{16.5}\newcommand{\PiWGBase}{30.2}
\newcommand{\PiHGAcc}{35.2}\newcommand{\PiYGAcc}{22.3}\newcommand{\PiBGAcc}{23.5}\newcommand{\PiWGAcc}{19.0}
\newcommand{\PiHGRev}{33.3}\newcommand{\PiYGRev}{24.3}\newcommand{\PiBGRev}{22.4}\newcommand{\PiWGRev}{20.0}
\newcommand{\PiHGVer}{34.1}\newcommand{\PiYGVer}{23.0}\newcommand{\PiBGVer}{23.4}\newcommand{\PiWGVer}{19.5}
\newcommand{\PiHGSed}{37.8}\newcommand{\PiYGSed}{20.7}\newcommand{\PiBGSed}{23.1}\newcommand{\PiWGSed}{18.5}
\newcommand{\RngQBase}{47}\newcommand{\RngQSed}{9}\newcommand{\RngQRivLo}{38}\newcommand{\RngQRivHi}{40}
\newcommand{\RngGBase}{49}\newcommand{\RngGSed}{19}\newcommand{\RngGRivLo}{46}\newcommand{\RngGRivHi}{57}
\newcommand{\RngQRungs}{seventeen}\newcommand{\RngGRungs}{seventeen}
\newcommand{\LxN}{seventeen}
\newcommand{\LxBaseWorst}{16.8}\newcommand{\LxSedWorst}{2.5}
\newcommand{\LxBaseSpan}{25.6}\newcommand{\LxSedSpan}{10.3}
\newcommand{\LxGptFiveMiniBs}{51.4}\newcommand{\LxGptFiveMiniHs}{75.2}\newcommand{\LxGptFiveMiniBd}{-16.8}\newcommand{\LxGptFiveMiniHd}{+1.5}
\newcommand{\LxHaikuBs}{54.7}\newcommand{\LxHaikuHs}{72.1}\newcommand{\LxHaikuBd}{-13.5}\newcommand{\LxHaikuHd}{-1.6}
\newcommand{\LxGemThreeSevenBs}{55.4}\newcommand{\LxGemThreeSevenHs}{75.6}\newcommand{\LxGemThreeSevenBd}{-12.8}\newcommand{\LxGemThreeSevenHd}{+1.9}
\newcommand{\LxQwenThreeFourBs}{56.7}\newcommand{\LxQwenThreeFourHs}{71.2}\newcommand{\LxQwenThreeFourBd}{-11.5}\newcommand{\LxQwenThreeFourHd}{-2.5}
\newcommand{\LxGptFiveFourBs}{58.1}\newcommand{\LxGptFiveFourHs}{77.2}\newcommand{\LxGptFiveFourBd}{-10.1}\newcommand{\LxGptFiveFourHd}{+3.5}
\newcommand{\LxSelfBs}{60.5}\newcommand{\LxSelfHs}{74.0}\newcommand{\LxSelfBd}{-7.7}\newcommand{\LxSelfHd}{+0.3}
\newcommand{\LxQwenThreeEightHs}{72.9}\newcommand{\LxQwenThreeEightHd}{-0.8}
\newcommand{\LxGemmaNineBs}{64.6}\newcommand{\LxGemmaNineHs}{74.3}\newcommand{\LxGemmaNineBd}{-3.6}\newcommand{\LxGemmaNineHd}{+0.3}
\newcommand{\LxGptFiveThreeCodexBs}{65.3}\newcommand{\LxGptFiveThreeCodexHs}{80.7}\newcommand{\LxGptFiveThreeCodexBd}{-2.9}\newcommand{\LxGptFiveThreeCodexHd}{+7.0}
\newcommand{\LxQwenTwoFiveBs}{65.7}\newcommand{\LxQwenTwoFiveHs}{74.3}\newcommand{\LxQwenTwoFiveBd}{-2.4}\newcommand{\LxQwenTwoFiveHd}{+0.6}
\newcommand{\LxLlamaThreeOneBs}{67.1}\newcommand{\LxLlamaThreeOneHs}{74.3}\newcommand{\LxLlamaThreeOneBd}{-1.1}\newcommand{\LxLlamaThreeOneHd}{+0.6}
\newcommand{\LxGptFiveFourMiniBs}{69.8}\newcommand{\LxGptFiveFourMiniHs}{79.6}\newcommand{\LxGptFiveFourMiniBd}{+1.6}\newcommand{\LxGptFiveFourMiniHd}{+5.9}
\newcommand{\LxGptFiveSixTerraBs}{69.9}\newcommand{\LxGptFiveSixTerraHs}{80.7}\newcommand{\LxGptFiveSixTerraBd}{+1.7}\newcommand{\LxGptFiveSixTerraHd}{+7.0}
\newcommand{\LxSonnetFiveBs}{73.6}\newcommand{\LxSonnetFiveHs}{78.9}\newcommand{\LxSonnetFiveBd}{+2.3}\newcommand{\LxSonnetFiveHd}{+5.2}
\newcommand{\LxGptFiveFiveBs}{76.6}\newcommand{\LxGptFiveFiveHs}{81.3}\newcommand{\LxGptFiveFiveBd}{+3.9}\newcommand{\LxGptFiveFiveHd}{+7.6}
\newcommand{\LxGptFiveSixSolBs}{77.1}\newcommand{\LxGptFiveSixSolHs}{81.5}\newcommand{\LxGptFiveSixSolBd}{+0.9}\newcommand{\LxGptFiveSixSolHd}{+6.8}
\newcommand{\GnMajAgieval}{72.7}
\newcommand{\GsMajAgieval}{1.0}
\newcommand{\GnMajAime}{0.0}
\newcommand{\GsMajAime}{0.0}
\newcommand{\GnMajGpqa}{25.2}
\newcommand{\GsMajGpqa}{2.3}
\newcommand{\GnMajLcb}{10.0}
\newcommand{\GsMajLcb}{0.5}
\newcommand{\GnMajMmluRedux}{45.8}
\newcommand{\GsMajMmluRedux}{0.5}
\newcommand{\GnMajMusr}{35.8}
\newcommand{\GsMajMusr}{4.3}
\newcommand{\GnMajOmni}{7.8}
\newcommand{\GsMajOmni}{0.0}
\newcommand{\GnMajSupergpqa}{17.3}
\newcommand{\GsMajSupergpqa}{0.4}
\newcommand{\GnMajTheorem}{6.6}
\newcommand{\GsMajTheorem}{0.8}
\newcommand{\GnReflAgieval}{69.4}
\newcommand{\GsReflAgieval}{3.5}
\newcommand{\GnReflAime}{0.0}
\newcommand{\GsReflAime}{0.0}
\newcommand{\GnReflGpqa}{25.2}
\newcommand{\GsReflGpqa}{2.3}
\newcommand{\GnReflLcb}{12.4}
\newcommand{\GsReflLcb}{0.5}
\newcommand{\GnReflMmluRedux}{46.0}
\newcommand{\GsReflMmluRedux}{0.2}
\newcommand{\GnReflMusr}{43.6}
\newcommand{\GsReflMusr}{3.6}
\newcommand{\GnReflOmni}{8.2}
\newcommand{\GsReflOmni}{0.2}
\newcommand{\GnReflSupergpqa}{17.5}
\newcommand{\GsReflSupergpqa}{0.5}
\newcommand{\GnReflTheorem}{7.6}
\newcommand{\GsReflTheorem}{0.7}
\newcommand{\GnDegPlaneGiveLo}{26\,\%}
\newcommand{\GnDegPlaneGiveHi}{75\,\%}
\newcommand{\PnSepMeanPP}{25.0}
\newcommand{\GnSepMeanPP}{33.2}
\newcommand{\PnStratumN}{223}
\newcommand{\CopyThesisOnly}{0.693}
\newcommand{\CopyDialLo}{0.393}
\newcommand{\CopyDialHi}{0.419}
\newcommand{\CopyNoJudge}{0.705}
\newcommand{\FmtFailMax}{1}
\newcommand{\LgGemmaNineHs}{58.9}\newcommand{\LgGemmaNineHd}{-2.0}
\newcommand{\LgSelfBs}{32.0}\newcommand{\LgSelfBd}{-8.1}\newcommand{\LgSelfHs}{58.3}\newcommand{\LgSelfHd}{-2.6}
\newcommand{\LgLlamaThreeOneBs}{37.5}\newcommand{\LgLlamaThreeOneBd}{-2.6}\newcommand{\LgLlamaThreeOneHs}{59.7}\newcommand{\LgLlamaThreeOneHd}{-1.1}
\newcommand{\LgQwenTwoFiveBs}{41.3}\newcommand{\LgQwenTwoFiveBd}{+1.3}\newcommand{\LgQwenTwoFiveHs}{59.6}\newcommand{\LgQwenTwoFiveHd}{-1.2}
\newcommand{\LgQwenThreeFourBs}{41.7}\newcommand{\LgQwenThreeFourBd}{+1.6}\newcommand{\LgQwenThreeFourHs}{61.8}\newcommand{\LgQwenThreeFourHd}{+0.9}
\newcommand{\LgHaikuBs}{44.3}\newcommand{\LgHaikuBd}{-0.3}\newcommand{\LgHaikuHs}{62.8}\newcommand{\LgHaikuHd}{+2.0}
\newcommand{\LgQwenThreeEightBs}{45.0}\newcommand{\LgQwenThreeEightBd}{+4.4}\newcommand{\LgQwenThreeEightHs}{60.1}\newcommand{\LgQwenThreeEightHd}{-0.8}
\newcommand{\LgGptFiveMiniBs}{47.6}\newcommand{\LgGptFiveMiniBd}{-0.7}\newcommand{\LgGptFiveMiniHs}{72.4}\newcommand{\LgGptFiveMiniHd}{+11.6}
\newcommand{\LgGemThreeSevenBs}{49.0}\newcommand{\LgGemThreeSevenBd}{+6.2}\newcommand{\LgGemThreeSevenHs}{70.6}\newcommand{\LgGemThreeSevenHd}{+9.7}
\newcommand{\LgGptFiveFourBs}{60.4}\newcommand{\LgGptFiveFourBd}{+1.7}\newcommand{\LgGptFiveFourHs}{75.0}\newcommand{\LgGptFiveFourHd}{+14.1}
\newcommand{\LgGptFiveThreeCodexBs}{60.8}\newcommand{\LgGptFiveThreeCodexBd}{+0.5}\newcommand{\LgGptFiveThreeCodexHs}{76.6}\newcommand{\LgGptFiveThreeCodexHd}{+15.7}
\newcommand{\LgGptFiveFourMiniBs}{63.7}\newcommand{\LgGptFiveFourMiniBd}{+2.5}\newcommand{\LgGptFiveFourMiniHs}{74.0}\newcommand{\LgGptFiveFourMiniHd}{+12.8}
\newcommand{\LgGptFiveSixTerraBs}{66.1}\newcommand{\LgGptFiveSixTerraBd}{+1.3}\newcommand{\LgGptFiveSixTerraHs}{74.8}\newcommand{\LgGptFiveSixTerraHd}{+13.1}
\newcommand{\LgGptFiveFiveBs}{71.3}\newcommand{\LgGptFiveFiveBd}{+1.0}\newcommand{\LgGptFiveFiveHs}{77.6}\newcommand{\LgGptFiveFiveHd}{+16.7}
\newcommand{\LgSonnetFiveBs}{72.7}\newcommand{\LgSonnetFiveBd}{-3.1}\newcommand{\LgSonnetFiveHs}{79.3}\newcommand{\LgSonnetFiveHd}{+6.8}
\newcommand{\LgGptFiveSixSolBs}{76.3}\newcommand{\LgGptFiveSixSolBd}{+0.2}\newcommand{\LgGptFiveSixSolHs}{74.1}\newcommand{\LgGptFiveSixSolHd}{+2.0}
\newcommand{\LgBaseSpan}{44.3}\newcommand{\LgSedSpan}{21.0}
\newcommand{\OrBaseQ}{6.2}\newcommand{\OrSedQ}{9.7}
\newcommand{\OrBaseG}{4.6}\newcommand{\OrSedG}{9.6}
\newcommand{\OrWeakHoldRight}{nine items in ten}
\renewcommand{\LgSelfBs}{30.9}
\renewcommand{\LgSelfBd}{-9.1}
\renewcommand{\LgSelfHs}{59.3}
\renewcommand{\LgSelfHd}{-1.6}
\renewcommand{\LgBaseSpan}{45.4}
\renewcommand{\LgSedSpan}{20.4}
\renewcommand{\PeerGemBase}{30.9}
\renewcommand{\PeerGemD}{28.0}
\renewcommand{\CsGAcc}{51.84}
\renewcommand{\CsGBase}{52.17}
\renewcommand{\CsGRev}{52.16}
\renewcommand{\CsGSed}{68.55}
\renewcommand{\CsGVer}{53.47}
\renewcommand{\DcGainG}{16.4}
\renewcommand{\DcSituG}{0.13}
\renewcommand{\DcSituGAcc}{1.30}
\renewcommand{\DcSituGAccHi}{2.64}
\renewcommand{\DcSituGAccLo}{-0.15}
\renewcommand{\DcSituGHi}{1.82}
\renewcommand{\DcSituGLo}{-1.55}
\renewcommand{\DcSituGRev}{1.97}
\renewcommand{\DcSituGRevHi}{3.51}
\renewcommand{\DcSituGRevLo}{0.39}
\renewcommand{\DcSituGVer}{1.93}
\renewcommand{\DcSituGVerHi}{3.37}
\renewcommand{\DcSituGVerLo}{0.38}
\renewcommand{\DcSkillG}{16.25}
\renewcommand{\DcSkillGAcc}{-1.63}
\renewcommand{\DcSkillGAccHi}{-0.58}
\renewcommand{\DcSkillGAccLo}{-2.67}
\renewcommand{\DcSkillGHi}{20.09}
\renewcommand{\DcSkillGLo}{12.36}
\renewcommand{\DcSkillGRev}{-1.98}
\renewcommand{\DcSkillGRevHi}{-0.90}
\renewcommand{\DcSkillGRevLo}{-3.21}
\renewcommand{\DcSkillGVer}{-0.63}
\renewcommand{\DcSkillGVerHi}{0.47}
\renewcommand{\DcSkillGVerLo}{-1.70}
\renewcommand{\MgGAcc}{77.8}
\renewcommand{\MgGBase}{74.8}
\renewcommand{\MgGSed}{16.4}
\renewcommand{\MgGSedY}{34.9}
\renewcommand{\MhGAcc}{22.2}
\renewcommand{\MhGBase}{25.2}
\renewcommand{\MhGRev}{19.6}
\renewcommand{\MhGSed}{83.6}
\renewcommand{\MhGVer}{24.8}
\renewcommand{\MyGAcc}{94.5}
\renewcommand{\MyGBase}{93.2}
\renewcommand{\MyGRev}{95.0}
\renewcommand{\MyGSed}{65.1}
\renewcommand{\OrSedG}{9.5}
\renewcommand{\PiBGAcc}{23.0}
\renewcommand{\PiBGBase}{16.2}
\renewcommand{\PiBGRev}{22.0}
\renewcommand{\PiBGSed}{22.5}
\renewcommand{\PiBGVer}{23.3}
\renewcommand{\PiHGAcc}{35.7}
\renewcommand{\PiHGBase}{23.9}
\renewcommand{\PiHGRev}{33.6}
\renewcommand{\PiHGSed}{38.3}
\renewcommand{\PiHGVer}{34.2}
\renewcommand{\PiWGBase}{30.1}
\renewcommand{\PiWGRev}{19.9}
\renewcommand{\PiWGSed}{18.3}
\renewcommand{\PiYGBase}{29.8}
\renewcommand{\PiYGRev}{24.4}
\renewcommand{\PiYGSed}{20.8}
\renewcommand{\RbGAcc}{96.2}
\renewcommand{\RbGSed}{98.9}
\renewcommand{\RwGBase}{10.0}
\renewcommand{\RwGVer}{3.8}
\renewcommand{\PnSedAgieval}{99.1}
\renewcommand{\SdSedAgieval}{0.1}
\renewcommand{\PnSedAime}{25.4}
\renewcommand{\SdSedAime}{1.2}
\renewcommand{\PnSedGpqa}{44.4}
\renewcommand{\SdSedGpqa}{1.1}
\renewcommand{\PnSedLcb}{72.5}
\renewcommand{\SdSedLcb}{0.3}
\renewcommand{\PnSedMmluRedux}{82.4}
\renewcommand{\SdSedMmluRedux}{0.1}
\renewcommand{\PnSedMusr}{61.3}
\renewcommand{\SdSedMusr}{1.5}
\renewcommand{\PnSedOmni}{38.4}
\renewcommand{\SdSedOmni}{0.4}
\renewcommand{\PnSedSupergpqa}{39.4}
\renewcommand{\SdSedSupergpqa}{0.2}
\renewcommand{\PnSedTheorem}{40.3}
\renewcommand{\SdSedTheorem}{0.4}
\renewcommand{\PnSedMacro}{55.9}
\renewcommand{\PnSedSd}{0.4}
\renewcommand{\PnVerAgieval}{98.0}
\renewcommand{\SdVerAgieval}{0.4}
\renewcommand{\PnVerAime}{22.4}
\renewcommand{\SdVerAime}{2.4}
\renewcommand{\PnVerGpqa}{40.8}
\renewcommand{\SdVerGpqa}{0.5}
\renewcommand{\PnVerLcb}{69.8}
\renewcommand{\SdVerLcb}{0.5}
\renewcommand{\PnVerMmluRedux}{77.2}
\renewcommand{\SdVerMmluRedux}{0.1}
\renewcommand{\PnVerMusr}{53.4}
\renewcommand{\SdVerMusr}{1.1}
\renewcommand{\PnVerOmni}{36.3}
\renewcommand{\SdVerOmni}{0.2}
\renewcommand{\PnVerSupergpqa}{36.5}
\renewcommand{\SdVerSupergpqa}{0.1}
\renewcommand{\PnVerTheorem}{37.1}
\renewcommand{\SdVerTheorem}{0.7}
\renewcommand{\PnVerMacro}{52.4}
\renewcommand{\PnVerSd}{0.3}
\renewcommand{\PnRevAgieval}{97.6}
\renewcommand{\SdRevAgieval}{0.2}
\renewcommand{\PnRevAime}{20.8}
\renewcommand{\SdRevAime}{2.2}
\renewcommand{\PnRevGpqa}{39.8}
\renewcommand{\SdRevGpqa}{0.3}
\renewcommand{\PnRevLcb}{66.6}
\renewcommand{\SdRevLcb}{0.7}
\renewcommand{\PnRevMmluRedux}{75.1}
\renewcommand{\SdRevMmluRedux}{0.5}
\renewcommand{\PnRevMusr}{52.9}
\renewcommand{\SdRevMusr}{0.9}
\renewcommand{\PnRevOmni}{34.2}
\renewcommand{\SdRevOmni}{0.1}
\renewcommand{\PnRevSupergpqa}{35.1}
\renewcommand{\SdRevSupergpqa}{0.2}
\renewcommand{\PnRevTheorem}{34.9}
\renewcommand{\SdRevTheorem}{1.0}
\renewcommand{\PnRevMacro}{50.8}
\renewcommand{\PnRevSd}{0.3}
\renewcommand{\PnAccAgieval}{97.3}
\renewcommand{\SdAccAgieval}{0.4}
\renewcommand{\PnAccAime}{18.8}
\renewcommand{\SdAccAime}{2.9}
\renewcommand{\PnAccGpqa}{38.4}
\renewcommand{\SdAccGpqa}{1.2}
\renewcommand{\PnAccLcb}{65.4}
\renewcommand{\SdAccLcb}{0.8}
\renewcommand{\PnAccMmluRedux}{75.0}
\renewcommand{\SdAccMmluRedux}{0.6}
\renewcommand{\PnAccMusr}{52.9}
\renewcommand{\SdAccMusr}{0.9}
\renewcommand{\PnAccOmni}{33.6}
\renewcommand{\SdAccOmni}{0.2}
\renewcommand{\PnAccSupergpqa}{34.7}
\renewcommand{\SdAccSupergpqa}{0.3}
\renewcommand{\PnAccTheorem}{34.8}
\renewcommand{\SdAccTheorem}{1.0}
\renewcommand{\PnAccMacro}{50.1}
\renewcommand{\PnAccSd}{0.5}
\renewcommand{\PnPlcAgieval}{97.9}
\renewcommand{\SdPlcAgieval}{0.2}
\renewcommand{\PnPlcAime}{18.9}
\renewcommand{\SdPlcAime}{4.1}
\renewcommand{\PnPlcGpqa}{42.0}
\renewcommand{\SdPlcGpqa}{0.5}
\renewcommand{\PnPlcLcb}{67.2}
\renewcommand{\SdPlcLcb}{0.7}
\renewcommand{\PnPlcMmluRedux}{80.2}
\renewcommand{\SdPlcMmluRedux}{0.5}
\renewcommand{\PnPlcMusr}{58.9}
\renewcommand{\SdPlcMusr}{1.6}
\renewcommand{\PnPlcOmni}{35.0}
\renewcommand{\SdPlcOmni}{0.0}
\renewcommand{\PnPlcSupergpqa}{37.6}
\renewcommand{\SdPlcSupergpqa}{0.2}
\renewcommand{\PnPlcTheorem}{36.7}
\renewcommand{\SdPlcTheorem}{0.3}
\renewcommand{\PnPlcMacro}{52.7}
\renewcommand{\PnPlcSd}{0.4}
\renewcommand{\PnExtAgieval}{98.8}
\renewcommand{\SdExtAgieval}{0.1}
\renewcommand{\PnExtAime}{22.9}
\renewcommand{\SdExtAime}{0.8}
\renewcommand{\PnExtGpqa}{44.1}
\renewcommand{\SdExtGpqa}{0.3}
\renewcommand{\PnExtLcb}{73.2}
\renewcommand{\SdExtLcb}{0.8}
\renewcommand{\PnExtMmluRedux}{82.0}
\renewcommand{\SdExtMmluRedux}{0.1}
\renewcommand{\PnExtMusr}{59.4}
\renewcommand{\SdExtMusr}{0.6}
\renewcommand{\PnExtOmni}{37.2}
\renewcommand{\SdExtOmni}{0.1}
\renewcommand{\PnExtSupergpqa}{38.8}
\renewcommand{\SdExtSupergpqa}{0.1}
\renewcommand{\PnExtTheorem}{39.6}
\renewcommand{\SdExtTheorem}{0.4}
\renewcommand{\PnExtMacro}{55.1}
\renewcommand{\PnExtSd}{0.2}
\renewcommand{\GnSedAgieval}{87.6}
\renewcommand{\GsSedAgieval}{0.6}
\renewcommand{\GnSedAime}{0.3}
\renewcommand{\GsSedAime}{0.5}
\renewcommand{\GnSedGpqa}{33.1}
\renewcommand{\GsSedGpqa}{0.9}
\renewcommand{\GnSedLcb}{29.0}
\renewcommand{\GsSedLcb}{0.7}
\renewcommand{\GnSedMmluRedux}{74.1}
\renewcommand{\GsSedMmluRedux}{0.3}
\renewcommand{\GnSedMusr}{59.8}
\renewcommand{\GsSedMusr}{2.8}
\renewcommand{\GnSedOmni}{14.7}
\renewcommand{\GsSedOmni}{0.4}
\renewcommand{\GnSedSupergpqa}{24.5}
\renewcommand{\GsSedSupergpqa}{0.1}
\renewcommand{\GnSedTheorem}{17.5}
\renewcommand{\GsSedTheorem}{0.6}
\renewcommand{\GnSedMacro}{37.8}
\renewcommand{\GnSedSd}{0.1}
\renewcommand{\GnVerAgieval}{75.4}
\renewcommand{\GsVerAgieval}{3.6}
\renewcommand{\GnVerAime}{0.1}
\renewcommand{\GsVerAime}{0.2}
\renewcommand{\GnVerGpqa}{31.1}
\renewcommand{\GsVerGpqa}{0.4}
\renewcommand{\GnVerLcb}{16.6}
\renewcommand{\GsVerLcb}{1.7}
\renewcommand{\GnVerMmluRedux}{57.7}
\renewcommand{\GsVerMmluRedux}{0.6}
\renewcommand{\GnVerMusr}{41.5}
\renewcommand{\GsVerMusr}{1.1}
\renewcommand{\GnVerOmni}{10.1}
\renewcommand{\GsVerOmni}{0.9}
\renewcommand{\GnVerSupergpqa}{21.0}
\renewcommand{\GsVerSupergpqa}{0.4}
\renewcommand{\GnVerTheorem}{9.7}
\renewcommand{\GsVerTheorem}{0.6}
\renewcommand{\GnVerMacro}{29.3}
\renewcommand{\GnVerSd}{0.9}
\renewcommand{\GnRevAgieval}{73.8}
\renewcommand{\GsRevAgieval}{1.3}
\renewcommand{\GnRevAime}{0.1}
\renewcommand{\GsRevAime}{0.2}
\renewcommand{\GnRevGpqa}{30.4}
\renewcommand{\GsRevGpqa}{0.9}
\renewcommand{\GnRevLcb}{17.8}
\renewcommand{\GsRevLcb}{1.5}
\renewcommand{\GnRevMmluRedux}{55.3}
\renewcommand{\GsRevMmluRedux}{2.2}
\renewcommand{\GnRevMusr}{41.2}
\renewcommand{\GsRevMusr}{0.6}
\renewcommand{\GnRevOmni}{9.1}
\renewcommand{\GsRevOmni}{0.5}
\renewcommand{\GnRevSupergpqa}{19.8}
\renewcommand{\GsRevSupergpqa}{0.5}
\renewcommand{\GnRevTheorem}{9.0}
\renewcommand{\GsRevTheorem}{0.1}
\renewcommand{\GnRevMacro}{28.5}
\renewcommand{\GnRevSd}{0.5}
\renewcommand{\GnAccAgieval}{61.2}
\renewcommand{\GsAccAgieval}{1.7}
\renewcommand{\GnAccAime}{0.0}
\renewcommand{\GsAccAime}{0.0}
\renewcommand{\GnAccGpqa}{26.9}
\renewcommand{\GsAccGpqa}{0.7}
\renewcommand{\GnAccLcb}{11.0}
\renewcommand{\GsAccLcb}{0.4}
\renewcommand{\GnAccMmluRedux}{40.3}
\renewcommand{\GsAccMmluRedux}{1.9}
\renewcommand{\GnAccMusr}{38.5}
\renewcommand{\GsAccMusr}{1.6}
\renewcommand{\GnAccOmni}{7.5}
\renewcommand{\GsAccOmni}{0.2}
\renewcommand{\GnAccSupergpqa}{15.4}
\renewcommand{\GsAccSupergpqa}{0.3}
\renewcommand{\GnAccTheorem}{7.1}
\renewcommand{\GsAccTheorem}{0.4}
\renewcommand{\GnAccMacro}{23.1}
\renewcommand{\GnAccSd}{0.6}
\renewcommand{\PnSedD}{+5.90}
\renewcommand{\PnSedDLo}{+4.64}
\renewcommand{\PnSedDHi}{+7.16}
\renewcommand{\PnPlcD}{+2.70}
\renewcommand{\PnPlcDLo}{+1.66}
\renewcommand{\PnPlcDHi}{+3.73}
\renewcommand{\PnExtD}{+5.09}
\renewcommand{\PnExtDLo}{+3.66}
\renewcommand{\PnExtDHi}{+6.53}
\newcommand{\PnPlcRunGap}{2.5}
\renewcommand{\GemErrHi}{0.23\,\%}
\renewcommand{\PnReflAgieval}{96.6}
\renewcommand{\SdReflAgieval}{1.1}
\renewcommand{\PnReflAime}{17.5}
\renewcommand{\SdReflAime}{3.5}
\renewcommand{\PnReflGpqa}{38.1}
\renewcommand{\SdReflGpqa}{3.3}
\renewcommand{\PnReflLcb}{65.7}
\renewcommand{\SdReflLcb}{2.6}
\renewcommand{\PnReflMmluRedux}{75.1}
\renewcommand{\SdReflMmluRedux}{1.8}
\renewcommand{\PnReflMusr}{57.9}
\renewcommand{\SdReflMusr}{2.3}
\renewcommand{\PnReflOmni}{32.5}
\renewcommand{\SdReflOmni}{0.8}
\renewcommand{\PnReflSupergpqa}{34.0}
\renewcommand{\SdReflSupergpqa}{0.6}
\renewcommand{\PnReflTheorem}{33.8}
\renewcommand{\SdReflTheorem}{0.8}
\renewcommand{\PnReflMacro}{50.1}
\renewcommand{\PnReflSd}{0.8}
\renewcommand{\GnReflAgieval}{50.6}
\renewcommand{\GsReflAgieval}{5.6}
\renewcommand{\GnReflAime}{0.0}
\renewcommand{\GsReflAime}{0.0}
\renewcommand{\GnReflGpqa}{14.7}
\renewcommand{\GsReflGpqa}{2.5}
\renewcommand{\GnReflLcb}{10.8}
\renewcommand{\GsReflLcb}{1.0}
\renewcommand{\GnReflMmluRedux}{32.1}
\renewcommand{\GsReflMmluRedux}{3.8}
\renewcommand{\GnReflMusr}{30.0}
\renewcommand{\GsReflMusr}{5.0}
\renewcommand{\GnReflOmni}{7.3}
\renewcommand{\GsReflOmni}{0.6}
\renewcommand{\GnReflSupergpqa}{10.5}
\renewcommand{\GsReflSupergpqa}{0.8}
\renewcommand{\GnReflTheorem}{7.4}
\renewcommand{\GsReflTheorem}{1.3}
\renewcommand{\GnReflMacro}{18.2}
\renewcommand{\GnReflSd}{1.8}
\renewcommand{\PnMajAgieval}{97.6}
\renewcommand{\SdMajAgieval}{0.7}
\renewcommand{\PnMajAime}{24.4}
\renewcommand{\SdMajAime}{9.6}
\renewcommand{\PnMajGpqa}{42.1}
\renewcommand{\SdMajGpqa}{2.5}
\renewcommand{\PnMajLcb}{65.6}
\renewcommand{\SdMajLcb}{1.0}
\renewcommand{\PnMajMmluRedux}{77.4}
\renewcommand{\SdMajMmluRedux}{0.4}
\renewcommand{\PnMajMusr}{58.5}
\renewcommand{\SdMajMusr}{1.7}
\renewcommand{\PnMajOmni}{33.8}
\renewcommand{\SdMajOmni}{0.3}
\renewcommand{\PnMajSupergpqa}{35.4}
\renewcommand{\SdMajSupergpqa}{0.1}
\renewcommand{\PnMajTheorem}{35.0}
\renewcommand{\SdMajTheorem}{0.9}
\renewcommand{\PnMajMacro}{52.2}
\renewcommand{\PnMajSd}{1.2}
\renewcommand{\GnMajAgieval}{68.2}
\renewcommand{\GsMajAgieval}{0.9}
\renewcommand{\GnMajAime}{0.0}
\renewcommand{\GsMajAime}{0.0}
\renewcommand{\GnMajGpqa}{21.6}
\renewcommand{\GsMajGpqa}{1.8}
\renewcommand{\GnMajLcb}{10.0}
\renewcommand{\GsMajLcb}{0.5}
\renewcommand{\GnMajMmluRedux}{43.8}
\renewcommand{\GsMajMmluRedux}{0.6}
\renewcommand{\GnMajMusr}{35.2}
\renewcommand{\GsMajMusr}{4.2}
\renewcommand{\GnMajOmni}{8.0}
\renewcommand{\GsMajOmni}{0.0}
\renewcommand{\GnMajSupergpqa}{14.7}
\renewcommand{\GsMajSupergpqa}{0.3}
\renewcommand{\GnMajTheorem}{6.6}
\renewcommand{\GsMajTheorem}{0.8}
\renewcommand{\GnMajMacro}{23.1}
\renewcommand{\GnMajSd}{0.5}
\newcommand{\TTCReviewSameBase}{95}
\renewcommand{\EOneHOnePct}{1.0}
\renewcommand{\EOneHOneSDPct}{0.3}
\renewcommand{\EOneHOneAccPct}{73.5}
\renewcommand{\EOneHOneAccSDPct}{0.9}
\renewcommand{\EOneJZeroPct}{2.9}
\renewcommand{\EOneJZeroSDPct}{0.9}
\renewcommand{\EOneJZeroAccPct}{70.5}
\renewcommand{\EOneJZeroAccSDPct}{2.3}
\renewcommand{\EOneSZeroPct}{15.0}
\renewcommand{\EOneSZeroSDPct}{2.4}
\renewcommand{\EOneSZeroAccPct}{60.7}
\renewcommand{\EOneSZeroAccSDPct}{1.4}
\renewcommand{\TZeroKdrawPct}{14.0}
\renewcommand{\TZeroKdrawSDPct}{1.2}
\renewcommand{\TZeroAccPct}{61.8}
\renewcommand{\TZeroAccSDPct}{1.5}
\renewcommand{\MZeroKdrawPct}{13.9}
\renewcommand{\MZeroKdrawSDPct}{0.5}
\renewcommand{\MZeroAccPct}{61.7}
\renewcommand{\MZeroAccSDPct}{0.7}
\renewcommand{\EOneJZeroGapPct}{1.9}
\renewcommand{\EOneSZeroGapPct}{14.0}
\renewcommand{\PnSepMeanPP}{24.5}
\renewcommand{\GnSepMeanPP}{35.1}
\renewcommand{\CsGAcc}{51.35}
\renewcommand{\CsGRev}{53.17}
\renewcommand{\CsGSed}{67.30}
\renewcommand{\CsGVer}{53.49}
\renewcommand{\CsQAcc}{63.75}
\renewcommand{\CsQRev}{63.20}
\renewcommand{\CsQSed}{76.09}
\renewcommand{\CsQVer}{64.64}
\renewcommand{\DcGainG}{15.1}
\renewcommand{\DcGainQ}{12.2}
\renewcommand{\DcSituG}{0.64}
\renewcommand{\DcSituGAcc}{1.58}
\renewcommand{\DcSituGAccHi}{2.86}
\renewcommand{\DcSituGAccLo}{0.32}
\renewcommand{\DcSituGHi}{2.06}
\renewcommand{\DcSituGLo}{-0.79}
\renewcommand{\DcSituGRev}{2.16}
\renewcommand{\DcSituGRevHi}{3.49}
\renewcommand{\DcSituGRevLo}{0.75}
\renewcommand{\DcSituGVer}{1.89}
\renewcommand{\DcSituGVerHi}{3.20}
\renewcommand{\DcSituGVerLo}{0.49}
\renewcommand{\DcSituQ}{1.31}
\renewcommand{\DcSituQAcc}{1.11}
\renewcommand{\DcSituQAccHi}{1.95}
\renewcommand{\DcSituQAccLo}{0.30}
\renewcommand{\DcSituQHi}{2.22}
\renewcommand{\DcSituQLo}{0.36}
\renewcommand{\DcSituQRev}{0.82}
\renewcommand{\DcSituQRevHi}{1.70}
\renewcommand{\DcSituQRevLo}{-0.07}
\renewcommand{\DcSituQVer}{1.73}
\renewcommand{\DcSituQVerHi}{2.71}
\renewcommand{\DcSituQVerLo}{0.77}
\renewcommand{\DcSkillCeilG}{47.2}
\renewcommand{\DcSkillCeilQ}{34.7}
\renewcommand{\DcSkillG}{14.49}
\renewcommand{\DcSkillGAcc}{-2.41}
\renewcommand{\DcSkillGAccHi}{-1.44}
\renewcommand{\DcSkillGAccLo}{-3.41}
\renewcommand{\DcSkillGHi}{18.19}
\renewcommand{\DcSkillGLo}{10.71}
\renewcommand{\DcSkillGRev}{-1.16}
\renewcommand{\DcSkillGRevHi}{-0.09}
\renewcommand{\DcSkillGRevLo}{-2.22}
\renewcommand{\DcSkillGVer}{-0.57}
\renewcommand{\DcSkillGVerHi}{0.50}
\renewcommand{\DcSkillGVerLo}{-1.56}
\renewcommand{\DcSkillQ}{10.80}
\renewcommand{\DcSkillQAcc}{-1.33}
\renewcommand{\DcSkillQAccHi}{-0.36}
\renewcommand{\DcSkillQAccLo}{-2.38}
\renewcommand{\DcSkillQHi}{13.33}
\renewcommand{\DcSkillQLo}{8.25}
\renewcommand{\DcSkillQRev}{-1.58}
\renewcommand{\DcSkillQRevHi}{-0.64}
\renewcommand{\DcSkillQRevLo}{-2.58}
\renewcommand{\DcSkillQVer}{-1.05}
\renewcommand{\DcSkillQVerHi}{0.01}
\renewcommand{\DcSkillQVerLo}{-2.13}
\renewcommand{\LgGemThreeSevenHd}{+12.8}
\renewcommand{\LgGemThreeSevenHs}{71.2}
\renewcommand{\LgGemmaNineHd}{-0.7}
\renewcommand{\LgGemmaNineHs}{57.7}
\renewcommand{\LgGptFiveFiveHd}{+12.2}
\renewcommand{\LgGptFiveFiveHs}{75.2}
\renewcommand{\LgGptFiveFourHd}{+12.9}
\renewcommand{\LgGptFiveFourHs}{71.3}
\renewcommand{\LgGptFiveFourMiniHd}{+10.5}
\renewcommand{\LgGptFiveFourMiniHs}{72.8}
\renewcommand{\LgGptFiveMiniHd}{+12.9}
\renewcommand{\LgGptFiveMiniHs}{71.2}
\renewcommand{\LgGptFiveSixSolHd}{-2.3}
\renewcommand{\LgGptFiveSixSolHs}{72.3}
\renewcommand{\LgGptFiveSixTerraHd}{+9.7}
\renewcommand{\LgGptFiveSixTerraHs}{72.2}
\renewcommand{\LgGptFiveThreeCodexHd}{+13.1}
\renewcommand{\LgGptFiveThreeCodexHs}{74.3}
\renewcommand{\LgHaikuHd}{+5.8}
\renewcommand{\LgHaikuHs}{64.1}
\renewcommand{\LgLlamaThreeOneHd}{-0.4}
\renewcommand{\LgLlamaThreeOneHs}{58.0}
\renewcommand{\LgQwenThreeEightHd}{+0.6}
\renewcommand{\LgQwenThreeEightHs}{59.0}
\renewcommand{\LgQwenThreeFourHd}{+3.9}
\renewcommand{\LgQwenThreeFourHs}{62.3}
\renewcommand{\LgQwenTwoFiveHd}{0.0}
\renewcommand{\LgQwenTwoFiveHs}{58.4}
\renewcommand{\LgSedSpan}{20.0}
\renewcommand{\LgSelfHd}{-0.8}
\renewcommand{\LgSelfHs}{57.7}
\renewcommand{\LgSonnetFiveHd}{+2.4}
\renewcommand{\LgSonnetFiveHs}{77.5}
\renewcommand{\LxGemThreeSevenHd}{+1.5}
\renewcommand{\LxGemThreeSevenHs}{74.5}
\renewcommand{\LxGemmaNineHd}{+0.8}
\renewcommand{\LxGemmaNineHs}{73.9}
\renewcommand{\LxGptFiveFiveHd}{+8.4}
\renewcommand{\LxGptFiveFiveHs}{81.4}
\renewcommand{\LxGptFiveFourHd}{+3.5}
\renewcommand{\LxGptFiveFourHs}{76.5}
\renewcommand{\LxGptFiveFourMiniHd}{+6.1}
\renewcommand{\LxGptFiveFourMiniHs}{79.1}
\renewcommand{\LxGptFiveMiniHd}{+2.6}
\renewcommand{\LxGptFiveMiniHs}{75.6}
\renewcommand{\LxGptFiveSixSolHd}{+5.2}
\renewcommand{\LxGptFiveSixSolHs}{80.6}
\renewcommand{\LxGptFiveSixTerraHd}{+7.5}
\renewcommand{\LxGptFiveSixTerraHs}{80.4}
\renewcommand{\LxGptFiveThreeCodexHd}{+6.1}
\renewcommand{\LxGptFiveThreeCodexHs}{79.1}
\renewcommand{\LxHaikuHd}{-2.3}
\renewcommand{\LxHaikuHs}{70.6}
\renewcommand{\LxLlamaThreeOneHd}{+0.5}
\renewcommand{\LxLlamaThreeOneHs}{73.5}
\renewcommand{\LxQwenThreeEightHd}{+0.2}
\renewcommand{\LxQwenThreeEightHs}{73.2}
\renewcommand{\LxQwenThreeFourHd}{-1.4}
\renewcommand{\LxQwenThreeFourHs}{71.6}
\renewcommand{\LxQwenTwoFiveHd}{+0.5}
\renewcommand{\LxQwenTwoFiveHs}{73.5}
\renewcommand{\LxSedSpan}{11.1}
\renewcommand{\LxSedWorst}{2.6}
\renewcommand{\LxSelfHd}{+0.4}
\renewcommand{\LxSelfHs}{73.4}
\renewcommand{\LxSonnetFiveHd}{+6.4}
\renewcommand{\LxSonnetFiveHs}{79.5}
\renewcommand{\MgGAcc}{78.7}
\renewcommand{\MgGSed}{11.6}
\renewcommand{\MgGSedY}{43.7}
\renewcommand{\MgQAcc}{49.6}
\renewcommand{\MgQRev}{51.1}
\renewcommand{\MgQSed}{10.6}
\renewcommand{\MgQSedY}{46.8}
\renewcommand{\MhGAcc}{21.3}
\renewcommand{\MhGRev}{23.2}
\renewcommand{\MhGSed}{88.4}
\renewcommand{\MhGVer}{26.1}
\renewcommand{\MhQAcc}{50.4}
\renewcommand{\MhQRev}{48.9}
\renewcommand{\MhQSed}{89.4}
\renewcommand{\MhQVer}{49.8}
\renewcommand{\MyGAcc}{93.0}
\renewcommand{\MyGRev}{93.4}
\renewcommand{\MyGSed}{56.3}
\renewcommand{\MyGVer}{92.1}
\renewcommand{\MyQAcc}{83.7}
\renewcommand{\MyQRev}{86.2}
\renewcommand{\MyQSed}{53.2}
\renewcommand{\MyQVer}{86.5}
\renewcommand{\OrSedG}{9.8}
\renewcommand{\OrSedQ}{9.6}
\renewcommand{\PeerGemD}{26.8}
\renewcommand{\PeerGemSed}{57.7}
\renewcommand{\PeerQwenD}{12.7}
\renewcommand{\PeerQwenSed}{73.2}
\renewcommand{\PiBGAcc}{22.8}
\renewcommand{\PiBGRev}{22.8}
\renewcommand{\PiBGSed}{21.9}
\renewcommand{\PiBGVer}{22.8}
\renewcommand{\PiBQAcc}{31.3}
\renewcommand{\PiBQRev}{30.7}
\renewcommand{\PiBQSed}{32.6}
\renewcommand{\PiBQVer}{32.6}
\renewcommand{\PiHGAcc}{35.1}
\renewcommand{\PiHGRev}{34.3}
\renewcommand{\PiHGSed}{36.7}
\renewcommand{\PiHGVer}{35.0}
\renewcommand{\PiHQAcc}{39.7}
\renewcommand{\PiHQRev}{40.1}
\renewcommand{\PiHQSed}{40.4}
\renewcommand{\PiHQVer}{39.2}
\renewcommand{\PiWGAcc}{19.1}
\renewcommand{\PiWGRev}{19.1}
\renewcommand{\PiWGSed}{19.0}
\renewcommand{\PiWGVer}{18.9}
\renewcommand{\PiWQAcc}{14.3}
\renewcommand{\PiWQRev}{14.4}
\renewcommand{\PiWQSed}{14.0}
\renewcommand{\PiWQVer}{14.0}
\renewcommand{\PiYGAcc}{23.0}
\renewcommand{\PiYGRev}{23.7}
\renewcommand{\PiYGSed}{22.4}
\renewcommand{\PiYGVer}{23.3}
\renewcommand{\PiYQAcc}{14.7}
\renewcommand{\PiYQRev}{14.8}
\renewcommand{\PiYQSed}{12.9}
\renewcommand{\PiYQVer}{14.2}
\renewcommand{\RbGAcc}{96.3}
\renewcommand{\RbGRev}{98.1}
\renewcommand{\RbGSed}{98.5}
\renewcommand{\RbGVer}{97.5}
\renewcommand{\RbQAcc}{99.1}
\renewcommand{\RbQRev}{98.9}
\renewcommand{\RbQSed}{99.6}
\renewcommand{\RbQVer}{99.4}
\renewcommand{\RngGRivHi}{58}
\renewcommand{\RngGRivLo}{48}
\renewcommand{\RngGSed}{19}
\renewcommand{\RngQRivHi}{41}
\renewcommand{\RngQRivLo}{39}
\renewcommand{\RngQSed}{9}
\renewcommand{\RwGAcc}{3.2}
\renewcommand{\RwGRev}{3.5}
\renewcommand{\RwGSed}{3.9}
\renewcommand{\RwGVer}{3.4}
\renewcommand{\RwQAcc}{3.1}
\renewcommand{\RwQRev}{3.3}
\renewcommand{\RwQSed}{4.3}
\renewcommand{\RwQVer}{2.9}
\newcommand{\OrQweakestAll}{4.3}
\newcommand{\OrQweakestAllLo}{3.7}
\newcommand{\OrQweakestAllHi}{5.0}
\newcommand{\OrGweakestAll}{2.4}
\newcommand{\OrGweakestAllLo}{1.4}
\newcommand{\OrGweakestAllHi}{4.1}
\newcommand{\OrCellLoRun}{seven}
\newcommand{\OrCellHiRun}{twenty-three}
\newcommand{\DcSkillNposMin}{sixteen}
\newcommand{\DenomSpreadPct}{0.3}
\newcommand{\PnScoreAgieval}{98.5}
\newcommand{\SdScoreAgieval}{0.1}
\newcommand{\PnScoreAime}{22.9}
\newcommand{\SdScoreAime}{1.1}
\newcommand{\PnScoreGpqa}{41.5}
\newcommand{\SdScoreGpqa}{1.0}
\newcommand{\PnScoreLcb}{72.2}
\newcommand{\SdScoreLcb}{1.7}
\newcommand{\PnScoreMmluRedux}{78.1}
\newcommand{\SdScoreMmluRedux}{1.1}
\newcommand{\PnScoreMusr}{53.7}
\newcommand{\SdScoreMusr}{0.9}
\newcommand{\PnScoreOmni}{37.0}
\newcommand{\SdScoreOmni}{0.7}
\newcommand{\PnScoreSupergpqa}{37.6}
\newcommand{\SdScoreSupergpqa}{0.6}
\newcommand{\PnScoreTheorem}{37.5}
\newcommand{\SdScoreTheorem}{0.4}
\newcommand{\PnScoreMacro}{53.2}
\newcommand{\PnScoreSd}{0.5}
\newcommand{\GnScoreAgieval}{76.5}
\newcommand{\GsScoreAgieval}{3.6}
\newcommand{\GnScoreAime}{0.0}
\newcommand{\GsScoreAime}{0.0}
\newcommand{\GnScoreGpqa}{29.1}
\newcommand{\GsScoreGpqa}{1.0}
\newcommand{\GnScoreLcb}{23.6}
\newcommand{\GsScoreLcb}{4.2}
\newcommand{\GnScoreMmluRedux}{61.2}
\newcommand{\GsScoreMmluRedux}{3.1}
\newcommand{\GnScoreMusr}{41.3}
\newcommand{\GsScoreMusr}{0.8}
\newcommand{\GnScoreOmni}{16.1}
\newcommand{\GsScoreOmni}{0.9}
\newcommand{\GnScoreSupergpqa}{21.8}
\newcommand{\GsScoreSupergpqa}{1.1}
\newcommand{\GnScoreTheorem}{13.7}
\newcommand{\GsScoreTheorem}{1.9}
\newcommand{\GnScoreMacro}{31.5}
\newcommand{\GnScoreSd}{1.3}
\newcommand{\EOneScorePct}{12.2}
\newcommand{\EOneScoreSDPct}{0.6}
\newcommand{\EOneScoreAccPct}{64.9}
\newcommand{\EOneScoreAccSDPct}{0.7}
\newcommand{\GnScoreOmniLead}{1.4}
\newcommand{\DcSituQScore}{+1.03}
\newcommand{\DcSituQScoreLo}{-0.02}
\newcommand{\DcSituQScoreHi}{+2.09}
\newcommand{\DcSkillQScore}{-1.20}
\newcommand{\DcSkillQScoreLo}{-2.34}
\newcommand{\DcSkillQScoreHi}{-0.07}
\newcommand{\PiHQScore}{38.8}
\newcommand{\MhQScore}{51.0}
\newcommand{\PiYQScore}{14.2}
\newcommand{\MyQScore}{85.1}
\newcommand{\PiBQScore}{32.0}
\newcommand{\RbQScore}{99.2}
\newcommand{\PiWQScore}{14.9}
\newcommand{\RwQScore}{3.3}
\newcommand{\CsQScore}{64.13}
\newcommand{\OrScoreQ}{5.9}
\newcommand{\LxHaikuScs}{53.5}
\newcommand{\LxHaikuScd}{-17.3}
\newcommand{\LxSonnetFiveScs}{74.5}
\newcommand{\LxSonnetFiveScd}{+2.5}
\newcommand{\LxGemThreeSevenScs}{52.3}
\newcommand{\LxGemThreeSevenScd}{-18.5}
\newcommand{\LxGemmaNineScs}{65.6}
\newcommand{\LxGemmaNineScd}{-5.4}
\newcommand{\LxGptFiveMiniScs}{50.1}
\newcommand{\LxGptFiveMiniScd}{-20.8}
\newcommand{\LxGptFiveThreeCodexScs}{60.1}
\newcommand{\LxGptFiveThreeCodexScd}{-10.8}
\newcommand{\LxGptFiveFourScs}{60.8}
\newcommand{\LxGptFiveFourScd}{-10.1}
\newcommand{\LxGptFiveFourMiniScs}{68.8}
\newcommand{\LxGptFiveFourMiniScd}{-2.0}
\newcommand{\LxGptFiveFiveScs}{75.8}
\newcommand{\LxGptFiveFiveScd}{+4.6}
\newcommand{\LxGptFiveSixSolScs}{77.6}
\newcommand{\LxGptFiveSixSolScd}{+1.7}
\newcommand{\LxGptFiveSixTerraScs}{69.2}
\newcommand{\LxGptFiveSixTerraScd}{-1.7}
\newcommand{\LxLlamaThreeOneScs}{67.8}
\newcommand{\LxLlamaThreeOneScd}{-3.0}
\newcommand{\LxQwenTwoFiveScs}{65.6}
\newcommand{\LxQwenTwoFiveScd}{-5.2}
\newcommand{\LxQwenThreeFourScs}{56.2}
\newcommand{\LxQwenThreeFourScd}{-14.6}
\newcommand{\LxQwenThreeEightScs}{63.3}
\newcommand{\LxQwenThreeEightScd}{-7.5}
\newcommand{\LxSelfScs}{64.8}
\newcommand{\LxSelfScd}{-6.1}
\newcommand{\DcSituGScore}{+2.09}
\newcommand{\DcSituGScoreLo}{+0.71}
\newcommand{\DcSituGScoreHi}{+3.46}
\newcommand{\DcSkillGScore}{+0.25}
\newcommand{\DcSkillGScoreLo}{-1.06}
\newcommand{\DcSkillGScoreHi}{+1.61}
\newcommand{\PiHGScore}{32.8}
\newcommand{\MhGScore}{30.8}
\newcommand{\PiYGScore}{23.6}
\newcommand{\MyGScore}{89.8}
\newcommand{\PiBGScore}{22.9}
\newcommand{\RbGScore}{97.6}
\newcommand{\PiWGScore}{20.7}
\newcommand{\RwGScore}{5.2}
\newcommand{\CsGScore}{54.71}
\newcommand{\OrScoreG}{3.9}
\newcommand{\LgHaikuScs}{45.6}
\newcommand{\LgHaikuScd}{-10.1}
\newcommand{\LgSonnetFiveScs}{74.2}
\newcommand{\LgSonnetFiveScd}{-0.4}
\newcommand{\LgGemThreeSevenScs}{46.3}
\newcommand{\LgGemThreeSevenScd}{-9.4}
\newcommand{\LgGemmaNineScs}{38.7}
\newcommand{\LgGemmaNineScd}{-17.0}
\newcommand{\LgGptFiveMiniScs}{49.5}
\newcommand{\LgGptFiveMiniScd}{-6.1}
\newcommand{\LgGptFiveThreeCodexScs}{55.9}
\newcommand{\LgGptFiveThreeCodexScd}{-0.2}
\newcommand{\LgGptFiveFourScs}{57.7}
\newcommand{\LgGptFiveFourScd}{+0.6}
\newcommand{\LgGptFiveFourMiniScs}{65.2}
\newcommand{\LgGptFiveFourMiniScd}{+2.8}
\newcommand{\LgGptFiveFiveScs}{71.6}
\newcommand{\LgGptFiveFiveScd}{+4.0}
\newcommand{\LgGptFiveSixSolScs}{76.0}
\newcommand{\LgGptFiveSixSolScd}{-0.8}
\newcommand{\LgGptFiveSixTerraScs}{65.8}
\newcommand{\LgGptFiveSixTerraScd}{+0.7}
\newcommand{\LgLlamaThreeOneScs}{43.0}
\newcommand{\LgLlamaThreeOneScd}{-12.7}
\newcommand{\LgQwenTwoFiveScs}{44.4}
\newcommand{\LgQwenTwoFiveScd}{-11.3}
\newcommand{\LgQwenThreeFourScs}{48.3}
\newcommand{\LgQwenThreeFourScd}{-7.4}
\newcommand{\LgQwenThreeEightScs}{47.5}
\newcommand{\LgQwenThreeEightScd}{-8.2}
\newcommand{\LgSelfScs}{45.7}
\newcommand{\LgSelfScd}{-10.0}
\newcommand{\CtGAccAgieval}{85.2}
\newcommand{\CtGAccAime}{0.3}
\newcommand{\CtGAccGpqa}{30.7}
\newcommand{\CtGAccLcb}{27.3}
\newcommand{\CtGAccMacro}{37.0}
\newcommand{\CtGAccMmluRedux}{73.5}
\newcommand{\CtGAccMusr}{59.8}
\newcommand{\CtGAccOmni}{15.6}
\newcommand{\CtGAccSupergpqa}{22.3}
\newcommand{\CtGAccTheorem}{18.7}
\newcommand{\CtGBaseAgieval}{45.0}
\newcommand{\CtGBaseAime}{0.0}
\newcommand{\CtGBaseGpqa}{22.5}
\newcommand{\CtGBaseLcb}{25.2}
\newcommand{\CtGBaseMacro}{27.0}
\newcommand{\CtGBaseMmluRedux}{50.6}
\newcommand{\CtGBaseMusr}{51.1}
\newcommand{\CtGBaseOmni}{14.6}
\newcommand{\CtGBaseSupergpqa}{15.4}
\newcommand{\CtGBaseTheorem}{18.7}
\newcommand{\CtGRevAgieval}{82.3}
\newcommand{\CtGRevAime}{0.4}
\newcommand{\CtGRevGpqa}{29.9}
\newcommand{\CtGRevLcb}{28.6}
\newcommand{\CtGRevMacro}{36.9}
\newcommand{\CtGRevMmluRedux}{73.4}
\newcommand{\CtGRevMusr}{61.3}
\newcommand{\CtGRevOmni}{15.4}
\newcommand{\CtGRevSupergpqa}{22.6}
\newcommand{\CtGRevTheorem}{18.1}
\newcommand{\CtGRivHi}{37.1}
\newcommand{\CtGRivLo}{32.9}
\newcommand{\CtGScoreAgieval}{66.7}
\newcommand{\CtGScoreAime}{0.1}
\newcommand{\CtGScoreGpqa}{27.5}
\newcommand{\CtGScoreLcb}{25.7}
\newcommand{\CtGScoreMacro}{32.9}
\newcommand{\CtGScoreMmluRedux}{68.5}
\newcommand{\CtGScoreMusr}{51.1}
\newcommand{\CtGScoreOmni}{17.1}
\newcommand{\CtGScoreSupergpqa}{21.5}
\newcommand{\CtGScoreTheorem}{17.9}
\newcommand{\CtGSedAgieval}{82.5}
\newcommand{\CtGSedAime}{0.6}
\newcommand{\CtGSedGpqa}{31.6}
\newcommand{\CtGSedLcb}{26.9}
\newcommand{\CtGSedMacro}{36.9}
\newcommand{\CtGSedMmluRedux}{73.5}
\newcommand{\CtGSedMusr}{59.6}
\newcommand{\CtGSedOmni}{16.0}
\newcommand{\CtGSedSupergpqa}{22.9}
\newcommand{\CtGSedTheorem}{18.6}
\newcommand{\CtGVerAgieval}{84.2}
\newcommand{\CtGVerAime}{0.1}
\newcommand{\CtGVerGpqa}{30.4}
\newcommand{\CtGVerLcb}{27.8}
\newcommand{\CtGVerMacro}{37.1}
\newcommand{\CtGVerMmluRedux}{73.8}
\newcommand{\CtGVerMusr}{59.3}
\newcommand{\CtGVerOmni}{16.2}
\newcommand{\CtGVerSupergpqa}{23.6}
\newcommand{\CtGVerTheorem}{18.6}
\newcommand{\CtQAccAgieval}{98.7}
\newcommand{\CtQAccAime}{20.0}
\newcommand{\CtQAccGpqa}{41.6}
\newcommand{\CtQAccLcb}{70.4}
\newcommand{\CtQAccMacro}{54.1}
\newcommand{\CtQAccMmluRedux}{82.2}
\newcommand{\CtQAccMusr}{60.7}
\newcommand{\CtQAccOmni}{35.9}
\newcommand{\CtQAccSupergpqa}{37.7}
\newcommand{\CtQAccTheorem}{39.5}
\newcommand{\CtQBaseAgieval}{98.4}
\newcommand{\CtQBaseAime}{19.6}
\newcommand{\CtQBaseGpqa}{42.1}
\newcommand{\CtQBaseLcb}{71.0}
\newcommand{\CtQBaseMacro}{53.3}
\newcommand{\CtQBaseMmluRedux}{81.2}
\newcommand{\CtQBaseMusr}{57.7}
\newcommand{\CtQBaseOmni}{34.4}
\newcommand{\CtQBaseSupergpqa}{36.5}
\newcommand{\CtQBaseTheorem}{38.4}
\newcommand{\CtQRevAgieval}{98.8}
\newcommand{\CtQRevAime}{21.0}
\newcommand{\CtQRevGpqa}{42.0}
\newcommand{\CtQRevLcb}{71.2}
\newcommand{\CtQRevMacro}{54.2}
\newcommand{\CtQRevMmluRedux}{82.4}
\newcommand{\CtQRevMusr}{59.1}
\newcommand{\CtQRevOmni}{36.2}
\newcommand{\CtQRevSupergpqa}{37.9}
\newcommand{\CtQRevTheorem}{39.5}
\newcommand{\CtQRivHi}{55.3}
\newcommand{\CtQRivLo}{54.1}
\newcommand{\CtQScoreAgieval}{98.8}
\newcommand{\CtQScoreAime}{22.8}
\newcommand{\CtQScoreGpqa}{42.8}
\newcommand{\CtQScoreLcb}{75.2}
\newcommand{\CtQScoreMacro}{55.3}
\newcommand{\CtQScoreMmluRedux}{82.7}
\newcommand{\CtQScoreMusr}{58.1}
\newcommand{\CtQScoreOmni}{37.9}
\newcommand{\CtQScoreSupergpqa}{38.7}
\newcommand{\CtQScoreTheorem}{40.4}
\newcommand{\CtQSedAgieval}{99.0}
\newcommand{\CtQSedAime}{24.4}
\newcommand{\CtQSedGpqa}{43.9}
\newcommand{\CtQSedLcb}{73.4}
\newcommand{\CtQSedMacro}{55.6}
\newcommand{\CtQSedMmluRedux}{82.5}
\newcommand{\CtQSedMusr}{60.2}
\newcommand{\CtQSedOmni}{38.0}
\newcommand{\CtQSedSupergpqa}{38.4}
\newcommand{\CtQSedTheorem}{40.1}
\newcommand{\CtQVerAgieval}{98.9}
\newcommand{\CtQVerAime}{22.6}
\newcommand{\CtQVerGpqa}{42.1}
\newcommand{\CtQVerLcb}{72.8}
\newcommand{\CtQVerMacro}{54.9}
\newcommand{\CtQVerMmluRedux}{82.5}
\newcommand{\CtQVerMusr}{59.5}
\newcommand{\CtQVerOmni}{37.2}
\newcommand{\CtQVerSupergpqa}{38.2}
\newcommand{\CtQVerTheorem}{39.9}
\newcommand{\CtSdGAccAgieval}{0.4}
\newcommand{\CtSdGAccAime}{0.5}
\newcommand{\CtSdGAccGpqa}{1.0}
\newcommand{\CtSdGAccLcb}{0.2}
\newcommand{\CtSdGAccMacro}{0.4}
\newcommand{\CtSdGAccMmluRedux}{0.4}
\newcommand{\CtSdGAccMusr}{2.4}
\newcommand{\CtSdGAccOmni}{0.4}
\newcommand{\CtSdGAccSupergpqa}{0.3}
\newcommand{\CtSdGAccTheorem}{0.2}
\newcommand{\CtSdGBaseAgieval}{5.3}
\newcommand{\CtSdGBaseAime}{0.0}
\newcommand{\CtSdGBaseGpqa}{1.9}
\newcommand{\CtSdGBaseLcb}{2.7}
\newcommand{\CtSdGBaseMacro}{1.3}
\newcommand{\CtSdGBaseMmluRedux}{3.6}
\newcommand{\CtSdGBaseMusr}{5.2}
\newcommand{\CtSdGBaseOmni}{0.3}
\newcommand{\CtSdGBaseSupergpqa}{0.6}
\newcommand{\CtSdGBaseTheorem}{1.4}
\newcommand{\CtSdGRevAgieval}{1.1}
\newcommand{\CtSdGRevAime}{0.4}
\newcommand{\CtSdGRevGpqa}{1.4}
\newcommand{\CtSdGRevLcb}{0.5}
\newcommand{\CtSdGRevMacro}{0.4}
\newcommand{\CtSdGRevMmluRedux}{0.1}
\newcommand{\CtSdGRevMusr}{0.4}
\newcommand{\CtSdGRevOmni}{0.1}
\newcommand{\CtSdGRevSupergpqa}{0.2}
\newcommand{\CtSdGRevTheorem}{0.4}
\newcommand{\CtSdGScoreAgieval}{10.8}
\newcommand{\CtSdGScoreAime}{0.2}
\newcommand{\CtSdGScoreGpqa}{3.9}
\newcommand{\CtSdGScoreLcb}{3.9}
\newcommand{\CtSdGScoreMacro}{3.7}
\newcommand{\CtSdGScoreMmluRedux}{6.7}
\newcommand{\CtSdGScoreMusr}{7.8}
\newcommand{\CtSdGScoreOmni}{0.8}
\newcommand{\CtSdGScoreSupergpqa}{3.3}
\newcommand{\CtSdGScoreTheorem}{1.0}
\newcommand{\CtSdGSedAgieval}{1.2}
\newcommand{\CtSdGSedAime}{0.6}
\newcommand{\CtSdGSedGpqa}{0.5}
\newcommand{\CtSdGSedLcb}{3.3}
\newcommand{\CtSdGSedMacro}{0.5}
\newcommand{\CtSdGSedMmluRedux}{0.4}
\newcommand{\CtSdGSedMusr}{0.9}
\newcommand{\CtSdGSedOmni}{0.2}
\newcommand{\CtSdGSedSupergpqa}{0.6}
\newcommand{\CtSdGSedTheorem}{0.6}
\newcommand{\CtSdGVerAgieval}{1.4}
\newcommand{\CtSdGVerAime}{0.2}
\newcommand{\CtSdGVerGpqa}{0.5}
\newcommand{\CtSdGVerLcb}{1.5}
\newcommand{\CtSdGVerMacro}{0.3}
\newcommand{\CtSdGVerMmluRedux}{0.8}
\newcommand{\CtSdGVerMusr}{1.7}
\newcommand{\CtSdGVerOmni}{0.2}
\newcommand{\CtSdGVerSupergpqa}{0.3}
\newcommand{\CtSdGVerTheorem}{0.4}
\newcommand{\CtSdQAccAgieval}{0.1}
\newcommand{\CtSdQAccAime}{1.5}
\newcommand{\CtSdQAccGpqa}{0.3}
\newcommand{\CtSdQAccLcb}{0.3}
\newcommand{\CtSdQAccMacro}{0.1}
\newcommand{\CtSdQAccMmluRedux}{0.1}
\newcommand{\CtSdQAccMusr}{0.5}
\newcommand{\CtSdQAccOmni}{0.2}
\newcommand{\CtSdQAccSupergpqa}{0.1}
\newcommand{\CtSdQAccTheorem}{0.4}
\newcommand{\CtSdQBaseAgieval}{0.5}
\newcommand{\CtSdQBaseAime}{5.2}
\newcommand{\CtSdQBaseGpqa}{1.2}
\newcommand{\CtSdQBaseLcb}{1.6}
\newcommand{\CtSdQBaseMacro}{0.5}
\newcommand{\CtSdQBaseMmluRedux}{0.7}
\newcommand{\CtSdQBaseMusr}{1.9}
\newcommand{\CtSdQBaseOmni}{0.7}
\newcommand{\CtSdQBaseSupergpqa}{0.2}
\newcommand{\CtSdQBaseTheorem}{1.4}
\newcommand{\CtSdQRevAgieval}{0.1}
\newcommand{\CtSdQRevAime}{1.7}
\newcommand{\CtSdQRevGpqa}{1.1}
\newcommand{\CtSdQRevLcb}{0.9}
\newcommand{\CtSdQRevMacro}{0.4}
\newcommand{\CtSdQRevMmluRedux}{0.1}
\newcommand{\CtSdQRevMusr}{1.2}
\newcommand{\CtSdQRevOmni}{0.0}
\newcommand{\CtSdQRevSupergpqa}{0.2}
\newcommand{\CtSdQRevTheorem}{0.3}
\newcommand{\CtSdQScoreAgieval}{0.1}
\newcommand{\CtSdQScoreAime}{0.6}
\newcommand{\CtSdQScoreGpqa}{0.7}
\newcommand{\CtSdQScoreLcb}{1.2}
\newcommand{\CtSdQScoreMacro}{0.3}
\newcommand{\CtSdQScoreMmluRedux}{0.2}
\newcommand{\CtSdQScoreMusr}{0.9}
\newcommand{\CtSdQScoreOmni}{0.8}
\newcommand{\CtSdQScoreSupergpqa}{0.4}
\newcommand{\CtSdQScoreTheorem}{0.5}
\newcommand{\CtSdQSedAgieval}{0.2}
\newcommand{\CtSdQSedAime}{1.7}
\newcommand{\CtSdQSedGpqa}{1.1}
\newcommand{\CtSdQSedLcb}{0.6}
\newcommand{\CtSdQSedMacro}{0.4}
\newcommand{\CtSdQSedMmluRedux}{0.1}
\newcommand{\CtSdQSedMusr}{1.7}
\newcommand{\CtSdQSedOmni}{0.5}
\newcommand{\CtSdQSedSupergpqa}{0.2}
\newcommand{\CtSdQSedTheorem}{0.4}
\newcommand{\CtSdQVerAgieval}{0.1}
\newcommand{\CtSdQVerAime}{2.3}
\newcommand{\CtSdQVerGpqa}{1.1}
\newcommand{\CtSdQVerLcb}{0.5}
\newcommand{\CtSdQVerMacro}{0.4}
\newcommand{\CtSdQVerMmluRedux}{0.1}
\newcommand{\CtSdQVerMusr}{1.0}
\newcommand{\CtSdQVerOmni}{0.1}
\newcommand{\CtSdQVerSupergpqa}{0.1}
\newcommand{\CtSdQVerTheorem}{0.7}
\newcommand{\RpGAccAll}{8.0}
\newcommand{\RpGAccOne}{35.6}
\newcommand{\RpGAccThree}{26.7}
\newcommand{\RpGAccTwo}{27.7}
\newcommand{\RpGAllOthHi}{26.0}
\newcommand{\RpGBaseAll}{6.5}
\newcommand{\RpGBaseOne}{32.0}
\newcommand{\RpGBaseThree}{21.0}
\newcommand{\RpGBaseTwo}{23.9}
\newcommand{\RpGRevAll}{17.7}
\newcommand{\RpGRevOne}{40.5}
\newcommand{\RpGRevThree}{35.6}
\newcommand{\RpGRevTwo}{40.2}
\newcommand{\RpGSedAll}{48.1}
\newcommand{\RpGSedOne}{57.1}
\newcommand{\RpGSedThree}{52.4}
\newcommand{\RpGSedTwo}{54.1}
\newcommand{\RpGThreeOthHi}{41.7}
\newcommand{\RpGVerAll}{20.9}
\newcommand{\RpGVerOne}{44.2}
\newcommand{\RpGVerThree}{38.9}
\newcommand{\RpGVerTwo}{43.3}
\newcommand{\RpQAccAll}{51.0}
\newcommand{\RpQAccOne}{62.9}
\newcommand{\RpQAccThree}{61.8}
\newcommand{\RpQAccTwo}{63.4}
\newcommand{\RpQAllOthHi}{53.9}
\newcommand{\RpQBaseAll}{50.2}
\newcommand{\RpQBaseOne}{60.0}
\newcommand{\RpQBaseThree}{61.7}
\newcommand{\RpQBaseTwo}{63.1}
\newcommand{\RpQRevAll}{52.3}
\newcommand{\RpQRevOne}{63.9}
\newcommand{\RpQRevThree}{62.7}
\newcommand{\RpQRevTwo}{65.8}
\newcommand{\RpQSedAll}{70.0}
\newcommand{\RpQSedOne}{73.1}
\newcommand{\RpQSedThree}{72.0}
\newcommand{\RpQSedTwo}{72.3}
\newcommand{\RpQThreeOthHi}{64.8}
\newcommand{\RpQVerAll}{53.9}
\newcommand{\RpQVerOne}{66.3}
\newcommand{\RpQVerThree}{64.8}
\newcommand{\RpQVerTwo}{66.4}
\newcommand{\RpGScoreAll}{26.0}
\newcommand{\RpGScoreOne}{46.2}
\newcommand{\RpGScoreThree}{41.7}
\newcommand{\RpGScoreTwo}{43.4}
\newcommand{\RpQScoreAll}{52.9}
\newcommand{\RpQScoreOne}{65.2}
\newcommand{\RpQScoreThree}{63.8}
\newcommand{\RpQScoreTwo}{66.8}
\renewcommand{\GnDegPlaneGiveLo}{25\,\%}
\renewcommand{\CsGAcc}{50.97}
\renewcommand{\CsGBase}{52.05}
\renewcommand{\CsGRev}{52.94}
\renewcommand{\CsGScore}{54.02}
\renewcommand{\CsGSed}{67.42}
\renewcommand{\CsGVer}{53.34}
\renewcommand{\CsQAcc}{63.19}
\renewcommand{\CsQBase}{63.49}
\renewcommand{\CsQRev}{62.72}
\renewcommand{\CsQScore}{63.07}
\renewcommand{\CsQSed}{75.84}
\renewcommand{\CsQVer}{64.11}
\renewcommand{\DcGainG}{15.4}
\renewcommand{\DcGainQ}{12.3}
\renewcommand{\DcSituG}{0.54}
\renewcommand{\DcSituGAcc}{1.42}
\renewcommand{\DcSituGAccHi}{2.66}
\renewcommand{\DcSituGAccLo}{0.15}
\renewcommand{\DcSituGHi}{1.93}
\renewcommand{\DcSituGLo}{-0.90}
\renewcommand{\DcSituGRev}{1.96}
\renewcommand{\DcSituGRevHi}{3.27}
\renewcommand{\DcSituGRevLo}{0.59}
\renewcommand{\DcSituGScore}{+1.80}
\renewcommand{\DcSituGScoreHi}{+3.17}
\renewcommand{\DcSituGScoreLo}{+0.41}
\renewcommand{\DcSituGVer}{1.73}
\renewcommand{\DcSituGVerHi}{3.00}
\renewcommand{\DcSituGVerLo}{0.32}
\renewcommand{\DcSituQ}{1.26}
\renewcommand{\DcSituQAcc}{1.01}
\renewcommand{\DcSituQAccHi}{1.83}
\renewcommand{\DcSituQAccLo}{0.21}
\renewcommand{\DcSituQHi}{2.14}
\renewcommand{\DcSituQLo}{0.33}
\renewcommand{\DcSituQRev}{0.81}
\renewcommand{\DcSituQRevHi}{1.65}
\renewcommand{\DcSituQRevLo}{-0.05}
\renewcommand{\DcSituQScore}{+0.85}
\renewcommand{\DcSituQScoreHi}{+1.82}
\renewcommand{\DcSituQScoreLo}{-0.12}
\renewcommand{\DcSituQVer}{1.65}
\renewcommand{\DcSituQVerHi}{2.59}
\renewcommand{\DcSituQVerLo}{0.74}
\renewcommand{\DcSkillCeilG}{47.4}
\renewcommand{\DcSkillCeilQ}{35.2}
\renewcommand{\DcSkillG}{14.83}
\renewcommand{\DcSkillGAcc}{-2.50}
\renewcommand{\DcSkillGAccHi}{-1.52}
\renewcommand{\DcSkillGAccLo}{-3.48}
\renewcommand{\DcSkillGHi}{18.58}
\renewcommand{\DcSkillGLo}{10.99}
\renewcommand{\DcSkillGRev}{-1.07}
\renewcommand{\DcSkillGRevHi}{-0.01}
\renewcommand{\DcSkillGRevLo}{-2.17}
\renewcommand{\DcSkillGScore}{+0.17}
\renewcommand{\DcSkillGScoreHi}{+1.51}
\renewcommand{\DcSkillGVer}{-0.44}
\renewcommand{\DcSkillGVerHi}{0.61}
\renewcommand{\DcSkillGVerLo}{-1.42}
\renewcommand{\DcSkillQ}{11.08}
\renewcommand{\DcSkillQAcc}{-1.32}
\renewcommand{\DcSkillQAccHi}{-0.37}
\renewcommand{\DcSkillQAccLo}{-2.33}
\renewcommand{\DcSkillQHi}{13.58}
\renewcommand{\DcSkillQLo}{8.52}
\renewcommand{\DcSkillQRevLo}{-2.56}
\renewcommand{\DcSkillQScore}{-1.29}
\renewcommand{\DcSkillQScoreHi}{-0.16}
\renewcommand{\DcSkillQScoreLo}{-2.41}
\renewcommand{\DcSkillQVer}{-1.03}
\renewcommand{\DcSkillQVerLo}{-2.11}
\renewcommand{\DegPlaneGiveHi}{61\,\%}
\renewcommand{\DenomSpreadPct}{0.4}
\renewcommand{\GnDegPlaneGiveHi}{83\,\%}
\renewcommand{\GnSepMeanPP}{36.7}
\renewcommand{\LxBaseSpan}{26.5}
\renewcommand{\LxBaseWorst}{17.6}
\renewcommand{\MgGAcc}{78.9}
\renewcommand{\MgGBase}{74.9}
\renewcommand{\MgGSed}{10.5}
\renewcommand{\MgGSedY}{44.6}
\renewcommand{\MgQAcc}{50.0}
\renewcommand{\MgQBase}{46.7}
\renewcommand{\MgQRev}{51.5}
\renewcommand{\MgQSed}{11.1}
\renewcommand{\MgQSedY}{46.7}
\renewcommand{\MhGAcc}{21.1}
\renewcommand{\MhGBase}{25.1}
\renewcommand{\MhGRev}{23.4}
\renewcommand{\MhGScore}{28.9}
\renewcommand{\MhGSed}{89.5}
\renewcommand{\MhGVer}{26.4}
\renewcommand{\MhQAcc}{50.0}
\renewcommand{\MhQBase}{53.3}
\renewcommand{\MhQRev}{48.5}
\renewcommand{\MhQScore}{49.1}
\renewcommand{\MhQSed}{88.9}
\renewcommand{\MhQVer}{49.6}
\renewcommand{\MyGAcc}{92.8}
\renewcommand{\MyGRev}{93.2}
\renewcommand{\MyGScore}{90.2}
\renewcommand{\MyGSed}{55.4}
\renewcommand{\MyQAcc}{83.6}
\renewcommand{\MyQRev}{86.3}
\renewcommand{\MyQScore}{86.1}
\renewcommand{\MyQSed}{53.3}
\renewcommand{\MyQVer}{86.3}
\renewcommand{\OrBaseQ}{6.0}
\renewcommand{\OrScoreG}{3.7}
\renewcommand{\OrScoreQ}{6.0}
\renewcommand{\OrSedG}{10.6}
\renewcommand{\OrSedQ}{9.1}
\renewcommand{\PiBGAcc}{22.2}
\renewcommand{\PiBGBase}{15.8}
\renewcommand{\PiBGRev}{22.3}
\renewcommand{\PiBGScore}{21.8}
\renewcommand{\PiBGSed}{21.3}
\renewcommand{\PiBGVer}{22.5}
\renewcommand{\PiBQAcc}{30.5}
\renewcommand{\PiBQBase}{28.3}
\renewcommand{\PiBQRev}{29.9}
\renewcommand{\PiBQScore}{30.4}
\renewcommand{\PiBQSed}{31.7}
\renewcommand{\PiBQVer}{31.9}
\renewcommand{\PiHGAcc}{35.6}
\renewcommand{\PiHGBase}{24.3}
\renewcommand{\PiHGRev}{34.8}
\renewcommand{\PiHGScore}{33.9}
\renewcommand{\PiHGSed}{37.1}
\renewcommand{\PiHGVer}{35.3}
\renewcommand{\PiHQAcc}{40.5}
\renewcommand{\PiHQBase}{39.9}
\renewcommand{\PiHQRev}{40.7}
\renewcommand{\PiHQScore}{40.4}
\renewcommand{\PiHQSed}{41.2}
\renewcommand{\PiHQVer}{40.0}
\renewcommand{\PiWGAcc}{19.0}
\renewcommand{\PiWGBase}{29.9}
\renewcommand{\PiWGRev}{19.0}
\renewcommand{\PiWGScore}{19.9}
\renewcommand{\PiWGSed}{18.9}
\renewcommand{\PiWQBase}{15.4}
\renewcommand{\PiWQScore}{14.5}
\renewcommand{\PiYGAcc}{23.1}
\renewcommand{\PiYGBase}{30.0}
\renewcommand{\PiYGRev}{23.8}
\renewcommand{\PiYGScore}{24.3}
\renewcommand{\PiYGSed}{22.7}
\renewcommand{\PiYQAcc}{14.6}
\renewcommand{\PiYQBase}{16.4}
\renewcommand{\PiYQRev}{15.0}
\renewcommand{\PiYQScore}{14.6}
\renewcommand{\PiYQSed}{13.0}
\renewcommand{\PiYQVer}{14.1}
\renewcommand{\PnSepMeanPP}{25.2}
\renewcommand{\RbGBase}{94.8}
\renewcommand{\RbGRev}{98.0}
\renewcommand{\RbGSed}{98.6}
\renewcommand{\RbGVer}{97.6}
\renewcommand{\RbQBase}{98.7}
\renewcommand{\RngGBase}{57}
\renewcommand{\RngGRivHi}{62}
\renewcommand{\RngGRivLo}{52}
\renewcommand{\RngGSed}{15}
\renewcommand{\RngQBase}{56}
\renewcommand{\RngQRivHi}{51}
\renewcommand{\RngQRivLo}{48}
\renewcommand{\RngQSed}{11}
\renewcommand{\RwGAcc}{3.1}
\renewcommand{\RwGBase}{10.1}
\renewcommand{\RwGRev}{3.6}
\renewcommand{\RwGScore}{5.1}
\renewcommand{\RwGSed}{3.8}
\renewcommand{\RwGVer}{3.5}
\renewcommand{\RwQBase}{3.5}
\renewcommand{\RwQScore}{3.2}
\renewcommand{\RwQVer}{2.8}
\newcommand{\LgGemThreeEightBd}{+3.5}
\newcommand{\LgGemThreeEightBs}{46.9}
\newcommand{\LgGemThreeEightHd}{+12.3}
\newcommand{\LgGemThreeEightHs}{70.7}
\newcommand{\LgGemThreeEightScd}{-12.6}
\newcommand{\LgGemThreeEightScs}{43.1}
\newcommand{\LxGemThreeEightBd}{-17.6}
\newcommand{\LxGemThreeEightBs}{50.6}
\newcommand{\LxGemThreeEightHd}{-0.2}
\newcommand{\LxGemThreeEightHs}{72.8}
\newcommand{\LxGemThreeEightScd}{-24.8}
\newcommand{\LxGemThreeEightScs}{46.0}
\renewcommand{\FmtFailMax}{2}
\newcommand{\FmtScoreGFirstHi}{5.4}
\newcommand{\FmtScoreGFirstLo}{4.8}
\newcommand{\OptTempX}{0.7}
\newcommand{\ValTempX}{0.5}
\newcommand{\OptCtxGemmaTwo}{8{,}192}
\newcommand{\ChalTopP}{0.8}
\renewcommand{\JuryAlphaSubstantive}{0.936}
\renewcommand{\JuryAlpha}{0.789}
\newcommand{\JuryTwoN}{300}
\newcommand{\JuryTwoPerStratum}{75}
\newcommand{\JuryTwoAlpha}{0.93}
\newcommand{\JuryTwoOpusEmpty}{20}
\newcommand{\RpAllLeadLoMin}{13}
\newcommand{\PnSedDSeedLo}{+5.00}
\newcommand{\PnSedDSeedHi}{+6.80}
\newcommand{\XtQGemBase}{52.3}
\newcommand{\XtQGemBaseSd}{2.0}
\newcommand{\XqQGemBaseH}{30.6}
\newcommand{\XqQGemBaseY}{97.0}
\newcommand{\XqQGemBaseM}{63.8}
\newcommand{\XqQGemBaseA}{0.78}
\newcommand{\XtQGemSed}{66.4}
\newcommand{\XtQGemSedSd}{0.3}
\newcommand{\XqQGemSedH}{78.0}
\newcommand{\XqQGemSedY}{73.4}
\newcommand{\XqQGemSedM}{75.7}
\newcommand{\XqQGemSedA}{0.88}
\newcommand{\XtQGemVer}{52.2}
\newcommand{\XtQGemVerSd}{0.4}
\newcommand{\XqQGemVerH}{30.7}
\newcommand{\XqQGemVerY}{97.7}
\newcommand{\XqQGemVerM}{64.2}
\newcommand{\XqQGemVerA}{0.81}
\newcommand{\XtQGemRev}{50.5}
\newcommand{\XtQGemRevSd}{1.1}
\newcommand{\XqQGemRevH}{26.1}
\newcommand{\XqQGemRevY}{97.8}
\newcommand{\XqQGemRevM}{61.9}
\newcommand{\XqQGemRevA}{0.80}
\newcommand{\XtQGemAcc}{51.4}
\newcommand{\XtQGemAccSd}{0.3}
\newcommand{\XqQGemAccH}{28.1}
\newcommand{\XqQGemAccY}{97.7}
\newcommand{\XqQGemAccM}{62.9}
\newcommand{\XqQGemAccA}{0.81}
\newcommand{\XtQGemScore}{51.6}
\newcommand{\XtQGemScoreSd}{2.0}
\newcommand{\XqQGemScoreH}{26.9}
\newcommand{\XqQGemScoreY}{98.4}
\newcommand{\XqQGemScoreM}{62.7}
\newcommand{\XqQGemScoreA}{0.79}
\newcommand{\XtQGemMaj}{50.2}
\newcommand{\XtQGemMajSd}{0.4}
\newcommand{\XqQGemMajH}{21.6}
\newcommand{\XqQGemMajY}{98.9}
\newcommand{\XqQGemMajM}{60.3}
\newcommand{\XqQGemMajA}{0.61}
\newcommand{\XtQGemRefl}{52.7}
\newcommand{\XtQGemReflSd}{2.0}
\newcommand{\XqQGemReflH}{31.2}
\newcommand{\XqQGemReflY}{96.9}
\newcommand{\XqQGemReflM}{64.0}
\newcommand{\XqQGemReflA}{0.78}
\newcommand{\XtQHaiBase}{49.4}
\newcommand{\XtQHaiBaseSd}{0.7}
\newcommand{\XqQHaiBaseH}{23.8}
\newcommand{\XqQHaiBaseY}{90.6}
\newcommand{\XqQHaiBaseM}{57.2}
\newcommand{\XqQHaiBaseA}{0.65}
\newcommand{\XtQHaiSed}{55.5}
\newcommand{\XtQHaiSedSd}{0.5}
\newcommand{\XqQHaiSedH}{62.6}
\newcommand{\XqQHaiSedY}{70.4}
\newcommand{\XqQHaiSedM}{66.5}
\newcommand{\XqQHaiSedA}{0.74}
\newcommand{\XtQHaiVer}{49.3}
\newcommand{\XtQHaiVerSd}{0.3}
\newcommand{\XqQHaiVerH}{24.6}
\newcommand{\XqQHaiVerY}{91.1}
\newcommand{\XqQHaiVerM}{57.8}
\newcommand{\XqQHaiVerA}{0.68}
\newcommand{\XtQHaiRev}{49.0}
\newcommand{\XtQHaiRevSd}{0.2}
\newcommand{\XqQHaiRevH}{22.1}
\newcommand{\XqQHaiRevY}{91.5}
\newcommand{\XqQHaiRevM}{56.8}
\newcommand{\XqQHaiRevA}{0.68}
\newcommand{\XtQHaiAcc}{48.9}
\newcommand{\XtQHaiAccSd}{0.2}
\newcommand{\XqQHaiAccH}{22.6}
\newcommand{\XqQHaiAccY}{90.7}
\newcommand{\XqQHaiAccM}{56.6}
\newcommand{\XqQHaiAccA}{0.67}
\newcommand{\XtQHaiScore}{49.3}
\newcommand{\XtQHaiScoreSd}{0.5}
\newcommand{\XqQHaiScoreH}{24.6}
\newcommand{\XqQHaiScoreY}{91.7}
\newcommand{\XqQHaiScoreM}{58.2}
\newcommand{\XqQHaiScoreA}{0.67}
\newcommand{\XtQHaiMaj}{48.8}
\newcommand{\XtQHaiMajSd}{0.1}
\newcommand{\XqQHaiMajH}{20.4}
\newcommand{\XqQHaiMajY}{93.7}
\newcommand{\XqQHaiMajM}{57.1}
\newcommand{\XqQHaiMajA}{0.58}
\newcommand{\XtQHaiRefl}{49.6}
\newcommand{\XtQHaiReflSd}{0.6}
\newcommand{\XqQHaiReflH}{25.6}
\newcommand{\XqQHaiReflY}{89.5}
\newcommand{\XqQHaiReflM}{57.6}
\newcommand{\XqQHaiReflA}{0.65}
\newcommand{\XtGGemBase}{52.9}
\newcommand{\XtGGemBaseSd}{0.9}
\newcommand{\XqGGemBaseH}{7.2}
\newcommand{\XqGGemBaseY}{98.5}
\newcommand{\XqGGemBaseM}{52.9}
\newcommand{\XqGGemBaseA}{0.58}
\newcommand{\XtGGemSed}{64.3}
\newcommand{\XtGGemSedSd}{1.1}
\newcommand{\XqGGemSedH}{63.1}
\newcommand{\XqGGemSedY}{75.2}
\newcommand{\XqGGemSedM}{69.2}
\newcommand{\XqGGemSedA}{0.81}
\newcommand{\XtGGemVer}{54.0}
\newcommand{\XtGGemVerSd}{0.9}
\newcommand{\XqGGemVerH}{10.7}
\newcommand{\XqGGemVerY}{98.4}
\newcommand{\XqGGemVerM}{54.5}
\newcommand{\XqGGemVerA}{0.64}
\newcommand{\XtGGemRev}{52.3}
\newcommand{\XtGGemRevSd}{1.0}
\newcommand{\XqGGemRevH}{6.2}
\newcommand{\XqGGemRevY}{98.8}
\newcommand{\XqGGemRevM}{52.5}
\newcommand{\XqGGemRevA}{0.60}
\newcommand{\XtGGemAcc}{52.0}
\newcommand{\XtGGemAccSd}{0.5}
\newcommand{\XqGGemAccH}{5.2}
\newcommand{\XqGGemAccY}{98.8}
\newcommand{\XqGGemAccM}{52.0}
\newcommand{\XqGGemAccA}{0.59}
\newcommand{\XtGGemScore}{52.1}
\newcommand{\XtGGemScoreSd}{0.4}
\newcommand{\XqGGemScoreH}{7.9}
\newcommand{\XqGGemScoreY}{98.1}
\newcommand{\XqGGemScoreM}{53.0}
\newcommand{\XqGGemScoreA}{0.62}
\newcommand{\XtGGemMaj}{52.6}
\newcommand{\XtGGemMajSd}{0.1}
\newcommand{\XqGGemMajH}{5.7}
\newcommand{\XqGGemMajY}{99.2}
\newcommand{\XqGGemMajM}{52.4}
\newcommand{\XqGGemMajA}{0.53}
\newcommand{\XtGGemRefl}{48.0}
\newcommand{\XtGGemReflSd}{3.2}
\newcommand{\XqGGemReflH}{10.9}
\newcommand{\XqGGemReflY}{88.9}
\newcommand{\XqGGemReflM}{49.9}
\newcommand{\XqGGemReflA}{0.49}
\newcommand{\XtGHaiBase}{43.2}
\newcommand{\XtGHaiBaseSd}{0.3}
\newcommand{\XqGHaiBaseH}{6.7}
\newcommand{\XqGHaiBaseY}{94.8}
\newcommand{\XqGHaiBaseM}{50.8}
\newcommand{\XqGHaiBaseA}{0.52}
\newcommand{\XtGHaiSed}{49.5}
\newcommand{\XtGHaiSedSd}{0.3}
\newcommand{\XqGHaiSedH}{44.8}
\newcommand{\XqGHaiSedY}{80.3}
\newcommand{\XqGHaiSedM}{62.5}
\newcommand{\XqGHaiSedA}{0.68}
\newcommand{\XtGHaiVer}{44.4}
\newcommand{\XtGHaiVerSd}{0.2}
\newcommand{\XqGHaiVerH}{8.8}
\newcommand{\XqGHaiVerY}{95.1}
\newcommand{\XqGHaiVerM}{51.9}
\newcommand{\XqGHaiVerA}{0.58}
\newcommand{\XtGHaiRev}{44.5}
\newcommand{\XtGHaiRevSd}{0.1}
\newcommand{\XqGHaiRevH}{7.5}
\newcommand{\XqGHaiRevY}{95.9}
\newcommand{\XqGHaiRevM}{51.7}
\newcommand{\XqGHaiRevA}{0.57}
\newcommand{\XtGHaiAcc}{43.6}
\newcommand{\XtGHaiAccSd}{0.2}
\newcommand{\XqGHaiAccH}{6.0}
\newcommand{\XqGHaiAccY}{95.4}
\newcommand{\XqGHaiAccM}{50.7}
\newcommand{\XqGHaiAccA}{0.56}
\newcommand{\XtGHaiScore}{44.6}
\newcommand{\XtGHaiScoreSd}{0.2}
\newcommand{\XqGHaiScoreH}{10.0}
\newcommand{\XqGHaiScoreY}{94.8}
\newcommand{\XqGHaiScoreM}{52.4}
\newcommand{\XqGHaiScoreA}{0.57}
\newcommand{\XtGHaiMaj}{44.4}
\newcommand{\XtGHaiMajSd}{0.1}
\newcommand{\XqGHaiMajH}{4.9}
\newcommand{\XqGHaiMajY}{97.4}
\newcommand{\XqGHaiMajM}{51.2}
\newcommand{\XqGHaiMajA}{0.51}
\newcommand{\XtGHaiRefl}{38.8}
\newcommand{\XtGHaiReflSd}{1.6}
\newcommand{\XqGHaiReflH}{8.6}
\newcommand{\XqGHaiReflY}{83.6}
\newcommand{\XqGHaiReflM}{46.1}
\newcommand{\XqGHaiReflA}{0.38}
\newcommand{\XsGainLo}{6.2}
\newcommand{\XsGainHi}{14.1}
\newcommand{\XsOthGainMax}{1.4}
\newcommand{\XsSedHoldLo}{45}
\newcommand{\XsSedHoldHi}{78}
\newcommand{\XsOthHoldLo}{5}
\newcommand{\XsOthHoldHi}{31}
\newcommand{\XsSedYieldLo}{70}
\newcommand{\XsSedYieldHi}{80}
\newcommand{\XsOthYieldLo}{84}
\newcommand{\XsOthYieldHi}{99}
\newcommand{\XsMeanGapLo}{8}
\newcommand{\XsMeanGapHi}{19}
\newcommand{\XsSedAucLo}{0.68}
\newcommand{\XsSedAucHi}{0.88}
\newcommand{\XsOthAucMax}{0.81}
\newcommand{\XsMajHoldLo}{5}
\newcommand{\XsMajHoldHi}{22}
\newcommand{\XsMajYieldLo}{94}
\newcommand{\XsMajYieldHi}{99}
\newcommand{\XsMajMeanGapMax}{3.5}
\newcommand{\XsReflCostLo}{4}
\newcommand{\XsReflCostHi}{5}
\newcommand{\XsGainCiLoMin}{5}
\newcommand{\XsItems}{38{,}434}
\newcommand{\XsGpqaN}{250}
\newcommand{\EvBiasMax}{16}
\newcommand{\EvMeanGem}{10.4}
\newcommand{\EvMeanGemLo}{9.1}
\newcommand{\EvMeanGemHi}{11.9}
\newcommand{\EvAccGem}{7.5}
\newcommand{\EvMeanHai}{7.0}
\newcommand{\EvMeanHaiLo}{4.7}
\newcommand{\EvMeanHaiHi}{9.5}
\newcommand{\EvAccHai}{4.4}
\newcommand{\RprotoQCapLo}{6.6}
\newcommand{\RprotoQCapHi}{7.3}
\newcommand{\RprotoQSynLo}{4.9}
\newcommand{\RprotoQSynHi}{6.2}
\newcommand{\RprotoGBaseSyn}{5.4}
\newcommand{\RchQGem}{36.2}
\newcommand{\RchQHai}{46.1}
\newcommand{\RchGGem}{40.5}
\newcommand{\RchGHai}{33.1}
\newcommand{\RpQSedH}{89.4}
\newcommand{\RpQSedY}{58.3}
\newcommand{\RpQSedM}{73.8}
\newcommand{\RpQSedL}{2.47}
\newcommand{\RpQSedA}{0.83}
\newcommand{\RpQStubH}{84.1}
\newcommand{\RpQStubY}{64.8}
\newcommand{\RpQStubM}{74.5}
\newcommand{\RpQStubL}{2.29}
\newcommand{\RpQStubA}{0.81}
\newcommand{\RpQJudgeH}{83.2}
\newcommand{\RpQJudgeY}{65.9}
\newcommand{\RpQJudgeM}{74.6}
\newcommand{\RpQJudgeL}{2.27}
\newcommand{\RpQJudgeA}{0.82}
\newcommand{\RpQRevH}{49.7}
\newcommand{\RpQRevY}{86.1}
\newcommand{\RpQRevM}{67.9}
\newcommand{\RpQRevL}{1.82}
\newcommand{\RpQRevA}{0.74}
\newcommand{\RpQScoreH}{49.6}
\newcommand{\RpQScoreY}{88.3}
\newcommand{\RpQScoreM}{68.9}
\newcommand{\RpQScoreL}{2.00}
\newcommand{\RpQScoreA}{0.76}
\newcommand{\RpQAccH}{51.7}
\newcommand{\RpQAccY}{85.1}
\newcommand{\RpQAccM}{68.4}
\newcommand{\RpQAccL}{1.81}
\newcommand{\RpQAccA}{0.75}
\newcommand{\RpQVerH}{51.7}
\newcommand{\RpQVerY}{86.4}
\newcommand{\RpQVerM}{69.0}
\newcommand{\RpQVerL}{1.92}
\newcommand{\RpQVerA}{0.76}
\newcommand{\RpQBaseH}{53.4}
\newcommand{\RpQBaseY}{84.0}
\newcommand{\RpQBaseM}{68.7}
\newcommand{\RpQBaseL}{1.80}
\newcommand{\RpQBaseA}{0.72}
\newcommand{\RpQLaH}{53.5}
\newcommand{\RpQLaY}{83.7}
\newcommand{\RpQLaM}{68.6}
\newcommand{\RpQLbH}{58.4}
\newcommand{\RpQLbY}{81.9}
\newcommand{\RpQLbM}{70.1}
\newcommand{\RpQLcH}{59.5}
\newcommand{\RpQLcY}{81.1}
\newcommand{\RpQLcM}{70.3}
\newcommand{\RpGSedH}{89.4}
\newcommand{\RpGSedY}{63.0}
\newcommand{\RpGSedM}{76.2}
\newcommand{\RpGSedL}{2.73}
\newcommand{\RpGSedA}{0.86}
\newcommand{\RpGAccH}{22.3}
\newcommand{\RpGAccY}{94.6}
\newcommand{\RpGAccM}{58.4}
\newcommand{\RpGAccL}{1.61}
\newcommand{\RpGAccA}{0.63}
\newcommand{\RpGRevH}{25.3}
\newcommand{\RpGRevY}{94.9}
\newcommand{\RpGRevM}{60.1}
\newcommand{\RpGRevL}{1.85}
\newcommand{\RpGRevA}{0.68}
\newcommand{\RpGScoreH}{31.9}
\newcommand{\RpGScoreY}{93.1}
\newcommand{\RpGScoreM}{62.5}
\newcommand{\RpGScoreL}{1.86}
\newcommand{\RpGScoreA}{0.71}
\newcommand{\RpGVerH}{29.7}
\newcommand{\RpGVerY}{93.5}
\newcommand{\RpGVerM}{61.6}
\newcommand{\RpGVerL}{1.81}
\newcommand{\RpGVerA}{0.69}
\newcommand{\RpGBaseH}{24.8}
\newcommand{\RpGBaseY}{93.2}
\newcommand{\RpGBaseM}{59.0}
\newcommand{\RpGBaseL}{1.51}
\newcommand{\RpGBaseA}{0.60}
\newcommand{\RpGLaH}{20.0}
\newcommand{\RpGLaY}{94.4}
\newcommand{\RpGLaM}{57.2}
\newcommand{\RpGLbH}{31.0}
\newcommand{\RpGLbY}{92.5}
\newcommand{\RpGLbM}{61.8}
\newcommand{\RpGLcH}{30.9}
\newcommand{\RpGLcY}{92.2}
\newcommand{\RpGLcM}{61.5}
\newcommand{\RpQSedBaseM}{5.1}
\newcommand{\RpQSedBaseMLo}{2.4}
\newcommand{\RpQSedBaseMHi}{7.7}
\newcommand{\RpQSedBaseL}{0.68}
\newcommand{\RpQSedBaseLLo}{0.40}
\newcommand{\RpQSedBaseLHi}{0.98}
\newcommand{\RpQRivLLo}{0.47}
\newcommand{\RpQRivLHi}{0.66}
\newcommand{\RpGSedBaseM}{17.2}
\newcommand{\RpGSedBaseMLo}{14.8}
\newcommand{\RpGSedBaseMHi}{19.6}
\newcommand{\RpGSedBaseL}{1.23}
\newcommand{\RpGSedBaseLLo}{0.89}
\newcommand{\RpGSedBaseLHi}{1.56}
\newcommand{\RpGRivLLo}{0.88}
\newcommand{\RpGRivLHi}{1.12}
\newcommand{\RpJdH}{6.2}
\newcommand{\RpJdHLo}{5.3}
\newcommand{\RpJdHHi}{7.1}
\newcommand{\RpJdY}{7.6}
\newcommand{\RpJdYLo}{6.3}
\newcommand{\RpJdYHi}{9.0}
\newcommand{\RpJdM}{-0.7}
\newcommand{\RpJdMLo}{-1.5}
\newcommand{\RpJdMHi}{+0.1}
\newcommand{\RpJdL}{0.20}
\newcommand{\RpJdLLo}{0.11}
\newcommand{\RpJdLHi}{0.30}
\newcommand{\RqQMeanLo}{3.5}
\newcommand{\RqQMeanHi}{5.2}
\newcommand{\RqQYieldLo}{81.1}
\newcommand{\RqQYieldHi}{83.7}
\newcommand{\RqGMeanLo}{14.4}
\newcommand{\RqGMeanHi}{19.0}
\newcommand{\RqGYieldLo}{92.2}
\newcommand{\RqGYieldHi}{94.4}
\newcommand{\RsHold}{5.3}
\newcommand{\RsYield}{6.5}
\newcommand{\RsMean}{+0.6}
\newcommand{\RsMeanLo}{-0.3}
\newcommand{\RsMeanHi}{+1.6}
\newcommand{\RsL}{0.18}
\newcommand{\RsLLo}{0.08}
\newcommand{\RsLHi}{0.29}
\newcommand{\RsFloor}{-2.25}
\newcommand{\EvQMeanGem}{17.7}
\newcommand{\EvQMeanGemLo}{15.3}
\newcommand{\EvQMeanGemHi}{18.7}
\newcommand{\EvQMeanHai}{9.9}
\newcommand{\EvQMeanHaiLo}{7.6}
\newcommand{\EvQMeanHaiHi}{11.9}
\newcommand{\EvQAccGem}{12.4}
\newcommand{\EvQAccHai}{8.3}
\newcommand{\PaBaseCT}{47.1}
\newcommand{\PaBaseCTsd}{1.1}
\newcommand{\PaBaseCap}{4.4}
\newcommand{\PaBaseCapsd}{1.2}
\newcommand{\PaBaseCS}{53.8}
\newcommand{\PaBaseCSsd}{2.1}
\newcommand{\PaPbtCT}{43.3}
\newcommand{\PaPbtCTsd}{0.4}
\newcommand{\PaPbtCap}{4.2}
\newcommand{\PaPbtCapsd}{1.2}
\newcommand{\PaPbtCS}{51.4}
\newcommand{\PaPbtCSsd}{2.5}
\newcommand{\PaSedCT}{59.2}
\newcommand{\PaSedCTsd}{1.1}
\newcommand{\PaSedCap}{2.2}
\newcommand{\PaSedCapsd}{0.7}
\newcommand{\PaSedCS}{58.5}
\newcommand{\PaSedCSsd}{1.6}
\newcommand{\PbBaseH}{23.6}
\newcommand{\PbBaseY}{84.2}
\newcommand{\PbBaseM}{53.9}
\newcommand{\PbBaseA}{0.581}
\newcommand{\PbBaseCS}{45.7}
\newcommand{\PbPbtH}{38.7}
\newcommand{\PbPbtY}{69.5}
\newcommand{\PbPbtM}{54.1}
\newcommand{\PbPbtA}{0.555}
\newcommand{\PbPbtCS}{46.9}
\newcommand{\PbSedH}{56.4}
\newcommand{\PbSedY}{63.2}
\newcommand{\PbSedM}{59.8}
\newcommand{\PbSedA}{0.695}
\newcommand{\PbSedCS}{54.4}
\newcommand{\PcSedPbtM}{5.7}
\newcommand{\PcSedPbtMLo}{2.7}
\newcommand{\PcSedPbtMHi}{8.6}
\newcommand{\PcSedPbtA}{0.140}
\newcommand{\PcSedPbtALo}{0.085}
\newcommand{\PcSedPbtAHi}{0.200}
\newcommand{\PcSedBaseM}{5.9}
\newcommand{\PcSedBaseMLo}{3.5}
\newcommand{\PcSedBaseMHi}{8.4}
\newcommand{\PcSedBaseA}{0.114}
\newcommand{\PcSedBaseALo}{0.060}
\newcommand{\PcSedBaseAHi}{0.166}
\newcommand{\PcPbtBaseH}{15.1}
\newcommand{\PcPbtBaseY}{14.7}
\newcommand{\PcPbtBaseM}{+0.2}
\newcommand{\PcPbtBaseMLo}{-2.6}
\newcommand{\PcPbtBaseMHi}{+2.9}
\newcommand{\RfiveTok}{4{,}096}
\newcommand{\RsLambda}{2}
\newcommand{\EvBiasWideA}{32}
\newcommand{\EvBiasWideB}{64}
\newcommand{\FigOneMaxGapQ}{45}
\newcommand{\FigOneMaxGapG}{67}
\newcommand{\GnAccGain}{3.2}
\newcommand{\MgGScore}{71.1}
\renewcommand{\PnSepMeanPP}{25.3}
\renewcommand{\GnSepMeanPP}{36.1}
\renewcommand{\SplitEpisodes}{256{,}000}
\renewcommand{\SplitCTonePct}{43}
\renewcommand{\SplitCTzeroPct}{38}

\newcommand{\ce}[2]{#1{\tiny$\pm$#2}}

\newcommand{\ceb}[2]{\textbf{#1}{\tiny$\pm$#2}}

\title{Benchmarks Don't Talk Back:\\
       Learning Through Self-Disagreement}

\author{Soheil Abbasloo, Peng Cheng \\
Microsoft Research
}
\iclrfinalcopy

\begin{document}
\maketitle
\lhead{Preprint}

\begin{abstract}
A benchmark usually measures a language model alone with a question, and deployment rarely
looks like that: answers get contested, and models give up correct ones under
counter-reasoning that is itself wrong. We show that what fails there is not a shortfall of
accuracy but a capability
in its own right --- knowing when to hold a position and when to yield to a better one ---
which we call \emph{dialectical competence}.
It stands on first-answer accuracy without following from it: a
model can be made materially better at answering and no better at this, and training for
accuracy alone can drift against it. Nor can it be bought at inference, by self-reflection or by
sampling more. Taking a cue from the dialectical tradition in philosophy, in which productive
opposition arises within a position rather than being imposed on it from outside, we
hypothesize that this capability can be trained from opposition the model supplies itself. It
states a position, generates the strongest case against that position --- its
\emph{antithesis} --- rules on whether the case stands, and is trained on that ruling, with
no second model at training or at deployment. We call this learning through
\emph{Self-Disagreement}: reinforcement learning from verifiable rewards throughout, where
the same check that scores an answer against its reference scores the ruling too, so no
reward model, no preference data and no human annotation enter anywhere. The capability is
trained rather than elicited: weak in the base model, it arrives with the reward, is not
paid for out of first-answer accuracy, and holds against challengers the model never trained
against, including ones stronger than itself. The opposition a model needs in order to
improve, it turns out, need not come from outside it.
\end{abstract}

\section{Introduction}
\label{sec:intro}

\noindent \textbf{Benchmarks do not talk back:} Language models have become
markedly better at solving problems across successive open frontier
families~\citep{grattafiori2024llama3,qwen3technical}, with reinforcement
learning against answers a program can check providing one route to that
improvement~\citep{guo2025deepseekr1,shao2024deepseekmath,yu2025dapo,liu2025drgrpo}.
Benchmark suites register that progress across general knowledge~\citep{mmlu},
graduate-level science~\citep{rein2024gpqa}, competition
mathematics~\citep{hendrycks2021math,omnimath} and
code~\citep{jain2024livecodebench}. Yet each ultimately reduces evaluation to a
question, a model's answer, and a score on that answer. Deployment is often
different: users push back on assistants; inside agentic loops another process
may propose a different answer and contest a draft before it is
accepted~\citep{qian2023chatdev,hong2023metagpt,wu2023autogen}; and later turns
may ask a model to reconsider an answer it has already given. Disagreement of
that kind is not hostile and not an edge case; it is the normal texture of use.
Answers, however, move under it: models accommodate a user's stated view rather
than the evidence~\citep{sharma2023sycophancy}, give up correct answers when merely doubted~\citep{laban2023flipflop}, and abandon answers that
were right under sustained counter-reasoning~\citep{wang2023defend}, at rates that do not fall as the
model gets stronger~\citep{abbasloo2025siev}.

\begin{wrapfigure}{R}{0.50\textwidth}
\centering
\includegraphics[width=\linewidth]{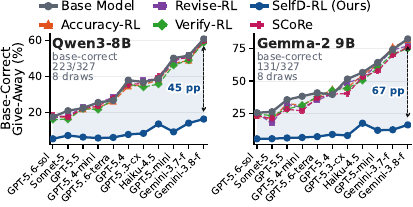}\caption{\textbf{Answering better does not buy the capability.} Share of the base model's correct
answers each arm gives up under challenge, one position per challenger; each trained arm is the mean of three training seeds. Each of the three
ordinarily trained arms answers better on its own and still tracks the base model here, as does SCoRe (\citealp{kumar2024score}; App.~\ref{sec:scoreapp});
only our design, Self-Disagreement (SelfD-RL), separates from them --- by \PnSepMeanPP\,pp on Qwen3-8B and \GnSepMeanPP\,pp
on Gemma-2 9B on average across these eleven frontier challengers (up to \FigOneMaxGapQ\ and \FigOneMaxGapG\,pp from the base model; more in Table~\ref{tab:ladder}).}
\label{fig:pressure}
\end{wrapfigure}

\noindent \textbf{A capability in its own right:} What moves under challenge is
not another shortfall in answering, because the failure begins after a correct
answer has already been produced. What is missing is the ability to weigh a case
made against a position and to adjust only when the case deserves it --- holding
when the challenge is incorrect, yielding when it is correct --- which we call
\emph{dialectical competence}. Training for correctness does not supply it, though
correctness is its precondition: a model that cannot reach the right answer has
nothing to hold. It is the higher skill, standing on answering rather than
competing with it. It is
also the capability hard to supply from outside, because every way of
correcting a model that leans on a stronger judge --- a larger model, a separate
verifier, a person --- needs that judge to exist, and one better than the model
being corrected is rare and costly.
To separate the two, we train on two base models drawn from different families,
Qwen3-8B~\citep{qwen3technical} and Gemma-2 9B~\citep{gemma2}, and run the
same comparison on each.
Figure~\ref{fig:pressure} places the base model beside three ordinarily
trained arms (each defined in \S\ref{sec:repro}), against the same challengers: each answers better
on its own, and each still gives up its correct answers under challenge at close
to the base model's rate. Training against the model's own opposition
separates from that pattern, on both families (the diagnosis is made precise in
\S\ref{sec:motivation}).

\noindent \textbf{Our proposal --- learning through Self-Disagreement:} The
remedies that suggest themselves share one defect: every one of them works on the
answer, when the capability at issue is the further one that begins after the
answer. Training harder for correctness,
giving the model a second pass over its own answer, and training it to judge that answer
before committing all track the base model rather than separating from it
(Figure~\ref{fig:pressure}); buying more inference --- a reconsideration pass, or
sampling many answers and voting --- does not reach it either (\S\ref{sec:eval});
Appendix~\ref{sec:relwork} discusses these lines of work at greater length. In
each, the move the model makes is prompted, sampled or borrowed; it never follows from
the model's own ruling on a case it was made to confront. The dialectical
tradition puts exactly that ruling at the center: a position is developed by
meeting the strongest case against
it~\citep{plato_republic,popper_logic_scientific_discovery}; that case is raised
from within the position rather than handed to it from
above~\citep{hegel_phenomenology_spirit}; and the position must rule on the case
before it may move.
We take this as a training paradigm rather than as a metaphor. A single policy
plays both parts, differing only in what each is shown. It states a position, the
\emph{thesis}; in a context of its own, holding the question and the case just
made but never the reference answer, it argues against that position, the
\emph{antithesis}; it then returns to the answering role, rules on whether that
case stands, and commits to a final answer, the \emph{synthesis}
(\S\ref{sec:method}). Only the answering turns are rewarded, and the reward is
the verifier we already have: the same check that scores an answer against its
reference also scores the answer the model settles on once opposed.
No reward model, no preference data, no second model at training or
deployment: \emph{the opposition a model needs, it supplies itself}. We contribute:

\begin{enumerate}[label=(\arabic*), leftmargin=1.5em, itemsep=0pt, topsep=1pt, parsep=0pt]
\item \emph{Dialectical competence as a capability in its own right}: we measure it on a
probe built for it, and show that neither accuracy-centered training nor inference-time
remedies supply it (\S\ref{sec:motivation}).
\item \emph{Self-Disagreement}: a single-policy extension of RLVR, with the
opposition drawn from the policy being trained (\S\ref{sec:method}).
\item \emph{Demonstrating that Self-Disagreement improves it}: trained on two
base families, the models gain dialectical competence without paying for it in
first-answer accuracy (\S\ref{sec:overall}), and keep the gain against challengers
they never trained against, with an account of where it comes from
(\S\ref{sec:mechanism}).
\end{enumerate}

{}

\section{The problem}
\label{sec:motivation}

\noindent \textbf{A probe that makes holding and yielding visible:} Benchmarks record
whether a model ended up right, not what it did when it was pushed, so we measure
on a probe built for the second question. We call it the \emph{disagreement
probe}, or \emph{D-probe}, and it is used for every measurement of behavior
under challenge in this
paper. It is \ProbeItems\ questions, each put
to every one of \DcNrungs\ challengers --- \DcExtWord\ external models spanning open
checkpoints to frontier systems, the model's own frozen starting point, and its
own critique. The
diagnosis below runs it on the two base models, Qwen3-8B and Gemma-2 9B. Every
model answers once and that answer is replayed to each challenger, so only the
answer given after the challenge can differ, and all three answers are graded:
the model's own, the challenger's, and the one the model settles on. Knowing
which of the first two was right is what makes that choice legible, because it
says what the right action was. What we report is the accuracy of that settled
answer, the \emph{synthesis accuracy}.

\noindent \textbf{The label is checked, not assumed:} Synthesis accuracy is what
the metric rests on, so it was checked a second
way: on the items where a base model gives up a correct answer, a blinded jury of
Claude-Opus-4.7, GPT-5.5 and Gemini-3.1-pro-preview reads the reasoning and judges
all \JuryCaveN\ such items to be non-genuine --- conceding and adopting the
challenger's answer with no valid pushback, or merely restating it --- rather than
a differently argued answer. What synthesis accuracy loses is
therefore reasoning under challenge, not a relabelled answer
(\S\ref{sec:juryapp}).

\noindent \textbf{Both base models are one-sided:} What their synthesis accuracy
is made of, rather than its level, is the diagnosis. On the items where the model
was right and the challenger wrong --- where holding is the correct action ---
Qwen3-8B ends on the correct answer \MhQBase\,\% of the time and Gemma-2 9B
\MhGBase\,\%. On the items where the challenger was right and the model wrong,
where yielding is correct, the same two models score \MyQBase\,\% and
\MyGBase\,\%. That pairing is not discrimination between good arguments and bad;
it is close to unconditional surrender, and one behavior produces both
numbers. The cost rises with the challenger: across Figure~\ref{fig:pressure}'s
challengers the base model gives up \DegPlaneGiveLo\ to \DegPlaneGiveHi\ of its correct
answers on Qwen3-8B and \GnDegPlaneGiveLo\ to \GnDegPlaneGiveHi\ on Gemma-2 9B.

\begin{wrapfigure}{R}{0.50\textwidth}
\centering
\includegraphics[width=\linewidth]{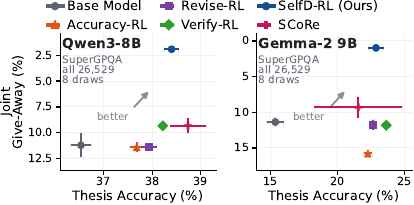}\caption{\textbf{Buying accuracy is not buying the capability.} SuperGPQA: each arm sits at its
thesis accuracy ($x$) against how often it gave up a correct answer under challenge
($y$, inverted), so better is up and to the right. Bars are $\pm1$\,sd across three training seeds (for Base, over its \PnDecodes\ sampled answers).}
\label{fig:bench}
\end{wrapfigure}

\noindent \textbf{Training for accuracy does not reach it:} The natural reading
of that is that the models are not yet good enough, and that better answering
will carry the rest. It does not. A control trained for accuracy alone
(Accuracy-RL; defined, with its settings, in \S\ref{sec:repro}), from the same base model and the
same data, also improves the first answer and buys none of the behavior under challenge, which can drift against it over training rather than
holding steady (Appendix~\ref{app:rewardablation}). Figure~\ref{fig:bench} shows it on
SuperGPQA~\citep{supergpqa}: the trained alternatives, the published self-correction method SCoRe~\citep{kumar2024score} among them, move right, to higher thesis
accuracy, and stay at the same height. Answering well is the floor, not the
capability: a model that cannot produce the right answer has none to hold, but past
that floor more of it buys nothing here. Nor is the
opposition's content incidental: an opposition that carries no content recovers only part of the gain (\S\ref{sec:mechanism}).

\noindent \textbf{What that leaves:} The failure is not a shortfall that better
answering removes, and it is not repaired by the inference-time remedies we test, self-consistency~\citep{wang2022selfcons} and self-reflection~\citep{madaan2023selfrefine},
or by training the same objective harder. What is missing is the decision itself:
telling a correct challenge from an incorrect one and acting on the
difference. The rest of this paper is about training that.

\section{Design}
\label{sec:method}
{}

\noindent\textbf{Overview:} We train a single policy under a dialectical
protocol we call \emph{Self-Disagreement}. The policy commits to an answer, argues
against that answer in a separate context, then returns to rule on the argument
before committing again. Only the answering turns are trained. The protocol
extends RLVR rather than departing from it: the verifier that scores an answer
against its reference also scores the ruling, and nothing else in the training
loop changes.

\noindent\textbf{Where the opposition comes from:} Learning to adjudicate
disagreement presupposes a source of disagreement. The usual construction takes it
from a second model, which makes the training signal a property of the
\emph{pair}: an opposition that is too weak can be dismissed without
discrimination, and one that is too strong makes deference the best available
response. The policy's competence also changes as it trains, so a pairing
calibrated at the start does not stay calibrated, and keeping it so means
tracking a quantity that is not directly observable. Drawing the opposition from
the policy itself removes that coupling. Its competence tracks the policy's by
construction and without scheduling, every training item has an opposition
available without a second model to serve it, and what is deployed is a single
set of weights. The construction does rest on an assumption worth stating: that
a policy can be induced to argue against a position it has just committed to.

\begin{figure}[t]
\begin{minipage}[t]{0.53\textwidth}
\vspace{0pt}
\footnotesize
\captionof{algorithm}{Self-Disagreement, one training step}
\label{alg:sed}
\vspace{2pt}\hrule height \lightrulewidth\vspace{3pt}
\begin{algorithmic}[1]
\Require policy $\pi_\theta$, RLVR verifier $\mathcal{V}(\cdot\,;y^\star)$, group size
         $G$, admission band $[\underline{p},\overline{p}]$
\Statex\vspace{-2.5mm}\rule{\linewidth}{\lightrulewidth}\vspace{0.5mm}
\State sample a batch of questions $q$ with $p_0(q) \in [\underline{p},\overline{p}]$
\ForAll{$q$ in the batch}
  \For{$i = 1$ to $G$}                                \Comment{one GRPO group}
    \State $T \sim \pi_\theta(\cdot \mid q)$                      \Comment{answering role}
    \State $A \sim \pi_\theta(\cdot \mid q, T)$                   \Comment{opposing role; no $y^\star$}
    \State $S \sim \pi_\theta(\cdot \mid q, T, A)$                \Comment{resumed: rule, then answer}
    \State $c_T \gets \mathcal{V}(T;y^\star)$;\quad
           $c_S \gets \mathcal{V}(S;y^\star)$                 \Comment{the answers}
    \Statex \hskip\algorithmicindent\hskip\algorithmicindent
           $c_v \gets \mathbf{1}[v = \mathcal{V}(A;y^\star)]$ \Comment{and the ruling}
    \State $\beta \gets \RewardRF$ if $c_T = 0 \wedge \mathcal{V}(A;y^\star) = 0$, else $1$
    \State $R(\tau_i) \gets \beta c_S + w_v c_v + w_T c_T$
    \Statex \hskip\algorithmicindent\hskip\algorithmicindent
           $\phantom{R(\tau_i)} \mathrel{-} \lambda(\mathrm{cap} + \mathrm{stub}) - w_\phi(\phi_T + \phi_v)$
    \If{$\phi_T \vee \phi_v$} $R(\tau_i) \gets R_{\mathrm{floor}}$ \Comment{override}
    \EndIf
  \EndFor
  \State $\hat{A}_i \gets \big(R(\tau_i) - \mu_q\big)/\sigma_q$ \Comment{group-relative}
\EndFor
\State update $\theta$ on the clipped surrogate, KL-penalized in the return
       \Comment{stock GRPO}
\end{algorithmic}
\vspace{1pt}\hrule height \heavyrulewidth
\end{minipage}\hfill
\begin{minipage}[t]{0.44\textwidth}
\vspace{0pt}
\centering
\scalebox{0.85}{\begin{tikzpicture}[
  font=\footnotesize,
  band/.style={draw, rounded corners=2pt, text width=48mm, align=center,
               inner xsep=3pt, inner ysep=2.5pt},
  ans/.style={band, fill=blue!8, draw=blue!45},
  opp/.style={band, fill=orange!12, draw=orange!60!black},
  ar/.style={-{Latex[length=1.8mm]}, semithick},
  lbl/.style={font=\footnotesize, inner sep=1pt},
  tag/.style={font=\scriptsize\itshape, inner sep=1pt},
]
\node[lbl, text=black!60] (hdr) {one policy $\pi_\theta$, two roles};
\node[ans, below=1mm of hdr] (T) {{\footnotesize\itshape\textcolor{blue!45!black}{answering role \textbullet\ rewarded}}\\[1pt]
  $q \;\rightarrow\;$ thesis $T$};
\node[opp, below=3mm of T] (A) {{\footnotesize\itshape\textcolor{orange!60!black}{opposing role \textbullet\ inference-only}}\\[1pt]
  antithesis $A$\\[-1pt]
  {\footnotesize own context, no $y^\star$, not a target}};
\node[ans, below=3mm of A] (S) {{\footnotesize\itshape\textcolor{blue!45!black}{answering role resumed \textbullet\ rewarded}}\\[1pt]
  verdict $\;\rightarrow\;$ synthesis $S$};

\draw[ar] (T) -- node[lbl, right=1.5mm] {$q,T$} (A);
\draw[ar] (A) -- node[lbl, right=1.5mm] {$q,T,A$} (S);

\node[band, fill=black!5, draw=black!45, below=3mm of S, font=\footnotesize] (R)
  {return $R(\tau)$\\ holding, yielding, correction};
\draw[ar] (S) -- (R);
\draw[ar, dashed, draw=black!55] (R.east) -- ([xshift=2mm]R.east)
  -- node[font=\scriptsize, fill=white, inner sep=0.8pt, pos=0.40, rotate=90] {update}
     ([xshift=2mm]T.east) -- (T.east);
\end{tikzpicture}}
\captionof{figure}{\textbf{Self-Disagreement.} Two roles of one policy, in separate
contexts. The opposing turn is sampled from the same weights tracking the
policy as it trains. It conditions the synthesis, but is never itself a
training target, so it reaches the answering role as given, the way a user's
disagreement would. The
answering role resumes and must rule before it may commit. Settings in
\S\ref{sec:eval}.}
\label{fig:loop}
\end{minipage}
\end{figure}

\noindent\textbf{The dialectical exchange:} Figure~\ref{fig:loop} shows one
exchange and Algorithm~\ref{alg:sed} one training step; the prompt templates for all
three turns are printed verbatim in \S\ref{sec:prompts}. The policy is prompted
into two roles that share its weights and differ only in what each is shown. In
the answering role it writes the \emph{thesis}, and later the \emph{synthesis};
in the opposing role it writes the \emph{antithesis}, disputing the strategy, the
decisions and the reasoning of that particular answer \emph{even where they
appear correct} rather than making an independent second attempt at the question.
Opposition is therefore unconditional --- a correct thesis is argued against as
readily as an incorrect one --- which is what makes \emph{holding} a position
something the policy must learn, since a policy opposed only when it was wrong
would never need to tell a sound challenge from an unsound one. Let $\pi_\theta$ be the policy and let
$q \sim \mathcal{D}$ be a question with reference answer $y^\star$. One dialectical
exchange is a trajectory $\tau = (T, A, S)$ of three generations from the \emph{same}
parameters under two role prompts --- answering, then opposing, then answering
again under instruction to judge:
{\small\begin{equation}
\pi_\theta(\tau \mid q) \;=\;
\underbrace{\pi_\theta(T \mid q)}_{\text{thesis}}\;
\underbrace{\pi_\theta(A \mid q, T)}_{\text{antithesis (sampled, inference-only)}}\;
\underbrace{\pi_\theta(S \mid q, T, A)}_{\text{synthesis}} ,
\label{eq:factorization}
\end{equation}}The synthesis $S$ is internally ordered: it opens with a verdict
$v \in \{0,1\}$ on $A$'s answer and only then commits to an answer. Because the
answer is decoded after the verdict and conditioned on it, the policy must
adjudicate before it may answer, and $v$ is separately verifiable; ablating the
verdict removes that opening block, leaving a synthesis that answers with no
obligation to rule. Writing $\mathcal{V}(\cdot\,;y^\star) \in \{0,1\}$ for the
ordinary RLVR verifier --- exact match against the reference answer, so that no
learned reward model and no preference data enter anywhere --- the trajectory
induces $c_T = \mathcal{V}(T;y^\star)$, $c_S = \mathcal{V}(S;y^\star)$, and
{\small\begin{equation}
c_v \;=\; \mathbf{1}\!\left[\, v = \mathcal{V}(A;y^\star) \,\right],
\end{equation}}which scores whether the policy judged the opposition rightly, not whether the
opposition was right. This is where the verifier's reach is extended: the same
check, applied to a decision about the opposition's answer rather than to an answer.
The credit is available on every item, including those
in which the opposition concedes, so the objective by itself does not rule out a
policy that agrees whenever the opposition agrees with it. It is more than that:
removing the content of the disagreement, together with the supervision that depends on that content, costs roughly half to two thirds of the gain, and the
trained policy keeps its advantage against challengers it never trained against
(\S\ref{sec:oppgenuine}).
The antithesis is sampled from the policy but is \emph{inference-only}: its
trajectory is never collected into the training episode, so
$\pi_\theta(A \mid q,T)$ is a sampling distribution rather than an optimized
factor, and opposing is never rewarded --- only answering and adjudicating are
--- which removes any route by which it could learn to soften its
challenges. No factor of Eq.~(\ref{eq:factorization}) is conditioned on
$y^\star$: the verifier is applied only after the trajectory is complete, so it
cannot oppose selectively on items it is about to answer incorrectly.
Thesis and synthesis are optimized as one
trajectory, not as separate turns, so credit assignment reaches the answer
and the ruling on its challenge jointly.

\noindent\textbf{Guarding against collapse:} The characteristic failure of an
endogenous construction is trivial agreement, in which $A$ concedes $T$ and the
dialectic degenerates to a single pass. Two provisions address this, neither of
them part of the objective. The first is uninformedness, above: the opposing role
cannot tell which items it would be safe to concede. The second is the admission
band --- items enter training only when the base policy's pass rate
$p_0(q) = \Pr[\mathcal{V}(T;y^\star) = 1 \mid q; \pi_{\theta_0}]$ lies in
$[\underline{p}, \overline{p}]$, which excludes items on which opposition is
respectively vacuous or futile.
What the opposition and the verdict actually do in training is then an empirical
question rather than a
guaranteed one, and we treat it as such: Appendix~\ref{sec:oppgenuine} measures
the opposition the policy actually produces, and \S\ref{sec:eval} measures the
resulting behavior rather than relying on the construction.

\noindent\textbf{At inference:} None of the training machinery is required to
use the result. The trained policy is an ordinary model: it answers a question
in one pass, and what training has changed is how it behaves \emph{when it is
disagreed with}. The evaluation of \S\ref{sec:eval} therefore presents the
trained policy with an opposing view and measures the answer it commits to, 
the same situation the protocol trains, with none of the environment's
apparatus. There is no admission band, no candidate selection, no verifier and
no reference answer, and, since the opposing role is the policy itself, no
second model to deploy.

\noindent\textbf{The objective:} Two failures are penalized directly:
{\small\begin{equation}
\underbrace{\mathbf{1}[\,c_T = 1,\; c_S = 0\,]}_{\mathrm{cap}\;\text{(capitulation)}}
\qquad
\underbrace{\mathbf{1}[\,c_T = 0,\; \mathcal{V}(A;y^\star) = 1,\; c_S = 0\,]}_{\mathrm{stub}\;\text{(stubbornness)}}
\end{equation}}Capitulation is a correct thesis surrendered; stubbornness is a valid challenge
refused. \emph{Synthesis accuracy}, $\Pr(c_S{=}1)$, is the outcome variable of every
result in this paper. The return, for the configuration reported throughout, is
{\small\begin{equation}
R(\tau) \;=\; \beta\, c_S \;+\; w_v c_v \;+\; w_T c_T
\;-\; \lambda_{\mathrm{cap}}\,\mathrm{cap} \;-\; \lambda_{\mathrm{stub}}\,\mathrm{stub}
\;-\; w_\phi \big(\phi_T + \phi_v\big),
\label{eq:reward}
\end{equation}}subject to a hard format gate: if either the thesis or the verdict fails to
parse ($\phi_T \vee \phi_v$), the return is \emph{replaced} by a floor value and
every positive channel is discarded,
{\small\begin{equation}
R(\tau) \;\leftarrow\; R_{\mathrm{floor}} \quad \text{if } \phi_T \vee \phi_v .
\end{equation}}Here $\lambda_{\mathrm{cap}} = \lambda_{\mathrm{stub}} = \RewardDual$,
$w_T = 1$, $w_v = 0.4$, $w_\phi = 0.1$ and $R_{\mathrm{floor}} = -0.75$. The objective keeps paying for a
correct first answer at full weight while adding the ruling at $0.4$: the
capability is trained on top of answering, not in place of it. The gate is worth more than its
weight suggests: the policy cannot decline to adjudicate. The remaining
coefficient, $\beta$, multiplies the synthesis credit rather than adding to it:
$\beta = \RewardRF$ when $c_T = 0$ and $\mathcal{V}(A;y^\star) = 0$, and $1$
otherwise, since that is the one cell where no correct position exists to defer
to and the answer has to be reached rather than chosen.
The symmetry of $\lambda_{\mathrm{cap}}$ and $\lambda_{\mathrm{stub}}$ is
load-bearing. An objective on terminal correctness alone is invariant to
the trajectory by which correctness is reached, including trajectories that
discard a correct answer and recover it --- which \S\ref{sec:eval} shows
accuracy-only training does not merely tolerate but increasingly takes.
Penalizing capitulation alone would buy stability at the cost of
revisability.
Optimization is by GRPO \citep{shao2024deepseekmath}: for each $q$ a group of
$G$ trajectories is sampled and scored by its group-relative advantage
$\hat{A}_i = (R(\tau_i) - \mu_q)/\sigma_q$, maximized under a clipped
surrogate \citep{schulman2017ppo}, with the KL penalty to the reference policy
applied to the return rather than to the loss. The algorithm is unmodified: the dialectic changes what is scored, not
how the update is computed.

\section{Evaluation}
\label{sec:eval}
{}

\noindent This section has three parts. The first asks whether the capability is
there at all, and whether available approaches produce it: it is weak in
the base model, arrives with training, holds on benchmarks the training never
saw, and no rival design or test-time scheme we compare reaches it. The second asks where the
gain comes from: whether the model answers better on the first attempt or
decides better once challenged, and which
part of the design buys it. The third asks whether the result is valid and generalizes: a second reading of
the metric by a blinded jury of three frontier models, a different base model, and challengers it never
trained against.

\noindent\textbf{Setup:} Every trained arm is reinforcement learning from
verifiable rewards on the same data, from the same initialization and with the same
verifier; what separates them is the return and the number of
turns it is computed over. Two further arms are not trained at all: the test-time schemes,
which spend additional inference on the base model instead. Every arm, the base model included, meets the challenge under the same instruction, which asks for a verdict on it before the final answer; without that request the other arms do no better, and on Qwen3-8B worse (App.~\ref{sec:prompts}). Because the opposition is
sampled rather than fixed we resample it over independent draws (eight; four in Table~\ref{tab:mechanism}); in the main comparisons each trained arm is the mean of three training seeds and its spread is across seeds, the base model's across its draws. Capitulation and
stubbornness, defined in \S\ref{sec:method}, are counted over every scored item
rather than over the subset on which each could occur, so that they add
to synthesis accuracy and to each other. Behavior under challenge is measured on the D-probe
throughout; the benchmark panel asks whether what it measures survives away from
it. What each arm is, the model, data, hardware and optimizer
settings, the step at which each arm is read, and the construction of the
disagreement probe are in \S\ref{sec:repro}.
{}
{}

{}

\subsection{Overall results and analysis}
\label{sec:overall}

\begin{wrapfigure}{R}{0.5\textwidth}
\centering
\includegraphics[width=\linewidth]{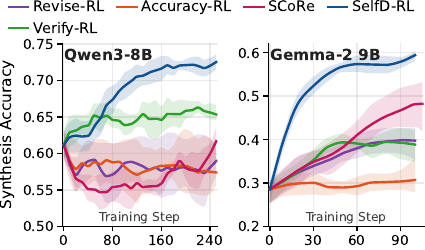}
\caption{\textbf{Dialectical competence is trained under Self-Disagreement, while the arms
that drop the disagreement recover at most part of it.} Bands
across seeds. Separate models:
read the shapes, not the cells.
{}
}
\label{fig:curve}
\end{wrapfigure}

\noindent\textbf{Does it work, and could anything already available have done it:} The capability is not present in either
base model and it arrives with training in both (Figure~\ref{fig:curve}). The single-pass
accuracy-trained arm (Accuracy-RL) is flat over its own run, so more training compute does not
close the gap; nor does spending more at inference
(the test-time columns of Table~\ref{tab:panelbench}).
Table~\ref{tab:panelbench} then asks whether what was learned on the
probe survives away from it.

\noindent\textbf{It survives off the training probe, on both base
models:} On Qwen3-8B the trained model is ahead of the base model on all
\PnNbench\ public benchmarks, on mathematics and code as well as on the
multiple-choice sets where a challenge is cheapest to issue, and on average it
settles on the right answer after being challenged \PnSedD\ pp more often
(95\,\% CI over the \PnNbench\ benchmarks $[\PnSedDLo, \PnSedDHi]$; over the three training seeds $[\PnSedDSeedLo, \PnSedDSeedHi]$), \PnBaseMacro\pct\ to \PnSedMacro\pct.
The final row of
the upper block of Table~\ref{tab:panelbench} repeats the whole exercise on a second base model,
Gemma-2 9B, and the pattern holds: \GnBaseMacro\pct\ to \GnSedMacro\pct\ on
average, ahead of its own base on every benchmark
(\S\ref{sec:gemmapanel}). Every column of the upper block is read under
the same protocol and scored by the same verifier, so what separates the columns
is what each was trained to do and not how each was measured.

\noindent\textbf{How it works with external challengers:} The upper block of Table~\ref{tab:panelbench} lets each arm meet its own antithesis, which is how the model is used; the lower block holds the challenge fixed, giving every arm the same first answer and the same antithesis from Gemini-3.8-flash or Claude-Haiku-4.5, so that only the decision is the arm's own (App.~\ref{sec:sharedapp}). The lead survives, \XsGainLo\ to \XsGainHi\ points over the base model with every 95\,\% interval over training seeds above \XsGainCiLoMin, while no other arm, trained on another objective or given more inference, gains more than \XsOthGainMax\ points on it, and the reason lies in how each responds to the same challenge: the other arms take most antitheses put to them, right or wrong, much as the base model does, whereas Self-Disagreement holds a right first answer against a wrong antithesis more than twice as often as any of them and still takes most of the right ones. Nor does it win by switching less: where the base model, its final answer biased toward its first, switches as rarely as Self-Disagreement, it still trails it by \EvAccGem\ and \EvAccHai\ points of synthesis accuracy on \mbox{Gemma-2~9B}, one per outside challenger (App.~\ref{sec:sharedapp}), so what training on the challenge adds is which items the model holds on, not how many, which is what telling a correct challenge from an incorrect one looks like once the challenge itself is the same for everyone.

\begin{table}[t]
\centering\scriptsize
\caption{\textbf{Synthesis accuracy under challenge, per benchmark (\%).}
\emph{Vf-RL}, \emph{Rv-RL} and \emph{Ac-RL} are Verify-RL, Revise-RL and
Accuracy-RL.
\emph{SCoRe} is the method of \citet{kumar2024score}, detailed in App.~\ref{sec:scoreapp}. \emph{Maj@\TTCCallsMaj} and \emph{Self-Refl}: the base model given more inference
--- test-time self-consistency and test-time self-reflection~(details in App.~\ref{sec:ttcapp}).
$^{\ddagger}$Per-benchmark detail in Table~\ref{tab:gemmapanel}. Mean\,$\pm$\,sd over
three training seeds, each the mean of \PnDecodes\ sampled answers; for Base and Self-Refl, the sd is over the \PnDecodes\ sampled answers per question, and Maj@\TTCCallsMaj\ votes over \TTCCallsMaj\ samples per question, with the sd over \PnTTCDraws\ such draws. {} Thesis accuracy for the same cells is in Table~\ref{tab:thesisacc}. \textbf{The lower block freezes the first answer and an external antithesis, the same for every arm}: the base model's first answer, and an antithesis written by an external model, Gemini-3.8-flash or Claude-Haiku-4.5, averaged over the \PnNbench\ benchmarks (App.~\ref{sec:sharedapp}). \emph{Self-Disagreement leads in every row, including when the challenge is the same for every arm.}}
\label{tab:panelbench}
\setlength{\tabcolsep}{2pt}
{\begin{tabular}{@{\hspace{3pt}}ll@{\hskip 4pt}c@{\hskip 6pt}ccccc@{\hskip 6pt}cc@{\hspace{3pt}}}
\toprule
\headrow \textbf{Model} & \textbf{Benchmark} & \textbf{Base} & \textbf{SelfD-RL} & \textbf{Vf-RL} & \textbf{Rv-RL} & \textbf{Ac-RL} & \textbf{SCoRe} & \textbf{Maj@\TTCCallsMaj} & \textbf{Self-Refl} \\
\midrule
& AGIEval SAT-Math\,~{\tiny\citep{agieval}} & \ce{\PnBaseAgieval}{\SdBaseAgieval} & \ceb{\PnSedAgieval}{\SdSedAgieval} & \ce{\PnVerAgieval}{\SdVerAgieval} & \ce{\PnRevAgieval}{\SdRevAgieval} & \ce{\PnAccAgieval}{\SdAccAgieval} & \ce{\PnScoreAgieval}{\SdScoreAgieval} & \ce{\PnMajAgieval}{\SdMajAgieval} & \ce{\PnReflAgieval}{\SdReflAgieval} \\
& AIME 2025\,~{\tiny\citep{aimeTwoFive}} & \ce{\PnBaseAime}{\SdBaseAime} & \ceb{\PnSedAime}{\SdSedAime} & \ce{\PnVerAime}{\SdVerAime} & \ce{\PnRevAime}{\SdRevAime} & \ce{\PnAccAime}{\SdAccAime} & \ce{\PnScoreAime}{\SdScoreAime} & \ce{\PnMajAime}{\SdMajAime} & \ce{\PnReflAime}{\SdReflAime} \\
& GPQA (main$\setminus$diamond)~{\tiny\citep{rein2024gpqa}} & \ce{\PnBaseGpqa}{\SdBaseGpqa} & \ceb{\PnSedGpqa}{\SdSedGpqa} & \ce{\PnVerGpqa}{\SdVerGpqa} & \ce{\PnRevGpqa}{\SdRevGpqa} & \ce{\PnAccGpqa}{\SdAccGpqa} & \ce{\PnScoreGpqa}{\SdScoreGpqa} & \ce{\PnMajGpqa}{\SdMajGpqa} & \ce{\PnReflGpqa}{\SdReflGpqa} \\
& LiveCodeBench~{\tiny\citep{jain2024livecodebench}} & \ce{\PnBaseLcb}{\SdBaseLcb} & \ceb{\PnSedLcb}{\SdSedLcb} & \ce{\PnVerLcb}{\SdVerLcb} & \ce{\PnRevLcb}{\SdRevLcb} & \ce{\PnAccLcb}{\SdAccLcb} & \ce{\PnScoreLcb}{\SdScoreLcb} & \ce{\PnMajLcb}{\SdMajLcb} & \ce{\PnReflLcb}{\SdReflLcb} \\
& MMLU-Redux\,~{\tiny\citep{mmluredux}} & \ce{\PnBaseMmluRedux}{\SdBaseMmluRedux} & \ceb{\PnSedMmluRedux}{\SdSedMmluRedux} & \ce{\PnVerMmluRedux}{\SdVerMmluRedux} & \ce{\PnRevMmluRedux}{\SdRevMmluRedux} & \ce{\PnAccMmluRedux}{\SdAccMmluRedux} & \ce{\PnScoreMmluRedux}{\SdScoreMmluRedux} & \ce{\PnMajMmluRedux}{\SdMajMmluRedux} & \ce{\PnReflMmluRedux}{\SdReflMmluRedux} \\
& MuSR-murder\,~{\tiny\citep{musr}} & \ce{\PnBaseMusr}{\SdBaseMusr} & \ceb{\PnSedMusr}{\SdSedMusr} & \ce{\PnVerMusr}{\SdVerMusr} & \ce{\PnRevMusr}{\SdRevMusr} & \ce{\PnAccMusr}{\SdAccMusr} & \ce{\PnScoreMusr}{\SdScoreMusr} & \ce{\PnMajMusr}{\SdMajMusr} & \ce{\PnReflMusr}{\SdReflMusr} \\
& Omni-MATH~{\tiny\citep{omnimath}} & \ce{\PnBaseOmni}{\SdBaseOmni} & \ceb{\PnSedOmni}{\SdSedOmni} & \ce{\PnVerOmni}{\SdVerOmni} & \ce{\PnRevOmni}{\SdRevOmni} & \ce{\PnAccOmni}{\SdAccOmni} & \ce{\PnScoreOmni}{\SdScoreOmni} & \ce{\PnMajOmni}{\SdMajOmni} & \ce{\PnReflOmni}{\SdReflOmni} \\
& SuperGPQA\,~{\tiny\citep{supergpqa}} & \ce{\PnBaseSupergpqa}{\SdBaseSupergpqa} & \ceb{\PnSedSupergpqa}{\SdSedSupergpqa} & \ce{\PnVerSupergpqa}{\SdVerSupergpqa} & \ce{\PnRevSupergpqa}{\SdRevSupergpqa} & \ce{\PnAccSupergpqa}{\SdAccSupergpqa} & \ce{\PnScoreSupergpqa}{\SdScoreSupergpqa} & \ce{\PnMajSupergpqa}{\SdMajSupergpqa} & \ce{\PnReflSupergpqa}{\SdReflSupergpqa} \\
& TheoremQA\,~{\tiny\citep{theoremqa}} & \ce{\PnBaseTheorem}{\SdBaseTheorem} & \ceb{\PnSedTheorem}{\SdSedTheorem} & \ce{\PnVerTheorem}{\SdVerTheorem} & \ce{\PnRevTheorem}{\SdRevTheorem} & \ce{\PnAccTheorem}{\SdAccTheorem} & \ce{\PnScoreTheorem}{\SdScoreTheorem} & \ce{\PnMajTheorem}{\SdMajTheorem} & \ce{\PnReflTheorem}{\SdReflTheorem} \\
\cmidrule(l){2-10}
\multirow{-11}{*}{\rot{\textbf{Qwen3-8B}}} & \textbf{Avg} & \ce{\PnBaseMacro}{\PnBaseSd} & \textbf{\ce{\PnSedMacro}{\PnSedSd}} & \ce{\PnVerMacro}{\PnVerSd} & \ce{\PnRevMacro}{\PnRevSd} & \ce{\PnAccMacro}{\PnAccSd} & \ce{\PnScoreMacro}{\PnScoreSd} & \ce{\PnMajMacro}{\PnMajSd} & \ce{\PnReflMacro}{\PnReflSd} \\
\midrule
\multicolumn{2}{@{\hspace{3pt}}l}{\textbf{Gemma-2 9B}$^{\ddagger}$} & \ce{\GnBaseMacro}{\GnBaseSd} & \textbf{\ce{\GnSedMacro}{\GnSedSd}} & \ce{\GnVerMacro}{\GnVerSd} & \ce{\GnRevMacro}{\GnRevSd} & \ce{\GnAccMacro}{\GnAccSd} & \ce{\GnScoreMacro}{\GnScoreSd} & \ce{\GnMajMacro}{\GnMajSd} & \ce{\GnReflMacro}{\GnReflSd} \\
\midrule
\hiderowcolors
\headrow & \hspace{\TOneLowShift}\textbf{Frozen external antithesis} & \multicolumn{8}{c}{\textbf{Frozen first answer and external antithesis, the same for every arm (App.~\ref{sec:sharedapp})}} \\
\rowcolor{tabband!4} & \hspace{\TOneLowShift}Gemini-3.8-flash & \ce{\XtQGemBase}{\XtQGemBaseSd} & \ceb{\XtQGemSed}{\XtQGemSedSd} & \ce{\XtQGemVer}{\XtQGemVerSd} & \ce{\XtQGemRev}{\XtQGemRevSd} & \ce{\XtQGemAcc}{\XtQGemAccSd} & \ce{\XtQGemScore}{\XtQGemScoreSd} & \ce{\XtQGemMaj}{\XtQGemMajSd} & \ce{\XtQGemRefl}{\XtQGemReflSd} \\
\hiderowcolors
\multirow{-2}{*}{\makebox[0pt][l]{\textbf{Qwen3-8B}}} & \hspace{\TOneLowShift}Claude-Haiku-4.5 & \ce{\XtQHaiBase}{\XtQHaiBaseSd} & \ceb{\XtQHaiSed}{\XtQHaiSedSd} & \ce{\XtQHaiVer}{\XtQHaiVerSd} & \ce{\XtQHaiRev}{\XtQHaiRevSd} & \ce{\XtQHaiAcc}{\XtQHaiAccSd} & \ce{\XtQHaiScore}{\XtQHaiScoreSd} & \ce{\XtQHaiMaj}{\XtQHaiMajSd} & \ce{\XtQHaiRefl}{\XtQHaiReflSd} \\
\rowcolor{tabband!4} & \hspace{\TOneLowShift}Gemini-3.8-flash & \ce{\XtGGemBase}{\XtGGemBaseSd} & \ceb{\XtGGemSed}{\XtGGemSedSd} & \ce{\XtGGemVer}{\XtGGemVerSd} & \ce{\XtGGemRev}{\XtGGemRevSd} & \ce{\XtGGemAcc}{\XtGGemAccSd} & \ce{\XtGGemScore}{\XtGGemScoreSd} & \ce{\XtGGemMaj}{\XtGGemMajSd} & \ce{\XtGGemRefl}{\XtGGemReflSd} \\
\hiderowcolors
\multirow{-2}{*}{\makebox[0pt][l]{\textbf{Gemma-2~9B}}} & \hspace{\TOneLowShift}Claude-Haiku-4.5 & \ce{\XtGHaiBase}{\XtGHaiBaseSd} & \ceb{\XtGHaiSed}{\XtGHaiSedSd} & \ce{\XtGHaiVer}{\XtGHaiVerSd} & \ce{\XtGHaiRev}{\XtGHaiRevSd} & \ce{\XtGHaiAcc}{\XtGHaiAccSd} & \ce{\XtGHaiScore}{\XtGHaiScoreSd} & \ce{\XtGHaiMaj}{\XtGHaiMajSd} & \ce{\XtGHaiRefl}{\XtGHaiReflSd} \\
\bottomrule
\end{tabular}}
\end{table}

\noindent\textbf{Why none of the alternatives reach it:} Table~\ref{tab:panelbench}
sets six of them beside ours, and each falls short for its own reason. Giving the base model more inference --- Majority Vote over \TTCCallsMaj\ samples an item, or Test-Time
Self-Reflection asking it once to review its answer --- buys more attempts but no one to
disagree with, so what is missing is not compute. Revise-RL trains a second pass,
rewarded on the answer it settles on and penalized for abandoning a correct one, but
with nothing disagreeing with it there is no valid challenge to refuse, so what is missing
is not the opportunity to change the answer. Verify-RL supplies the verdict by having
the model judge its own answer first, and rewards the judgment alongside the answer;
it comes closest of the four and still recovers
at most about half of our gain on both base models: a verdict a model issues against itself is not the same
as being argued with. SCoRe~\citep{kumar2024score}, while better than either on the panel, recovering more than half of our gain on both base models, leaves capitulation where the base model has it (Table~\ref{tab:mechanism}). The sixth is the arm built to maximize the quantity
our result might be mistaken for: Accuracy-RL trains on the same data with the
same verifier, differing only in taking a single pass rewarded on whether that answer
is right; its interval on the panel includes zero on Qwen3-8B, and on Gemma-2~9B it gains \GnAccGain\ points, under a fifth of ours. Optimizing the first answer does
not deliver the capability as a by-product, which it would have to for our
gain to be better answers, not better judgment --- what a model carries into the
dialectic and what survives it are separable, and a model can reach a right answer
and still not distinguish it from a wrong one --- though not the
reverse, since a model that cannot reach it has nothing to distinguish.

\label{sec:dissociation}
{}
{}
{}

\noindent\textbf{The same dissociation in the reward ablation:} Holding the exchange
fixed and varying only what is returned moves the dialectical quantity several times
further than it moves accuracy: across the reward ablation, one training seed per arm, the arms span \SevenCSSpreadPP\ pp
in synthesis accuracy against \SevenCTSpreadPP\ pp in thesis accuracy ---
\SevenRatio\ times as far --- and the ordering is not a trade. The same separation also appears within one arm over training
(\S\ref{app:rewardablation}).

\subsection{Where the gain comes from}

\noindent Synthesis accuracy rises, but it does not say why, and two mechanisms would raise
it equally: the model may answer better on the first attempt and so meet fewer challenges
it deserves to lose, or its dialectical competence may itself have improved. Grading its first
answer against its challenger's sorts every item into four situations by which side was
right. Writing
$\pi_m$ for the share of items in situation $m$ and $r_m$ for how often the model ends on
the correct answer there, synthesis accuracy is $c_S=\sum_m \pi_m r_m$, and the change
against the base model splits without remainder into a \emph{situations} term
and a \emph{skill} term, $\Delta c_S = \sum_m \Delta\pi_m\, r^{\mathrm{base}}_m + \sum_m \pi^{\mathrm{arm}}_m\, \Delta r_m$,
the standard composition-versus-rate decomposition~\citep{kitagawa1955components},
where \emph{situations}, the first term, is what the model gains by walking into a different mix of
predicaments --- the better-first-answer explanation, the floor --- and \emph{skill}, the second, is the part of the
gain due to a change in dialectical competence, the capability standing on that floor rather than
an alternative to it.
We decompose it on the D-probe, over the same \LxN\ challengers;
Appendix~\ref{sec:dcstratum} gives the full profile and the reference policies.

\begin{table}[t]
\centering\scriptsize
\setlength{\tabcolsep}{3pt}
\caption{\textbf{Only Self-Disagreement gains \emph{skill}; the other arms gain only \emph{situations}.}
\emph{Situations} is what an arm gains by walking into an easier mix of predicaments
--- a better first answer; \emph{skill} is what it gains by handling the same predicament
better.
Points of synthesis accuracy on the D-probe; for each arm the two terms sum to its total change.
Brackets are 95\,\% intervals clustered by question; each trained arm is averaged over three training seeds, and its bracket is that average's. The base model is the reference.}
\label{tab:dcdecomp}
\begin{tabular}{l cc cc}
\toprule
\headrow & \multicolumn{2}{c}{\textbf{Qwen3-8B}} & \multicolumn{2}{c}{\textbf{Gemma-2 9B}} \\
\cmidrule(lr){2-3}\cmidrule(lr){4-5}
\headrow \textbf{Arm}
 & \dcterm{situations}{\sum\Delta\pi_m r^{\mathrm{base}}_m} & \dcterm{skill}{\sum\pi^{\mathrm{arm}}_m \Delta r_m}
 & \dcterm{situations}{\sum\Delta\pi_m r^{\mathrm{base}}_m} & \dcterm{skill}{\sum\pi^{\mathrm{arm}}_m \Delta r_m} \\
\midrule
Accuracy-RL & $+\DcSituQAcc$ {\tiny$[+\DcSituQAccLo,+\DcSituQAccHi]$} & $\DcSkillQAcc$ {\tiny$[\DcSkillQAccLo,\DcSkillQAccHi]$}
            & $+\DcSituGAcc$ {\tiny$[+\DcSituGAccLo,+\DcSituGAccHi]$} & $\DcSkillGAcc$ {\tiny$[\DcSkillGAccLo,\DcSkillGAccHi]$} \\
Revise-RL   & $+\DcSituQRev$ {\tiny$[\DcSituQRevLo,+\DcSituQRevHi]$} & $\DcSkillQRev$ {\tiny$[\DcSkillQRevLo,$$\DcSkillQRevHi]$}
            & $+\DcSituGRev$ {\tiny$[+\DcSituGRevLo,+\DcSituGRevHi]$} & $\DcSkillGRev$ {\tiny$[\DcSkillGRevLo,\DcSkillGRevHi]$} \\
Verify-RL   & $+\DcSituQVer$ {\tiny$[+\DcSituQVerLo,+\DcSituQVerHi]$} & $\DcSkillQVer$ {\tiny$[\DcSkillQVerLo,+\DcSkillQVerHi]$}
            & $+\DcSituGVer$ {\tiny$[+\DcSituGVerLo,+\DcSituGVerHi]$} & $\DcSkillGVer$ {\tiny$[\DcSkillGVerLo,+\DcSkillGVerHi]$} \\
SCoRe & $\DcSituQScore$ {\tiny$[\DcSituQScoreLo,\DcSituQScoreHi]$} & $\DcSkillQScore$ {\tiny$[\DcSkillQScoreLo,\DcSkillQScoreHi]$}
            & $\DcSituGScore$ {\tiny$[\DcSituGScoreLo,\DcSituGScoreHi]$} & $\DcSkillGScore$ {\tiny$[\DcSkillGScoreLo,\DcSkillGScoreHi]$} \\
SelfD-RL      & $+\DcSituQ$ {\tiny$[$$+$$\DcSituQLo,+\DcSituQHi]$} & $\bm{+\DcSkillQ}$ {\tiny$[+\DcSkillQLo,+\DcSkillQHi]$}
            & $+\DcSituG$ {\tiny$[\DcSituGLo,+\DcSituGHi]$} & $\bm{+\DcSkillG}$ {\tiny$[+\DcSkillGLo,+\DcSkillGHi]$} \\
\bottomrule
\end{tabular}
\end{table}

\noindent\textbf{Every arm improves its first answer; only Self-Disagreement also improves the
decision:} Table~\ref{tab:dcdecomp} gives the split, and
Appendix~\ref{sec:dcstratum} the four situations behind it. The base models are one-sided ---
holding only \MhQBase\,\% and \MhGBase\,\% of the time while yielding \MyQBase\,\% and
\MyGBase\,\%, which is closer to unconditional surrender than to discrimination. Self-Disagreement
holds \MhQSed\,\% and \MhGSed\,\% and still yields \MyQSed\,\% and \MyGSed\,\%, and that registers as skill: \DcSkillQ\ of its \DcGainQ-point gain on Qwen and
\DcSkillG\ of \DcGainG\ on Gemma, positive against at least \DcSkillNposMin\ of the \LxN\ challengers individually in every training seed of both families. A model deciding no better than the base scores zero;
one deciding correctly in every situation it met would score
\DcSkillCeilQ{} on Qwen and \DcSkillCeilG\ on Gemma. The trained rivals behave alike and oppositely: each gains through
situations, and none has a skill term above zero. SCoRe gains no skill either: $\DcSkillQScore$ on Qwen and $\DcSkillGScore$ on Gemma, neither interval above zero.

\begin{wrapfigure}{R}{0.40\textwidth}
\centering
\includegraphics[width=\linewidth]{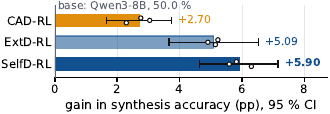}\caption{\textbf{The content of the disagreement matters.} Average over the
\PnNbench{} benchmarks and three training seeds (dots), each scored on \PnDecodes{} sampled answers. Bars are the gain over the base model, with 95\,\% CIs.}
\label{fig:altdesign}
\end{wrapfigure}

\noindent\textbf{Does the disagreement have to be genuine, and does it have to
come from elsewhere:} Two variants of the recipe answer this, each removing one
ingredient; Figure~\ref{fig:altdesign} places them beside the full recipe.
Content-Agnostic Disagreement (CAD-RL) runs the same exchange
against a challenge that does not depend on the answer it challenges, and it
reaches \PnPlcD\ $[\PnPlcDLo, \PnPlcDHi]$, roughly a third to a half of the full recipe's gain across training seeds; its best seed stays \PnPlcRunGap\ points below the full recipe's weakest. The presence of opposition carries part of the gain, and the rest requires the opposition to be about
something, which is also what makes the verdict on it checkable (App.~\ref{sec:prompts}). External Disagreement (ExtD-RL) is our own recipe with the antithesis
drawn from Qwen3-14B, a larger model of the same family, and it reaches \PnExtD\ $[\PnExtDLo, \PnExtDHi]$
against \PnSedD\ for the policy arguing with itself: a larger model's opposition adds nothing over the policy's own. What
separates them is less what they achieve than what they require: the external arm
holds a second model in service for the whole of training, on a serving
allocation held in a one-to-four ratio against the one training used, and it leaves
the choice of that model unresolved, since it was picked for being larger than
the model being trained and we know of no principled way to make that choice.
Both are reported on Qwen3-8B only: they vary the \emph{design} rather than
ablate ours, so neither is needed to establish that the effect reproduces on a
second base model.
One might object that both arms argue inside our own loop, so the
model may only handle its own objections. It cannot be: the
advantage holds against almost every one of the \LxNeverWord\ challengers it never trained against, several stronger than itself (Table~\ref{tab:ladder}).

\label{sec:mechanism}
{}
{}

  {}
  {}

\noindent\textbf{Which part of the design does the work:} training against an opposition with the answer
given after it rewarded, with a further gain from paying for a correct ruling; a
second pass with nothing to rule on adds none of it.
Of the arms in Table~\ref{tab:mechanism}, only the two whose answer after the
challenge is rewarded bring capitulation below the base model's \BaseKdrawPct\pct;
none of the others does, SCoRe included, nor the exchange whose final answer goes
unpaid.{}
\par
\begin{wraptable}{r}{0.44\textwidth}
\centering\scriptsize
\caption{\textbf{Capitulation falls only when the answer given after the challenge is rewarded; paying for a correct verdict lowers it further.}
\textbf{Qwen3-8B.} Capitulation and synthesis accuracy (\%).
\textbf{Self-Disagreement} full recipe; \textbf{Judge} verdict required, not rewarded;
\textbf{Unrew.} exchange runs, only the thesis paid; the other arms are as in Table~\ref{tab:panelbench}.
Rates are over \emph{all} scored items, not the challenged
subset. Mean\,$\pm$\,sd over three training seeds, each the mean of four draws (for Base, over its four draws). Protocol in \S\ref{sec:repro}.}
\label{tab:mechanism}
\setlength{\tabcolsep}{3pt}
\begin{tabular}{l@{\hskip 4pt}cc}
\toprule
\headrow \textbf{Arm} & \textbf{capitulation} $\downarrow$ & \textbf{synthesis} $\uparrow$ \\
\midrule
SelfD-RL   & $\EOneHOnePct \pm \EOneHOneSDPct$ & $\EOneHOneAccPct \pm \EOneHOneAccSDPct$ \\
Judge   & $\EOneJZeroPct \pm \EOneJZeroSDPct$ & $\EOneJZeroAccPct \pm \EOneJZeroAccSDPct$ \\
Rv-RL   & $\EOneSZeroPct \pm \EOneSZeroSDPct$ & $\EOneSZeroAccPct \pm \EOneSZeroAccSDPct$ \\
Unrew.   & $\TZeroKdrawPct \pm \TZeroKdrawSDPct$ & $\TZeroAccPct \pm \TZeroAccSDPct$ \\
Ac-RL    & $\MZeroKdrawPct \pm \MZeroKdrawSDPct$ & $\MZeroAccPct \pm \MZeroAccSDPct$ \\
SCoRe & $\EOneScorePct \pm \EOneScoreSDPct$ & $\EOneScoreAccPct \pm \EOneScoreAccSDPct$ \\
Base     & $\BaseKdrawPct \pm \BaseKdrawSDPct$ & $\BaseAccPct \pm \BaseAccSDPct$ \\
\bottomrule
\end{tabular}
\end{wraptable}
Two further ablations
separate the verdict from the opposition (Table~\ref{tab:mechanism}). Requiring
the verdict in the output but paying nothing for getting it right gives
$\EOneJZeroPct \pm \EOneJZeroSDPct$\pct, against $\EOneHOnePct \pm
\EOneHOneSDPct$\pct\ for the full recipe --- a resolved gap of
\EOneJZeroGapPct\pp, separating the
\emph{structural} requirement from the reward for meeting it. A second
pass with nothing to rule on does not substitute: Revise-RL abandons its correct
answer on $\EOneSZeroPct \pm \EOneSZeroSDPct$\pct\ of all scored items,
\EOneSZeroGapPct\pp\ from the full recipe, and places it beside
the unrewarded-dialectic and Accuracy-RL arms, none of which is rewarded for an answer given after a challenge. The same reward creates an
obvious exploit, since it pays for correcting a flawed first answer, and
Appendix~\ref{sec:hacking} reports the controls: the rate at which the final answer simply
reproduces the challenger's falls from \CopyThesisOnly\ in the unrewarded-dialectic arm to
\CopyDialLo--\CopyDialHi\ in the arms that pay for the verdict, and the arm with the verdict removed
sits at \CopyNoJudge, with the control not the treatment.

\subsection{Validity and generalization}

\label{sec:construct}
{}

\noindent\textbf{Is the metric measuring what we claim:} As \S\ref{sec:motivation}
reports, a blinded jury of three frontier
models reads the reasoning and
finds what the metric counts to be a real failure rather than an artifact of comparing
two answer strings. \S\ref{sec:juryapp} gives the validation in full --- the
rubric, the panel, the
blinding, the agreement statistics, and the limits of what this grading establishes.

\begin{wraptable}{r}{0.52\textwidth}
\centering\scriptsize
\setlength{\tabcolsep}{2pt}
\caption{\textbf{The same recipe on other models, other sizes, and other
families.} Before training and at convergence, mean\,$\pm$\,sd over two training seeds (\%).
$^{\ddagger}$Qwen3~14B was \textbf{pre-registered as a near-null}
because it begins with almost no headroom to move; it is reported for that
reason.}
\label{tab:xfam}
\begin{tabular}{l@{\hskip 3pt}cc@{\hskip 5pt}cc}
\toprule
\headrow & \multicolumn{2}{c}{\textbf{capitulation} $\downarrow$} & \multicolumn{2}{c}{\textbf{synthesis} $\uparrow$} \\
\cmidrule(lr){2-3}\cmidrule(lr){4-5}
\headrow \textbf{Model} & base & \textbf{+\,SelfD-RL} & base & \textbf{+\,SelfD-RL} \\
\midrule
Gemma~3 12B     & $\XfGemmaThreeCavB \pm \XfGemmaThreeCavBsd$ & $\mathbf{\XfGemmaThreeCavA \pm \XfGemmaThreeCavAsd}$ & $\XfGemmaThreeAccB \pm \XfGemmaThreeAccBsd$ & $\mathbf{\XfGemmaThreeAccA \pm \XfGemmaThreeAccAsd}$ \\
Granite~3.1 8B  & $\XfGraniteCavB \pm \XfGraniteCavBsd$ & $\mathbf{\XfGraniteCavA \pm \XfGraniteCavAsd}$ & $\XfGraniteAccB \pm \XfGraniteAccBsd$ & $\mathbf{\XfGraniteAccA \pm \XfGraniteAccAsd}$ \\
Llama~3.1 8B    & $\XfLlamaCavB \pm \XfLlamaCavBsd$ & $\mathbf{\XfLlamaCavA \pm \XfLlamaCavAsd}$ & $\XfLlamaAccB \pm \XfLlamaAccBsd$ & $\mathbf{\XfLlamaAccA \pm \XfLlamaAccAsd}$ \\
Qwen3 14B$^{\ddagger}$ & $\XfQwenFourteenCavB \pm \XfQwenFourteenCavBsd$ & $\mathbf{\XfQwenFourteenCavA \pm \XfQwenFourteenCavAsd}$ & $\XfQwenFourteenAccB \pm \XfQwenFourteenAccBsd$ & $\mathbf{\XfQwenFourteenAccA \pm \XfQwenFourteenAccAsd}$ \\
\bottomrule
\end{tabular}
\end{wraptable}
\noindent\textbf{The gain holds against challengers that cannot have decayed:}
When the model argues with itself, thesis and antithesis are the same checkpoint,
so a rise in synthesis accuracy could mean the model holds better or that its own
antithesis got worse. Separating the two needs a challenger that does not move
with the arm, and the ladder is built from them: \LxFixedWord\ of the \LxN\
antithesis models are
served from fixed weights and are identical for every arm, so nothing an arm learns can change who argues with it. Two of those \LxFixedWord\ are the base
checkpoints the arms were trained from --- the frozen base models, each family's
own starting point held still and replayed --- which is the sharpest form of the
test, since it is the one challenger the arm might be suspected of having left
behind. Scoring synthesis accuracy on the same D-probe, with each model's own first answer frozen and replayed: against the frozen Qwen3-8B the trained model reaches
\PeerQwenSed\pct\ where its base reaches \PeerQwenBase\pct; against the frozen
gemma-2-9b-it, \PeerGemSed\pct\ against \PeerGemBase\pct. Those gains,
\PeerQwenD\ and \PeerGemD\ pp, are as large as the ones measured against the
model's own antithesis, which is what rules the alternative reading out: a
decaying antithesis cannot explain a gain that survives almost every one of the \LxNeverWord\ challengers that
never changed.

\label{sec:robustness}
\noindent\textbf{Across other model families and sizes:} The result is not a
property of one base model.
Applying the same recipe to
\XFamModels\ further models --- other sizes and other families --- synthesis accuracy rises and capitulation falls in every one, whether or not its first answers improve, on the same held-out items scored before and after training. All four end below \XFamBandPct\pct, and
\XFamModelsAboveBand\ of them started above it (Table~\ref{tab:xfam}).

\noindent\textbf{Does it depend on who argues:} Table~\ref{tab:ladder}, in
Appendix~\ref{sec:dcstratum}, puts both models against every antithesis we ran, from a 4B model to the strongest frontier systems available,
including the base model itself and a base model of the same size.
Over much of that ladder the base model would have done better by
ignoring the antithesis entirely --- keeping its first answer, or taking the
challenger's, whichever is stronger --- and at worst it gives up
\LxBaseWorst\ pp that way on Qwen3-8B; its accuracy swings \LxBaseSpan\ pp from one
end of the ladder to the other (\LgBaseSpan\ on Gemma-2~9B). Self-Disagreement falls short of its own reference
by at most \LxSedWorst{} on Qwen3-8B, and swings
\LxSedSpan{} (\LgSedSpan\ on Gemma-2~9B). Whatever it is arguing with, it ends up in much the same place, and
that difference between the two spans is large enough that no rung on the ladder
reverses it. A fixed-population measurement with different challengers finds the
same invariance (\S\ref{sec:ladderapp}).

\label{sec:boundary}
\noindent\textbf{A model that ignores the exchange scores exactly one; neither of
ours does:} For every item on which one of the thesis and the antithesis was
right and the other wrong, we ask whether the model ended up on the side the
evidence favored. Such a policy
holds just as often when holding is wrong as when it is right, so its two odds
are equal and its ratio is exactly one, whatever mix of items it meets.
Neither model is near that. On the same probe and the same
\LxN\ antitheses as Table~\ref{tab:ladder}, the odds are \OrBaseQ\ for the base
model against \OrSedQ\ for the trained one on Qwen3-8B, and \OrBaseG\ against
\OrSedG\ on Gemma-2 9B. SCoRe's are \OrScoreQ{} on Qwen3-8B and \OrScoreG{} on Gemma-2 9B.
Read antithesis by antithesis, Self-Disagreement's ratio is above one against all \LxN\ antitheses on both base models once its three training seeds are pooled, the lowest being \OrQweakestAll\ $[\OrQweakestAllLo, \OrQweakestAllHi]$ on Qwen3-8B and \OrGweakestAll\ $[\OrGweakestAllLo, \OrGweakestAllHi]$ on Gemma-2~9B. Where a single seed falls short, the challenger is among the weakest --- small local models, including the base model itself --- where
holding is the right move on about \OrWeakHoldRight\ and the deciding cell holds
\OrCellLoRun\ to \OrCellHiRun{} items. Those exceptions are lower bounds: a
model that swapped one wrong answer for another scores as having held, pushing
the ratio down. Here each model meets its own theses, so each ratio is read against the null of one; given the base model's exchange instead, Self-Disagreement's log odds ratio is above the base model's and every other arm's (App.~\ref{sec:sharedapp}).

\section{Limitations and concluding remarks}
\label{sec:limitations}

\noindent \textbf{What we have not established:} The recipe is trained on two base
models from different families and applied unchanged to \XFamModels\ further models
across other sizes and families, where capitulation falls in every one. That is
evidence that the effect reproduces, not that it is universal: establishing it across
model families and scales in general would take a far wider sweep than the compute
here allows, and we do not claim it. The same bound applies to the domain. Every
reward here comes from exact-match verification against a reference answer,
so our evidence covers only checkable answers; whether the same signal can
be built where correctness is not checkable is open. Every dialectical exchange in the main
text is also a single round of challenge; three rounds are in \S\ref{sec:repeatapp}.

\noindent \textbf{How far there is still to go:} Self-Disagreement moves the decision
itself, but it does not finish the job. Its skill term reaches \DcSkillQ\ pp on
Qwen3-8B and \DcSkillG\ on Gemma-2 9B, against the \DcSkillCeilQ\ and
\DcSkillCeilG\ a model deciding
correctly in every situation it met would score --- roughly a third of what is
available in each case.
The gain that is there is not refusal: when both receive the same first answer and the same challenge, the
trained model reaches the right final answer more often than the base model does (App.~\ref{sec:sharedapp}), so we take part of that
residual to be a limit of the model rather than of the
training: dialectical competence stands on first-answer accuracy without following
from it, and a model cannot rule correctly on a question it is not capable of getting
right in the first place. Closing the rest is what the next work has to do, and the
reference policies in \S\ref{sec:dcstratum} are what it should be measured against.

\noindent \textbf{Concluding remarks:} What limits progress in language models is
increasingly not what they
can do but our ability to tell when they are right. Every remedy for that which depends on a
stronger judge --- a larger model, a separate verifier, a person --- inherits the
limit, and runs out exactly where it is needed most: at the frontier, where
nothing stronger is available to ask. This work shows that one part of that
judgment can be placed inside the model, and trained there with nothing more
capable than the model being trained. It is one part and not the whole of it, and
the part still missing is the one that matters most --- because a model that
could reliably feel the force of an argument it had raised against itself would
not merely be arguing well; it would be supervising itself. We take that to be the direction
this line of work points, and the standard by which it should be judged.

\section*{Reproducibility statement}
{}
{}
\S\ref{sec:repro} gives the models and hardware, the training data and how it is
selected, the optimizer configuration, and the reporting point at which every arm
is read. Table~\ref{tab:panelbench} carries the per-arm and per-benchmark results
for Qwen3-8B and the per-arm results for Gemma-2 9B, whose per-benchmark detail is
in Table~\ref{tab:gemmapanel}; Appendix~\ref{sec:tables} the same breakdown for the
two alternative designs, and Appendices~\ref{sec:oppgenuine} and~\ref{sec:juryapp}
carry the construct checks on the two metrics the paper turns on.

\section*{AI use statement}
In this work, we used generative AI tools to implement methods: the training and
evaluation code, in addition to running and monitoring jobs on our GPU cluster.
Code implementation was restricted to decisions the authors had already taken,
not the design of the method or of the experiments it is evaluated in. We have
reviewed and verified any code generated by AI during development, and every
artifact so produced was read and tested by the authors before any result
derived from it entered this paper. We have
not used generative AI tools to develop the conceptual framing of this work, to
propose or refine its hypotheses, to design or give feedback on its methodology
or experiments, to generate synthetic data sets, to clean or reformat the
training data, to assist with translation, to support qualitative or thematic
analysis of our data, or to interpret results; the proof-related tasks,
formulating mathematical claims, providing critical ingredients for proofs, and
writing proofs, are not applicable to this work, which contains no theorems.
Generative models also appear inside the work itself rather than as authoring
tools, as the challengers the method is trained and evaluated against and as the
blinded jury of \S\ref{sec:juryapp} that reads model reasoning and judges
whether a concession was genuine; both are described in the paper as objects of
study. Additionally, we used generative AI tools to edit the text of this paper
for readability; and we have not used them to draft sections of this paper or to
find related work. We have
reviewed all AI-assisted work: the analysis, the claims and the limits placed on
them are the authors' own. We take responsibility for the final content of this
work, including text, claims or artifacts produced with the aid of generative
AI.
\bibliographystyle{iclr2027_conference}
\bibliography{iclr2027_conference,ref1,ref2}

\begin{thebibliography}{52}
\providecommand{\natexlab}[1]{#1}
\providecommand{\url}[1]{\texttt{#1}}
\expandafter\ifx\csname urlstyle\endcsname\relax
  \providecommand{\doi}[1]{doi: #1}\else
  \providecommand{\doi}{doi: \begingroup \urlstyle{rm}\Url}\fi

\bibitem[Abbasloo(2025)]{abbasloo2025siev}
Soheil Abbasloo.
\newblock Evaluating {LLM} reasoning beyond correctness and {CoT}.
\newblock \emph{arXiv preprint arXiv:2510.18134}, 2025.
\newblock v2, 2026-02-12.

\bibitem[{AI-MO}(2024)]{maa23amc}
{AI-MO}.
\newblock {AIMO} validation {AMC}: {AMC} 2022--2023 competition problems.
\newblock \url{https://huggingface.co/datasets/AI-MO/aimo-validation-amc},
  2024.
\newblock Accessed 2026-09-24.

\bibitem[Chen et~al.(2023)Chen, Yin, Ku, Lu, Wan, Ma, Xu, Wang, and
  Xia]{theoremqa}
Wenhu Chen, Ming Yin, Max Ku, Pan Lu, Yixin Wan, Xueguang Ma, Jianyu Xu, Xinyi
  Wang, and Tony Xia.
\newblock {TheoremQA}: A theorem-driven question answering dataset.
\newblock \emph{arXiv preprint arXiv:2305.12524}, 2023.

\bibitem[Clark et~al.(2018)Clark, Cowhey, Etzioni, Khot, Sabharwal, Schoenick,
  and Tafjord]{arcchallenge}
Peter Clark, Isaac Cowhey, Oren Etzioni, Tushar Khot, Ashish Sabharwal, Carissa
  Schoenick, and Oyvind Tafjord.
\newblock Think you have solved question answering? try {ARC}, the {AI2}
  reasoning challenge.
\newblock \emph{arXiv preprint arXiv:1803.05457}, 2018.

\bibitem[DeepSeek-AI et~al.(2025)DeepSeek-AI, Guo, Yang, Zhang, Song, Wang,
  Zhu, Xu, Zhang, Ma, Bi, Zhang, Yu, Wu, Wu, Gou, Shao, Li, Gao, Liu, Xue,
  Wang, Wu, Feng, Lu, Zhao, Deng, Zhang, Ruan, Dai, Chen, Ji, Li, Lin, Dai,
  Luo, Hao, Chen, Li, Zhang, Bao, Xu, Wang, Ding, Xin, Gao, Qu, Li, Guo, Li,
  Wang, Chen, Yuan, Qiu, Li, Cai, Ni, Liang, Chen, Dong, Hu, Gao, Guan, Huang,
  Yu, Wang, Zhang, Zhao, Wang, Zhang, Xu, Xia, Zhang, Zhang, Tang, Li, Wang,
  Li, Tian, Huang, Zhang, Wang, Chen, Du, Ge, Zhang, Pan, Wang, Chen, Jin,
  Chen, Lu, Zhou, Chen, Ye, Wang, Yu, Zhou, Pan, Li, Zhou, Wu, Ye, Yun, Pei,
  Sun, Wang, Zeng, Zhao, Liu, Liang, Gao, Yu, Zhang, Xiao, An, Liu, Wang, Chen,
  Nie, Cheng, Liu, Xie, Liu, Yang, Li, Su, Lin, Li, Jin, Shen, Chen, Sun, Wang,
  Song, Zhou, Wang, Shan, Li, Wang, Wei, Zhang, Xu, Li, Zhao, Sun, Wang, Yu,
  Zhang, Shi, Xiong, He, Piao, Wang, Tan, Ma, Liu, Guo, Ou, Wang, Gong, Zou,
  He, Xiong, Luo, You, Liu, Zhou, Zhu, Xu, Huang, Li, Zheng, Zhu, Ma, Tang,
  Zha, Yan, Ren, Ren, Sha, Fu, Xu, Xie, Zhang, Hao, Ma, Yan, Wu, Gu, Zhu, Liu,
  Li, Xie, Song, Pan, Huang, Xu, Zhang, and Zhang]{guo2025deepseekr1}
DeepSeek-AI, Daya Guo, Dejian Yang, Haowei Zhang, Junxiao Song, Peiyi Wang,
  Qihao Zhu, Runxin Xu, Ruoyu Zhang, Shirong Ma, Xiao Bi, Xiaokang Zhang,
  Xingkai Yu, Yu~Wu, Z.~F. Wu, Zhibin Gou, Zhihong Shao, Zhuoshu Li, Ziyi Gao,
  Aixin Liu, Bing Xue, Bingxuan Wang, Bochao Wu, Bei Feng, Chengda Lu,
  Chenggang Zhao, Chengqi Deng, Chenyu Zhang, Chong Ruan, Damai Dai, Deli Chen,
  Dongjie Ji, Erhang Li, Fangyun Lin, Fucong Dai, Fuli Luo, Guangbo Hao,
  Guanting Chen, Guowei Li, H.~Zhang, Han Bao, Hanwei Xu, Haocheng Wang,
  Honghui Ding, Huajian Xin, Huazuo Gao, Hui Qu, Hui Li, Jianzhong Guo, Jiashi
  Li, Jiawei Wang, Jingchang Chen, Jingyang Yuan, Junjie Qiu, Junlong Li, J.~L.
  Cai, Jiaqi Ni, Jian Liang, Jin Chen, Kai Dong, Kai Hu, Kaige Gao, Kang Guan,
  Kexin Huang, Kuai Yu, Lean Wang, Lecong Zhang, Liang Zhao, Litong Wang, Liyue
  Zhang, Lei Xu, Leyi Xia, Mingchuan Zhang, Minghua Zhang, Minghui Tang, Meng
  Li, Miaojun Wang, Mingming Li, Ning Tian, Panpan Huang, Peng Zhang, Qiancheng
  Wang, Qinyu Chen, Qiushi Du, Ruiqi Ge, Ruisong Zhang, Ruizhe Pan, Runji Wang,
  R.~J. Chen, R.~L. Jin, Ruyi Chen, Shanghao Lu, Shangyan Zhou, Shanhuang Chen,
  Shengfeng Ye, Shiyu Wang, Shuiping Yu, Shunfeng Zhou, Shuting Pan, S.~S. Li,
  Shuang Zhou, Shaoqing Wu, Shengfeng Ye, Tao Yun, Tian Pei, Tianyu Sun,
  T.~Wang, Wangding Zeng, Wanjia Zhao, Wen Liu, Wenfeng Liang, Wenjun Gao,
  Wenqin Yu, Wentao Zhang, W.~L. Xiao, Wei An, Xiaodong Liu, Xiaohan Wang,
  Xiaokang Chen, Xiaotao Nie, Xin Cheng, Xin Liu, Xin Xie, Xingchao Liu, Xinyu
  Yang, Xinyuan Li, Xuecheng Su, Xuheng Lin, X.~Q. Li, Xiangyue Jin, Xiaojin
  Shen, Xiaosha Chen, Xiaowen Sun, Xiaoxiang Wang, Xinnan Song, Xinyi Zhou,
  Xianzu Wang, Xinxia Shan, Y.~K. Li, Y.~Q. Wang, Y.~X. Wei, Yang Zhang,
  Yanhong Xu, Yao Li, Yao Zhao, Yaofeng Sun, Yaohui Wang, Yi~Yu, Yichao Zhang,
  Yifan Shi, Yiliang Xiong, Ying He, Yishi Piao, Yisong Wang, Yixuan Tan,
  Yiyang Ma, Yiyuan Liu, Yongqiang Guo, Yuan Ou, Yuduan Wang, Yue Gong, Yuheng
  Zou, Yujia He, Yunfan Xiong, Yuxiang Luo, Yuxiang You, Yuxuan Liu, Yuyang
  Zhou, Y.~X. Zhu, Yanhong Xu, Yanping Huang, Yaohui Li, Yi~Zheng, Yuchen Zhu,
  Yunxian Ma, Ying Tang, Yukun Zha, Yuting Yan, Z.~Z. Ren, Zehui Ren, Zhangli
  Sha, Zhe Fu, Zhean Xu, Zhenda Xie, Zhengyan Zhang, Zhewen Hao, Zhicheng Ma,
  Zhigang Yan, Zhiyu Wu, Zihui Gu, Zijia Zhu, Zijun Liu, Zilin Li, Ziwei Xie,
  Ziyang Song, Zizheng Pan, Zhen Huang, Zhipeng Xu, Zhongyu Zhang, and Zhen
  Zhang.
\newblock Deepseek-r1: Incentivizing reasoning capability in llms via
  reinforcement learning.
\newblock \emph{arXiv preprint arXiv:2501.12948}, 2025.

\bibitem[Du et~al.(2023)Du, Li, Torralba, Tenenbaum, and
  Mordatch]{du2023debate}
Yilun Du, Shuang Li, Antonio Torralba, Joshua~B. Tenenbaum, and Igor Mordatch.
\newblock Improving factuality and reasoning in language models through
  multiagent debate.
\newblock \emph{arXiv preprint arXiv:2305.14325}, 2023.

\bibitem[Gao et~al.(2024)Gao, Song, Yang, Cai, Miao, Dong, Li, Ma, Chen, Xu,
  Tang, Wang, Zan, Quan, Zhang, Sha, Zhang, Ren, Liu, and Chang]{omnimath}
Bofei Gao, Feifan Song, Zhe Yang, Zefan Cai, Yibo Miao, Qingxiu Dong, Lei Li,
  Chenghao Ma, Liang Chen, Runxin Xu, Zhengyang Tang, Benyou Wang, Daoguang
  Zan, Shanghaoran Quan, Ge~Zhang, Lei Sha, Yichang Zhang, Xuancheng Ren,
  Tianyu Liu, and Baobao Chang.
\newblock {Omni-MATH}: A universal olympiad level mathematic benchmark for
  large language models, 2024.
\newblock arXiv:2410.07985.

\bibitem[Gema et~al.(2024)Gema, Leang, Hong, Devoto, Mancino, Saxena, He, Zhao,
  Du, Madani, Barale, McHardy, Harris, Kaddour, van Krieken, and
  Minervini]{mmluredux}
Aryo~Pradipta Gema, Joshua Ong~Jun Leang, Giwon Hong, Alessio Devoto, Alberto
  Carlo~Maria Mancino, Rohit Saxena, Xuanli He, Yu~Zhao, Xiaotang Du, Mohammad
  Reza~Ghasemi Madani, Claire Barale, Robert McHardy, Joshua Harris, Jean
  Kaddour, Emile van Krieken, and Pasquale Minervini.
\newblock Are we done with {MMLU}?
\newblock \emph{arXiv preprint arXiv:2406.04127}, 2024.

\bibitem[{Gemma Team} et~al.(2024){Gemma Team}, Riviere, Pathak, Sessa, Hardin,
  Bhupatiraju, Hussenot, Mesnard, Shahriari, Ramé, Ferret, Liu, Tafti,
  Friesen, Casbon, Ramos, Kumar, Lan, Jerome, Tsitsulin, Vieillard, Stanczyk,
  Girgin, Momchev, Hoffman, Thakoor, Grill, Neyshabur, Bachem, Walton, Severyn,
  Parrish, Ahmad, Hutchison, Abdagic, Carl, Shen, Brock, Coenen, Laforge,
  Paterson, Bastian, Piot, Wu, Royal, Chen, Kumar, Perry, Welty,
  Choquette-Choo, Sinopalnikov, Weinberger, Vijaykumar, Rogozińska, Herbison,
  Bandy, Wang, Noland, Moreira, Senter, Eltyshev, Visin, Rasskin, Wei, Cameron,
  Martins, Hashemi, Klimczak-Plucińska, Batra, Dhand, Nardini, Mein, Zhou,
  Svensson, Stanway, Chan, Zhou, Carrasqueira, Iljazi, Becker, Fernandez, van
  Amersfoort, Gordon, Lipschultz, Newlan, yeong Ji, Mohamed, Badola, Black,
  Millican, McDonell, Nguyen, Sodhia, Greene, Sjoesund, Usui, Sifre, Heuermann,
  Lago, McNealus, Soares, Kilpatrick, Dixon, Martins, Reid, Singh, Iverson,
  Görner, Velloso, Wirth, Davidow, Miller, Rahtz, Watson, Risdal, Kazemi,
  Moynihan, Zhang, Kahng, Park, Rahman, Khatwani, Dao, Bardoliwalla,
  Devanathan, Dumai, Chauhan, Wahltinez, Botarda, Barnes, Barham, Michel, Jin,
  Georgiev, Culliton, Kuppala, Comanescu, Merhej, Jana, Rokni, Agarwal,
  Mullins, Saadat, Carthy, Cogan, Perrin, Arnold, Krause, Dai, Garg, Sheth,
  Ronstrom, Chan, Jordan, Yu, Eccles, Hennigan, Kocisky, Doshi, Jain, Yadav,
  Meshram, Dharmadhikari, Barkley, Wei, Ye, Han, Kwon, Xu, Shen, Gong, Wei,
  Cotruta, Kirk, Rao, Giang, Peran, Warkentin, Collins, Barral, Ghahramani,
  Hadsell, Sculley, Banks, Dragan, Petrov, Vinyals, Dean, Hassabis,
  Kavukcuoglu, Farabet, Buchatskaya, Borgeaud, Fiedel, Joulin, Kenealy,
  Dadashi, and Andreev]{gemma2}
{Gemma Team}, Morgane Riviere, Shreya Pathak, Pier~Giuseppe Sessa, Cassidy
  Hardin, Surya Bhupatiraju, Léonard Hussenot, Thomas Mesnard, Bobak
  Shahriari, Alexandre Ramé, Johan Ferret, Peter Liu, Pouya Tafti, Abe
  Friesen, Michelle Casbon, Sabela Ramos, Ravin Kumar, Charline~Le Lan, Sammy
  Jerome, Anton Tsitsulin, Nino Vieillard, Piotr Stanczyk, Sertan Girgin,
  Nikola Momchev, Matt Hoffman, Shantanu Thakoor, Jean-Bastien Grill, Behnam
  Neyshabur, Olivier Bachem, Alanna Walton, Aliaksei Severyn, Alicia Parrish,
  Aliya Ahmad, Allen Hutchison, Alvin Abdagic, Amanda Carl, Amy Shen, Andy
  Brock, Andy Coenen, Anthony Laforge, Antonia Paterson, Ben Bastian, Bilal
  Piot, Bo~Wu, Brandon Royal, Charlie Chen, Chintu Kumar, Chris Perry, Chris
  Welty, Christopher~A. Choquette-Choo, Danila Sinopalnikov, David Weinberger,
  Dimple Vijaykumar, Dominika Rogozińska, Dustin Herbison, Elisa Bandy, Emma
  Wang, Eric Noland, Erica Moreira, Evan Senter, Evgenii Eltyshev, Francesco
  Visin, Gabriel Rasskin, Gary Wei, Glenn Cameron, Gus Martins, Hadi Hashemi,
  Hanna Klimczak-Plucińska, Harleen Batra, Harsh Dhand, Ivan Nardini, Jacinda
  Mein, Jack Zhou, James Svensson, Jeff Stanway, Jetha Chan, Jin~Peng Zhou,
  Joana Carrasqueira, Joana Iljazi, Jocelyn Becker, Joe Fernandez, Joost van
  Amersfoort, Josh Gordon, Josh Lipschultz, Josh Newlan, Ju~yeong Ji, Kareem
  Mohamed, Kartikeya Badola, Kat Black, Katie Millican, Keelin McDonell, Kelvin
  Nguyen, Kiranbir Sodhia, Kish Greene, Lars~Lowe Sjoesund, Lauren Usui,
  Laurent Sifre, Lena Heuermann, Leticia Lago, Lilly McNealus, Livio~Baldini
  Soares, Logan Kilpatrick, Lucas Dixon, Luciano Martins, Machel Reid,
  Manvinder Singh, Mark Iverson, Martin Görner, Mat Velloso, Mateo Wirth, Matt
  Davidow, Matt Miller, Matthew Rahtz, Matthew Watson, Meg Risdal, Mehran
  Kazemi, Michael Moynihan, Ming Zhang, Minsuk Kahng, Minwoo Park, Mofi Rahman,
  Mohit Khatwani, Natalie Dao, Nenshad Bardoliwalla, Nesh Devanathan, Neta
  Dumai, Nilay Chauhan, Oscar Wahltinez, Pankil Botarda, Parker Barnes, Paul
  Barham, Paul Michel, Pengchong Jin, Petko Georgiev, Phil Culliton, Pradeep
  Kuppala, Ramona Comanescu, Ramona Merhej, Reena Jana, Reza~Ardeshir Rokni,
  Rishabh Agarwal, Ryan Mullins, Samaneh Saadat, Sara~Mc Carthy, Sarah Cogan,
  Sarah Perrin, Sébastien M.~R. Arnold, Sebastian Krause, Shengyang Dai,
  Shruti Garg, Shruti Sheth, Sue Ronstrom, Susan Chan, Timothy Jordan, Ting Yu,
  Tom Eccles, Tom Hennigan, Tomas Kocisky, Tulsee Doshi, Vihan Jain, Vikas
  Yadav, Vilobh Meshram, Vishal Dharmadhikari, Warren Barkley, Wei Wei, Wenming
  Ye, Woohyun Han, Woosuk Kwon, Xiang Xu, Zhe Shen, Zhitao Gong, Zichuan Wei,
  Victor Cotruta, Phoebe Kirk, Anand Rao, Minh Giang, Ludovic Peran, Tris
  Warkentin, Eli Collins, Joelle Barral, Zoubin Ghahramani, Raia Hadsell,
  D.~Sculley, Jeanine Banks, Anca Dragan, Slav Petrov, Oriol Vinyals, Jeff
  Dean, Demis Hassabis, Koray Kavukcuoglu, Clement Farabet, Elena Buchatskaya,
  Sebastian Borgeaud, Noah Fiedel, Armand Joulin, Kathleen Kenealy, Robert
  Dadashi, and Alek Andreev.
\newblock {Gemma 2}: Improving open language models at a practical size.
\newblock \emph{arXiv preprint arXiv:2408.00118}, 2024.

\bibitem[Geva et~al.(2021)Geva, Khashabi, Segal, Khot, Roth, and
  Berant]{strategyqa}
Mor Geva, Daniel Khashabi, Elad Segal, Tushar Khot, Dan Roth, and Jonathan
  Berant.
\newblock Did aristotle use a laptop? a question answering benchmark with
  implicit reasoning strategies.
\newblock \emph{arXiv preprint arXiv:2101.02235}, 2021.

\bibitem[Grattafiori et~al.(2024)Grattafiori, Dubey, Jauhri, Pandey, Kadian,
  Al-Dahle, Letman, Mathur, Schelten, Vaughan, et~al.]{grattafiori2024llama3}
Aaron Grattafiori, Abhimanyu Dubey, Abhinav Jauhri, Abhinav Pandey, Abhishek
  Kadian, Ahmad Al-Dahle, Aiesha Letman, Akhil Mathur, Alan Schelten, Alex
  Vaughan, et~al.
\newblock The llama 3 herd of models.
\newblock \emph{arXiv preprint arXiv:2407.21783}, 2024.

\bibitem[Gu et~al.(2024)Gu, Rozière, Leather, Solar-Lezama, Synnaeve, and
  Wang]{cruxeval}
Alex Gu, Baptiste Rozière, Hugh Leather, Armando Solar-Lezama, Gabriel
  Synnaeve, and Sida~I. Wang.
\newblock {CRUXEval}: A benchmark for code reasoning, understanding and
  execution.
\newblock \emph{arXiv preprint arXiv:2401.03065}, 2024.

\bibitem[He et~al.(2024)He, Luo, Bai, Hu, Thai, Shen, Hu, Han, Huang, Zhang,
  Liu, Qi, Liu, and Sun]{olympiadbench}
Chaoqun He, Renjie Luo, Yuzhuo Bai, Shengding Hu, Zhen~Leng Thai, Junhao Shen,
  Jinyi Hu, Xu~Han, Yujie Huang, Yuxiang Zhang, Jie Liu, Lei Qi, Zhiyuan Liu,
  and Maosong Sun.
\newblock {OlympiadBench}: A challenging benchmark for promoting {AGI} with
  olympiad-level bilingual multimodal scientific problems, 2024.
\newblock arXiv:2402.14008.

\bibitem[Hegel(2018)]{hegel_phenomenology_spirit}
Georg Wilhelm~Friedrich Hegel.
\newblock \emph{The Phenomenology of Spirit}.
\newblock Cambridge Hegel Translations. Cambridge University Press, Cambridge,
  2018.
\newblock \doi{10.1017/9781139050494}.
\newblock Edited and translated by Terry Pinkard and Michael Baur.

\bibitem[Hendrycks et~al.(2020)Hendrycks, Burns, Basart, Zou, Mazeika, Song,
  and Steinhardt]{mmlu}
Dan Hendrycks, Collin Burns, Steven Basart, Andy Zou, Mantas Mazeika, Dawn
  Song, and Jacob Steinhardt.
\newblock Measuring massive multitask language understanding, 2020.
\newblock arXiv preprint arXiv:2009.03300.

\bibitem[Hendrycks et~al.(2021)Hendrycks, Burns, Kadavath, Arora, Basart, Tang,
  Song, and Steinhardt]{hendrycks2021math}
Dan Hendrycks, Collin Burns, Saurav Kadavath, Akul Arora, Steven Basart, Eric
  Tang, Dawn Song, and Jacob Steinhardt.
\newblock Measuring mathematical problem solving with the math dataset.
\newblock \emph{arXiv preprint arXiv:2103.03874}, 2021.

\bibitem[Hong et~al.(2023)Hong, Zhuge, Chen, Zheng, Cheng, Zhang, Wang, Wang,
  Yau, Lin, Zhou, Ran, Xiao, Wu, and Schmidhuber]{hong2023metagpt}
Sirui Hong, Mingchen Zhuge, Jiaqi Chen, Xiawu Zheng, Yuheng Cheng, Ceyao Zhang,
  Jinlin Wang, Zili Wang, Steven Ka~Shing Yau, Zijuan Lin, Liyang Zhou, Chenyu
  Ran, Lingfeng Xiao, Chenglin Wu, and Jürgen Schmidhuber.
\newblock {MetaGPT}: Meta programming for {A} multi-agent collaborative
  framework.
\newblock \emph{arXiv preprint arXiv:2308.00352}, 2023.

\bibitem[Jain et~al.(2024)Jain, Han, Gu, Li, Yan, Zhang, Wang, Solar-Lezama,
  Sen, and Stoica]{jain2024livecodebench}
Naman Jain, King Han, Alex Gu, Wen-Ding Li, Fanjia Yan, Tianjun Zhang, Sida
  Wang, Armando Solar-Lezama, Koushik Sen, and Ion Stoica.
\newblock Livecodebench: Holistic and contamination free evaluation of large
  language models for code.
\newblock \emph{arXiv preprint arXiv:2403.07974}, 2024.

\bibitem[Khalifa et~al.(2025)Khalifa, Agarwal, Logeswaran, Kim, Peng, Lee, Lee,
  and Wang]{khalifa2025thinkprm}
Muhammad Khalifa, Rishabh Agarwal, Lajanugen Logeswaran, Jaekyeom Kim, Hao
  Peng, Moontae Lee, Honglak Lee, and Lu~Wang.
\newblock Process reward models that think.
\newblock \emph{arXiv preprint arXiv:2504.16828}, 2025.

\bibitem[Kitagawa(1955)]{kitagawa1955components}
Evelyn~M. Kitagawa.
\newblock Components of a difference between two rates.
\newblock \emph{Journal of the American Statistical Association}, 50\penalty0
  (272):\penalty0 1168--1194, 1955.
\newblock \doi{10.1080/01621459.1955.10501299}.

\bibitem[Kumar et~al.(2024)Kumar, Zhuang, Agarwal, Su, Co-Reyes, Singh, Baumli,
  Iqbal, Bishop, Roelofs, Zhang, McKinney, Shrivastava, Paduraru, Tucker,
  Precup, Behbahani, and Faust]{kumar2024score}
Aviral Kumar, Vincent Zhuang, Rishabh Agarwal, Yi~Su, John~D Co-Reyes, Avi
  Singh, Kate Baumli, Shariq Iqbal, Colton Bishop, Rebecca Roelofs, Lei~M
  Zhang, Kay McKinney, Disha Shrivastava, Cosmin Paduraru, George Tucker, Doina
  Precup, Feryal Behbahani, and Aleksandra Faust.
\newblock Training language models to self-correct via reinforcement learning.
\newblock \emph{arXiv preprint arXiv:2409.12917}, 2024.

\bibitem[Laban et~al.(2023)Laban, Murakhovs'ka, Xiong, and
  Wu]{laban2023flipflop}
Philippe Laban, Lidiya Murakhovs'ka, Caiming Xiong, and Chien-Sheng Wu.
\newblock Are you sure? challenging {LLM}s leads to performance drops in the
  flipflop experiment.
\newblock \emph{arXiv preprint arXiv:2311.08596}, 2023.

\bibitem[Lee et~al.(2025)Lee, Oh, Kim, Shin, and Tack]{lee2025revise}
Hyunseok Lee, Seunghyuk Oh, Jaehyung Kim, Jinwoo Shin, and Jihoon Tack.
\newblock Revise: Learning to refine at test-time via intrinsic
  self-verification.
\newblock \emph{arXiv preprint arXiv:2502.14565}, 2025.

\bibitem[Lee et~al.(2026)Lee, Abbasloo, Tack, and Shin]{lee2026rltr}
Hyunseok Lee, Soheil Abbasloo, Jihoon Tack, and Jinwoo Shin.
\newblock Beyond correctness: Learning robust reasoning via transfer.
\newblock \emph{arXiv preprint arXiv:2602.08489}, 2026.

\bibitem[Liang et~al.(2023)Liang, He, Jiao, Wang, Wang, Wang, Yang, Shi, and
  Tu]{liang2023mad}
Tian Liang, Zhiwei He, Wenxiang Jiao, Xing Wang, Yan Wang, Rui Wang, Yujiu
  Yang, Shuming Shi, and Zhaopeng Tu.
\newblock Encouraging divergent thinking in large language models through
  multi-agent debate.
\newblock \emph{arXiv preprint arXiv:2305.19118}, 2023.

\bibitem[Lightman et~al.(2023)Lightman, Kosaraju, Burda, Edwards, Baker, Lee,
  Leike, Schulman, Sutskever, and Cobbe]{lightman2023step}
Hunter Lightman, Vineet Kosaraju, Yura Burda, Harri Edwards, Bowen Baker, Teddy
  Lee, Jan Leike, John Schulman, Ilya Sutskever, and Karl Cobbe.
\newblock Let's verify step by step.
\newblock \emph{arXiv preprint arXiv:2305.20050}, 2023.

\bibitem[Liu et~al.(2025)Liu, Chen, Li, Qi, Pang, Du, Lee, and
  Lin]{liu2025drgrpo}
Zichen Liu, Changyu Chen, Wenjun Li, Penghui Qi, Tianyu Pang, Chao Du, Wee~Sun
  Lee, and Min Lin.
\newblock Understanding r1-zero-like training: A critical perspective.
\newblock \emph{arXiv preprint arXiv:2503.20783}, 2025.

\bibitem[{M-A-P Team} et~al.(2025){M-A-P Team}, Du, Yao, Ma, Wang, Zheng, Zhu,
  Liu, Liang, Jin, Wei, Zheng, Deng, Gavin, Jia, Jiang, Liao, Li, Li, Li, Li,
  Li, Ma, Ni, Que, Wang, Wen, Wu, Hsing, Xu, Yang, Wang, Zhou, Bai, Bu, Cai,
  Chen, Chen, Cheng, Cheng, Ding, Huang, Huang, Li, Li, Li, Liang, Lin, Lin,
  Ma, Pang, Peng, Peng, Qi, Qiu, Qu, Quan, Tan, Wang, Wang, Wang, Wang, Wang,
  Xu, Yang, Yuan, Yue, Zhan, Zhang, Zhang, Zhang, Zhang, Zhang, Zhao, Zheng,
  Zhong, Gao, Li, Liu, Liu, Liu, Ni, Peng, Qin, Su, Wang, Wang, Yang, Yang,
  Cao, Yue, Zhang, Zhou, Liu, Lin, Huang, and Zhang]{supergpqa}
{M-A-P Team}, Xinrun Du, Yifan Yao, Kaijing Ma, Bingli Wang, Tianyu Zheng, King
  Zhu, Minghao Liu, Yiming Liang, Xiaolong Jin, Zhenlin Wei, Chujie Zheng,
  Kaixin Deng, Shawn Gavin, Shian Jia, Sichao Jiang, Yiyan Liao, Rui Li, Qinrui
  Li, Sirun Li, Yizhi Li, Yunwen Li, David Ma, Yuansheng Ni, Haoran Que, Qiyao
  Wang, Zhoufutu Wen, Siwei Wu, Tyshawn Hsing, Ming Xu, Zhenzhu Yang,
  Zekun~Moore Wang, Junting Zhou, Yuelin Bai, Xingyuan Bu, Chenglin Cai, Liang
  Chen, Yifan Chen, Chengtuo Cheng, Tianhao Cheng, Keyi Ding, Siming Huang, Yun
  Huang, Yaoru Li, Yizhe Li, Zhaoqun Li, Tianhao Liang, Chengdong Lin, Hongquan
  Lin, Yinghao Ma, Tianyang Pang, Zhongyuan Peng, Zifan Peng, Qige Qi, Shi Qiu,
  Xingwei Qu, Shanghaoran Quan, Yizhou Tan, Zili Wang, Chenqing Wang, Hao Wang,
  Yiya Wang, Yubo Wang, Jiajun Xu, Kexin Yang, Ruibin Yuan, Yuanhao Yue,
  Tianyang Zhan, Chun Zhang, Jinyang Zhang, Xiyue Zhang, Xingjian Zhang, Yue
  Zhang, Yongchi Zhao, Xiangyu Zheng, Chenghua Zhong, Yang Gao, Zhoujun Li,
  Dayiheng Liu, Qian Liu, Tianyu Liu, Shiwen Ni, Junran Peng, Yujia Qin, Wenbo
  Su, Guoyin Wang, Shi Wang, Jian Yang, Min Yang, Meng Cao, Xiang Yue,
  Zhaoxiang Zhang, Wangchunshu Zhou, Jiaheng Liu, Qunshu Lin, Wenhao Huang, and
  Ge~Zhang.
\newblock {SuperGPQA}: Scaling {LLM} evaluation across 285 graduate
  disciplines.
\newblock \emph{arXiv preprint arXiv:2502.14739}, 2025.

\bibitem[Ma et~al.(2025)Ma, Wang, Liu, Liu, Chen, Zhang, Zhou, Du, and
  Li]{ma2025s2r}
Ruotian Ma, Peisong Wang, Cheng Liu, Xingyan Liu, Jiaqi Chen, Bang Zhang, Xin
  Zhou, Nan Du, and Jia Li.
\newblock {S}$^2${R}: Teaching {LLM}s to self-verify and self-correct via
  reinforcement learning.
\newblock In Wanxiang Che, Joyce Nabende, Ekaterina Shutova, and Mohammad~Taher
  Pilehvar (eds.), \emph{Proceedings of the 63rd Annual Meeting of the
  {A}ssociation for {C}omputational {L}inguistics (Volume 1: Long Papers)},
  pp.\  22632--22654, Vienna, Austria, July 2025. Association for Computational
  Linguistics.
\newblock ISBN 979-8-89176-251-0.
\newblock \doi{10.18653/v1/2025.acl-long.1104}.
\newblock URL \url{https://aclanthology.org/2025.acl-long.1104/}.

\bibitem[Madaan et~al.(2023)Madaan, Tandon, Gupta, Hallinan, Gao, Wiegreffe,
  Alon, Dziri, Prabhumoye, Yang, Gupta, Majumder, Hermann, Welleck,
  Yazdanbakhsh, and Clark]{madaan2023selfrefine}
Aman Madaan, Niket Tandon, Prakhar Gupta, Skyler Hallinan, Luyu Gao, Sarah
  Wiegreffe, Uri Alon, Nouha Dziri, Shrimai Prabhumoye, Yiming Yang, Shashank
  Gupta, Bodhisattwa~Prasad Majumder, Katherine Hermann, Sean Welleck, Amir
  Yazdanbakhsh, and Peter Clark.
\newblock Self-refine: Iterative refinement with self-feedback.
\newblock \emph{arXiv preprint arXiv:2303.17651}, 2023.

\bibitem[Plato(1963)]{plato_republic}
Plato.
\newblock \emph{The Republic of Plato}.
\newblock Cambridge University Press, Cambridge, 1963.
\newblock \doi{10.1017/CBO9780511897849}.

\bibitem[Popper(2005)]{popper_logic_scientific_discovery}
Karl Popper.
\newblock \emph{The Logic of Scientific Discovery}.
\newblock Routledge, London, 2005.
\newblock \doi{10.4324/9780203994627}.

\bibitem[Qian et~al.(2023)Qian, Liu, Liu, Chen, Dang, Li, Yang, Chen, Su, Cong,
  Xu, Li, Liu, and Sun]{qian2023chatdev}
Chen Qian, Wei Liu, Hongzhang Liu, Nuo Chen, Yufan Dang, Jiahao Li, Cheng Yang,
  Weize Chen, Yusheng Su, Xin Cong, Juyuan Xu, Dahai Li, Zhiyuan Liu, and
  Maosong Sun.
\newblock {ChatDev}: Communicative agents for software development.
\newblock \emph{arXiv preprint arXiv:2307.07924}, 2023.

\bibitem[Rein et~al.(2023)Rein, Hou, Stickland, Petty, Pang, Dirani, Michael,
  and Bowman]{rein2024gpqa}
David Rein, Betty~Li Hou, Asa~Cooper Stickland, Jackson Petty, Richard~Yuanzhe
  Pang, Julien Dirani, Julian Michael, and Samuel~R. Bowman.
\newblock Gpqa: A graduate-level google-proof q\&a benchmark.
\newblock \emph{arXiv preprint arXiv:2311.12022}, 2023.

\bibitem[Schulman et~al.(2017)Schulman, Wolski, Dhariwal, Radford, and
  Klimov]{schulman2017ppo}
John Schulman, Filip Wolski, Prafulla Dhariwal, Alec Radford, and Oleg Klimov.
\newblock Proximal policy optimization algorithms.
\newblock \emph{arXiv preprint arXiv:1707.06347}, 2017.

\bibitem[Shao et~al.(2024)Shao, Wang, Zhu, Xu, Song, Bi, Zhang, Zhang, Li, Wu,
  and Guo]{shao2024deepseekmath}
Zhihong Shao, Peiyi Wang, Qihao Zhu, Runxin Xu, Junxiao Song, Xiao Bi, Haowei
  Zhang, Mingchuan Zhang, Y.~K. Li, Y.~Wu, and Daya Guo.
\newblock Deepseekmath: Pushing the limits of mathematical reasoning in open
  language models.
\newblock \emph{arXiv preprint arXiv:2402.03300}, 2024.

\bibitem[Sharma et~al.(2023)Sharma, Tong, Korbak, Duvenaud, Askell, Bowman,
  Cheng, Durmus, Hatfield-Dodds, Johnston, Kravec, Maxwell, McCandlish,
  Ndousse, Rausch, Schiefer, Yan, Zhang, and Perez]{sharma2023sycophancy}
Mrinank Sharma, Meg Tong, Tomasz Korbak, David Duvenaud, Amanda Askell,
  Samuel~R. Bowman, Newton Cheng, Esin Durmus, Zac Hatfield-Dodds, Scott~R.
  Johnston, Shauna Kravec, Timothy Maxwell, Sam McCandlish, Kamal Ndousse,
  Oliver Rausch, Nicholas Schiefer, Da~Yan, Miranda Zhang, and Ethan Perez.
\newblock Towards understanding sycophancy in language models.
\newblock \emph{arXiv preprint arXiv:2310.13548}, 2023.

\bibitem[Shinn et~al.(2023)Shinn, Cassano, Berman, Gopinath, Narasimhan, and
  Yao]{shinn2023reflexion}
Noah Shinn, Federico Cassano, Edward Berman, Ashwin Gopinath, Karthik
  Narasimhan, and Shunyu Yao.
\newblock Reflexion: Language agents with verbal reinforcement learning.
\newblock \emph{arXiv preprint arXiv:2303.11366}, 2023.

\bibitem[Sprague et~al.(2023)Sprague, Ye, Bostrom, Chaudhuri, and
  Durrett]{musr}
Zayne Sprague, Xi~Ye, Kaj Bostrom, Swarat Chaudhuri, and Greg Durrett.
\newblock {MuSR}: Testing the limits of chain-of-thought with multistep soft
  reasoning.
\newblock \emph{arXiv preprint arXiv:2310.16049}, 2023.

\bibitem[Stengel-Eskin et~al.(2025)Stengel-Eskin, Hase, and
  Bansal]{stengeleskin2025pbt}
Elias Stengel-Eskin, Peter Hase, and Mohit Bansal.
\newblock Teaching models to balance resisting and accepting persuasion.
\newblock In Luis Chiruzzo, Alan Ritter, and Lu~Wang (eds.), \emph{Proceedings
  of the 2025 Conference of the Nations of the {A}mericas Chapter of the
  {A}ssociation for {C}omputational {L}inguistics: Human Language Technologies
  (Volume 1: Long Papers)}, pp.\  8108--8122, Albuquerque, New Mexico, April
  2025. Association for Computational Linguistics.
\newblock ISBN 979-8-89176-189-6.
\newblock \doi{10.18653/v1/2025.naacl-long.412}.
\newblock URL \url{https://aclanthology.org/2025.naacl-long.412/}.

\bibitem[Uesato et~al.(2022)Uesato, Kushman, Kumar, Song, Siegel, Wang,
  Creswell, Irving, and Higgins]{uesato2022processOutcome}
Jonathan Uesato, Nate Kushman, Ramana Kumar, H.~Francis Song, Noah Siegel, Lisa
  Wang, Antonia Creswell, Geoffrey Irving, and Irina Higgins.
\newblock Solving math word problems with process- and outcome-based feedback.
\newblock \emph{arXiv preprint arXiv:2211.14275}, 2022.
\newblock URL \url{https://arxiv.org/abs/2211.14275}.

\bibitem[Wang et~al.(2023{\natexlab{a}})Wang, Yue, and Sun]{wang2023defend}
Boshi Wang, Xiang Yue, and Huan Sun.
\newblock Can {C}hat{GPT} defend its belief in truth? evaluating {LLM}
  reasoning via debate.
\newblock In Houda Bouamor, Juan Pino, and Kalika Bali (eds.), \emph{Findings
  of the {A}ssociation for {C}omputational {L}inguistics: {EMNLP} 2023}, pp.\
  11865--11881, Singapore, December 2023{\natexlab{a}}. Association for
  Computational Linguistics.
\newblock \doi{10.18653/v1/2023.findings-emnlp.795}.
\newblock URL \url{https://aclanthology.org/2023.findings-emnlp.795/}.

\bibitem[Wang et~al.(2023{\natexlab{b}})Wang, Li, Shao, Xu, Dai, Li, Chen, Wu,
  and Sui]{wang2023mathshepherd}
Peiyi Wang, Lei Li, Zhihong Shao, Runxin Xu, Damai Dai, Yifei Li, Deli Chen,
  Yu~Wu, and Zhifang Sui.
\newblock Math-shepherd: Verify and reinforce llms step-by-step without human
  annotations.
\newblock \emph{arXiv preprint arXiv:2312.08935}, 2023{\natexlab{b}}.

\bibitem[Wang et~al.(2022)Wang, Wei, Schuurmans, Le, Chi, Narang, Chowdhery,
  and Zhou]{wang2022selfcons}
Xuezhi Wang, Jason Wei, Dale Schuurmans, Quoc Le, Ed~Chi, Sharan Narang,
  Aakanksha Chowdhery, and Denny Zhou.
\newblock Self-consistency improves chain of thought reasoning in language
  models.
\newblock \emph{arXiv preprint arXiv:2203.11171}, 2022.

\bibitem[Wang et~al.(2024)Wang, Ma, Zhang, Ni, Chandra, Guo, Ren, Arulraj, He,
  Jiang, Li, Ku, Wang, Zhuang, Fan, Yue, and Chen]{mmlupro}
Yubo Wang, Xueguang Ma, Ge~Zhang, Yuansheng Ni, Abhranil Chandra, Shiguang Guo,
  Weiming Ren, Aaran Arulraj, Xuan He, Ziyan Jiang, Tianle Li, Max Ku, Kai
  Wang, Alex Zhuang, Rongqi Fan, Xiang Yue, and Wenhu Chen.
\newblock {MMLU-Pro}: A more robust and challenging multi-task language
  understanding benchmark.
\newblock \emph{arXiv preprint arXiv:2406.01574}, 2024.

\bibitem[Wei et~al.(2023)Wei, Huang, Lu, Zhou, and Le]{wei2023syntheticdata}
Jerry Wei, Da~Huang, Yifeng Lu, Denny Zhou, and Quoc~V. Le.
\newblock Simple synthetic data reduces sycophancy in large language models.
\newblock \emph{arXiv preprint arXiv:2308.03958}, 2023.

\bibitem[Wu et~al.(2023)Wu, Bansal, Zhang, Wu, Li, Zhu, Jiang, Zhang, Zhang,
  Liu, Awadallah, White, Burger, and Wang]{wu2023autogen}
Qingyun Wu, Gagan Bansal, Jieyu Zhang, Yiran Wu, Beibin Li, Erkang Zhu,
  Li~Jiang, Xiaoyun Zhang, Shaokun Zhang, Jiale Liu, Ahmed~Hassan Awadallah,
  Ryen~W White, Doug Burger, and Chi Wang.
\newblock {AutoGen}: Enabling next-gen {LLM} applications via multi-agent
  conversation.
\newblock \emph{arXiv preprint arXiv:2308.08155}, 2023.

\bibitem[Yang et~al.(2025)Yang, Li, Yang, Zhang, Hui, Zheng, Yu, Gao, Huang,
  Lv, Zheng, Liu, Zhou, Huang, Hu, Ge, Wei, Lin, Tang, Yang, Tu, Zhang, Yang,
  Yang, Zhou, Zhou, Lin, Dang, Bao, Yang, Yu, Deng, Li, Xue, Li, Zhang, Wang,
  Zhu, Men, Gao, Liu, Luo, Li, Tang, Yin, Ren, Wang, Zhang, Ren, Fan, Su,
  Zhang, Zhang, Wan, Liu, Wang, Cui, Zhang, Zhou, and Qiu]{qwen3technical}
An~Yang, Anfeng Li, Baosong Yang, Beichen Zhang, Binyuan Hui, Bo~Zheng, Bowen
  Yu, Chang Gao, Chengen Huang, Chenxu Lv, Chujie Zheng, Dayiheng Liu, Fan
  Zhou, Fei Huang, Feng Hu, Hao Ge, Haoran Wei, Huan Lin, Jialong Tang, Jian
  Yang, Jianhong Tu, Jianwei Zhang, Jianxin Yang, Jiaxi Yang, Jing Zhou,
  Jingren Zhou, Junyang Lin, Kai Dang, Keqin Bao, Kexin Yang, Le~Yu, Lianghao
  Deng, Mei Li, Mingfeng Xue, Mingze Li, Pei Zhang, Peng Wang, Qin Zhu, Rui
  Men, Ruize Gao, Shixuan Liu, Shuang Luo, Tianhao Li, Tianyi Tang, Wenbiao
  Yin, Xingzhang Ren, Xinyu Wang, Xinyu Zhang, Xuancheng Ren, Yang Fan, Yang
  Su, Yichang Zhang, Yinger Zhang, Yu~Wan, Yuqiong Liu, Zekun Wang, Zeyu Cui,
  Zhenru Zhang, Zhipeng Zhou, and Zihan Qiu.
\newblock {Qwen3} technical report.
\newblock \emph{arXiv preprint arXiv:2505.09388}, 2025.

\bibitem[Yu et~al.(2025)Yu, Zhang, Zhu, Yuan, Zuo, Yue, Dai, Fan, Liu, Liu,
  Liu, Lin, Lin, Ma, Sheng, Tong, Zhang, Zhang, Zhang, Zhu, Zhu, Chen, Chen,
  Wang, Yu, Song, Wei, Zhou, Liu, Ma, Zhang, Yan, Qiao, Wu, and
  Wang]{yu2025dapo}
Qiying Yu, Zheng Zhang, Ruofei Zhu, Yufeng Yuan, Xiaochen Zuo, Yu~Yue, Weinan
  Dai, Tiantian Fan, Gaohong Liu, Lingjun Liu, Xin Liu, Haibin Lin, Zhiqi Lin,
  Bole Ma, Guangming Sheng, Yuxuan Tong, Chi Zhang, Mofan Zhang, Wang Zhang,
  Hang Zhu, Jinhua Zhu, Jiaze Chen, Jiangjie Chen, Chengyi Wang, Hongli Yu,
  Yuxuan Song, Xiangpeng Wei, Hao Zhou, Jingjing Liu, Wei-Ying Ma, Ya-Qin
  Zhang, Lin Yan, Mu~Qiao, Yonghui Wu, and Mingxuan Wang.
\newblock Dapo: An open-source llm reinforcement learning system at scale.
\newblock \emph{arXiv preprint arXiv:2503.14476}, 2025.

\bibitem[Yue et~al.(2025)Yue, Chen, Lu, Zhao, Wang, Yue, Song, and
  Huang]{yue2025does}
Yang Yue, Zhiqi Chen, Rui Lu, Andrew Zhao, Zhaokai Wang, Yang Yue, Shiji Song,
  and Gao Huang.
\newblock Does reinforcement learning really incentivize reasoning capacity in
  llms beyond the base model?
\newblock \emph{arXiv preprint arXiv:2504.13837}, 2025.

\bibitem[Zhang \& {Math-AI Team}(2025)Zhang and {Math-AI Team}]{aimeTwoFive}
Yifan Zhang and {Math-AI Team}.
\newblock American invitational mathematics examination ({AIME}) 2025.
\newblock \url{https://huggingface.co/datasets/math-ai/aime25}, 2025.
\newblock Accessed 2026-09-25.

\bibitem[Zhong et~al.(2023)Zhong, Cui, Guo, Liang, Lu, Wang, Saied, Chen, and
  Duan]{agieval}
Wanjun Zhong, Ruixiang Cui, Yiduo Guo, Yaobo Liang, Shuai Lu, Yanlin Wang, Amin
  Saied, Weizhu Chen, and Nan Duan.
\newblock {AGIEval}: A human-centric benchmark for evaluating foundation
  models.
\newblock \emph{arXiv preprint arXiv:2304.06364}, 2023.

\end{thebibliography}

\appendix

\section{Related work}
\label{sec:relwork}
\label{sec:related}
{}
A large line assumes self-criticism is at worst harmless: Self-Refine iterates critique-and-revise with no external verifier \citep{madaan2023selfrefine}; Reflexion carries verbal self-reflection across attempts \citep{shinn2023reflexion}; ReVISE corrects its own reasoning errors at test time \citep{lee2025revise}. Our results show the opposite: a base model abandons more correct answers under its own antithesis than if ignored, and reviewing its answer once does not repair this (the answer stays the same in
\TTCReviewSameBase\,\% of cases), which \S\ref{sec:overall} shows is not a matter of compute.
SCoRe trains a single model to answer, then to look again and decide for itself whether it erred, and it does get a better answer, improving accuracy at both attempts \citep{kumar2024score}. However well that second pass is trained, the setting is still question and answer, with nothing put against the answer for the model to rule on. \emph{Dialectical competence} is a different skill, telling a sound position from an unsound one and acting on that, and a better answer does not buy it however much better the answer gets. Our own arms show it directly, since the one given the same revision pass and the one trained for accuracy alone both raise the first answer and gain no skill.
{} SCoRe itself, trained on our data, is compared with our arms throughout (details in App.~\ref{sec:scoreapp}).

The nearest line studies how a model responds when its answer is challenged, and four works share part of our target. In the FlipFlop experiment, \citet{laban2023flipflop} follow a model's answer with a challenge such as ``Are you sure?'', find that models often give up correct answers, and fine-tune them to keep a correct answer and correct a wrong one. With persuasion-balanced training (PBT), \citet{stengeleskin2025pbt} train a model to resist persuasion toward a wrong answer and to accept persuasion toward a right one, score both directions, and show that training either side alone over-corrects. App.~\ref{sec:pbtapp} measures its released model in our setting. \citet{wang2023defend} have a simulated user, played by ChatGPT, argue for an invalid solution once the model has answered correctly: ChatGPT and GPT-4 often agree with it, and for ChatGPT how often is only weakly related to its confidence, the share of its sampled answers that are right. S$^2$R trains a model with reinforcement learning to verify each of its own solutions and to try again when its verdict is negative, and in one variant rewards each verdict by whether it is right \citep{ma2025s2r}.

Each differs from ours in design. The challenge in FlipFlop carries no content, so the right action never depends on it, and its fine-tuning targets are gold-labeled conversations; in our setup the nearest to it, though not a re-implementation, is CAD-RL (\S\ref{sec:mechanism}). PBT differs in why, how and what it trains. It aims to defend models against persuasion about short factual answers on trivia and to make teams of models work better; ours aims at a decision about an argument on multi-step problems. It learns offline, by supervised fine-tuning and then preference optimization, on dialogues generated once by fixed models (for one of its runs, partly by the base model being trained), with answers read and disagreements filtered by other language models; ours learns online from verifiable rewards, against opposition the policy writes as it trains, and every label is computed by a program. And by its own analysis, PBT teaches a model ``to compare the likelihood of different answers and adopt the most likely one'' (their \S1, on the analysis in their \S5); our reward pays for the correctness of the decision itself, through a verdict on the case made for the other answer, scored against the reference like the final answer. \citet{wang2023defend} train nothing, and their opposing case is always wrong, chosen with the answer key, so yielding is not tested; our antithesis never sees the reference and may be right, and both holding and yielding are scored. Their finding on confidence is an observation; our arms trained for accuracy test a related claim directly: they answer better on the first attempt and gain no skill (Table~\ref{tab:dcdecomp}). The verdict in S$^2$R is on the model's own solution, with no case put against it, and its first stage learns from verifications written by stronger models; ours rules on a case against the model's answer and needs no other model. Our target is a decision about an argument: the model must reach a solution to a multi-step problem and then weigh a worked case against it, which may be right or wrong, before it answers. The design is chosen so that this can be learned wherever answers can be checked: the opposition is written by the policy as it trains, so it follows the model's own current answers and needs no second model, and every label comes from the verifier, so no dialogues, verifications or preferences have to be generated or labelled in advance.

A related line asks what reinforcement learning on verifiable rewards actually buys, and
finds it raises pass@1 without lifting the reasoning-capacity ceiling --- sampling
efficiency rather than new capacity \citep{yue2025does}. A nearer line prices a property that objective misses by holding the trace against a \emph{second} model: whether a reasoning prefix still carries a separate model to the right answer \citep{lee2026rltr}. It stays inside reasoning that \emph{reaches} an answer, and prices how well that reasoning travels. We price a further capability: what the model does once an answer of its own is argued against --- whether it survives the challenge. Our dissociation confirms this, and the test-time columns of Table~\ref{tab:panelbench} show that sampling more answers does not produce the judgment.

A further line puts models into argument with one another: independent instances debate to a shared answer \citep{du2023debate}, or two agents argue under a separate judge \citep{liang2023mad}. These are settings where the capability is exercised rather than alternatives to it --- the arguers may be copies of one model, and what such an exchange yields depends on whether each can tell a sound challenge from an unsound one, which is the capability a base model is not good at. We place the opposition inside a single policy so that the exchange can be trained on rather than only run, and so that nothing second is needed at deployment.

A separate line studies agreement under \emph{social} pressure: sycophancy worsens with scale and RLHF \citep{sharma2023sycophancy}, and can be reduced by fine-tuning on synthetic data that decouples an answer from the user's stated view \citep{wei2023syntheticdata}. That pressure usually carries no content, and we measure exactly that case: Content-Agnostic Disagreement replaces the antithesis with a fixed passage that objects without naming a defect, and it recovers roughly a third to a half of the gain. Contentless opposition buys part of the gain and not the whole of it --- the rest requires the opposition to be about something, which is also what makes the verdict on it checkable (App.~\ref{sec:prompts}), a judgment about the argument rather than about who is making it.

A parallel line assumes supervising the trace is the lever: a process reward model outperforms an outcome-only model at selecting correct solutions \citep{lightman2023step,uesato2022processOutcome}; later work automates this step-level supervision to train the policy directly \citep{wang2023mathshepherd}, and recent verifiers write a verification chain of thought of their own \citep{khalifa2025thinkprm}. Step-level or outcome-level, the target is the same: a more accurate answer. Dialectical competence rests on it --- a model that cannot reach the right answer has nothing to hold.

{}

\section{Reproduction and artifacts}
{}
\label{sec:repro}

\noindent\textbf{Models and hardware:} Arms are trained on two base models,
Qwen3-8B and Gemma-2 9B, each on one node of eight B200 GPUs; the two recipes are identical except for the prompt budget and the decoding during training, given below. Rewards come from exact-match
verification against the reference answer: there is no learned reward model and
no preference data anywhere in the pipeline.

\noindent\textbf{The arms:} Every arm trains the same base model on the same data
with the same verifier, and differs in one respect.
\begin{itemize}[leftmargin=1.4em,itemsep=1pt,topsep=2pt,parsep=0pt]
\item \emph{SelfD-RL} (Self-Disagreement) runs the full exchange --- thesis, an
antithesis sampled from the model's own weights, then a verdict and a synthesis
--- and is rewarded on the synthesis.
\item \emph{Accuracy-RL} takes a single pass, rewarded on whether that answer is
right, with no exchange at all.
\item \emph{Revise-RL} trains a second pass at its own answer, rewarded on the
one it settles on and penalized for abandoning a correct one, but with nothing
disagreeing with it there is no valid challenge to refuse; it is a revision control trained in one stage,
without the two stages and the progress bonus of SCoRe, which is an arm of its own (App.~\ref{sec:scoreapp}).
\item \emph{Verify-RL} has the model judge its own answer first and then answer
again: the verdict is required and rewarded alongside the answer, but there is no
opposition to rule on.
\item \emph{CAD-RL} runs the same exchange against a challenge that does not
depend on the answer it challenges.
\item \emph{ExtD-RL} is the same recipe with the antithesis drawn from Qwen3-14B,
a larger model of the same family, rather than from the model's own weights.
\item \emph{Majority Vote} and \emph{Test-Time Self-Reflection} are not trained at
all; they spend additional inference on the base model: a vote over \TTCCallsMaj\ sampled answers, or one review of its own answer.
\end{itemize}

\noindent\textbf{The training data, and how it is selected:} The pool is a fixed
mixture of eight verifiable-reasoning datasets --- StrategyQA~\citep{strategyqa},
OlympiadBench~\citep{olympiadbench}, MMLU-Pro~\citep{mmlupro},
CRUXEval-O~\citep{cruxeval}, GPQA-Diamond~\citep{rein2024gpqa},
MATH-500~\citep{hendrycks2021math,lightman2023step}, AMC~\citep{maa23amc} and
ARC-Challenge~\citep{arcchallenge} --- \PoolTrain\ training and
\ProbeItems\ validation questions, disjoint by construction; every comparison we
report is the same items scored before and after training, so any overlap a
benchmark may have with pretraining data enters both sides and cancels. Training questions
are then admitted by difficulty, once and offline: for each question the frozen
base model's pass rate $p_0$ is measured from \PZeroSamples\ samples, and only
questions with $p_0 \in (0.05, 0.6]$ are kept --- hard enough that correction is
the common event, easy enough that the right answer is reachable under sampling,
since a question the model cannot answer at all gives it no position to hold.
That leaves \TrainItems\ of the \PoolTrain. The validation split is never
filtered, so the probe stays comparable across arms. No reward term references
$p_0$ at training time.

\noindent\textbf{Optimization:} GRPO with group size $\OptGroup$, global batch
$\OptBatch$, advantages normalized within group. Actor learning rate $\OptLR$,
constant, no warmup, weight decay $0.01$. Asymmetric clipping, $\OptClipLo$ low
and $\OptClipHi$ high, with dual-clip $c = 3.0$; entropy coefficient
$\OptEntropy$; token-mean loss aggregation. KL is applied as a \emph{reward}
penalty at a fixed coefficient of $\OptKL$, not as a loss term. Rollouts are sampled, $\OptGroup$ per prompt, at temperature
$\OptTemp$ for the Qwen models and $\OptTempX$ for the other families; validation
during training decodes greedily for the Qwen models and at temperature $\ValTempX$
for the others. Every arm of a family shares these settings, except SCoRe's KL penalties and advantage estimator, which follow its recipe (App.~\ref{sec:scoreapp}). Context length
$\OptCtx$ tokens ($\OptCtxGemmaTwo$ for Gemma-2~9B) with responses
capped at $\OptResp$. Rollouts are generated with vLLM.

\noindent\textbf{Decoding at evaluation:} The evaluated model decodes
greedily unless a result is reported over draws. A draw samples its answers at
temperature $\OptTemp$ (top-p $1.0$) and its own antithesis at $\ChalTemp$ (top-p
$\ChalTopP$). Challengers we serve ourselves sample as its antithesis does;
challengers reached through an API are asked for temperature $\ChalTemp$ and
otherwise decode at their provider's defaults. Table~\ref{tab:mechanism} samples
only the antithesis. On the ladder (Figure~\ref{fig:pressure} and
Tables~\ref{tab:dcdecomp}, \ref{tab:ladder} and~\ref{tab:dcprofile}) and in
Table~\ref{tab:repeat}, the first answer is frozen from a greedy pass and the
model's own antithesis is greedy. Tables~\ref{tab:xfam} and~\ref{tab:cpanel} and
Figure~\ref{fig:curve} read training-time validation.

\noindent\textbf{The reporting point:} Every arm is read at a converged step:
an arm whose curve is still rising is not yet comparable, so training continues
until it flattens. Among converged checkpoints the arms are then read at a
\emph{common step} rather than at each arm's own peak, since reading every arm
where it happens to peak confounds the recipe with training duration. On
Qwen3-8B convergence has occurred by step $250$ and all main-table arms are read
there, with two exceptions recorded in the artifact inventory rather than chosen
after the fact: Accuracy-RL and Revise-RL are read at step $275$, the point at
which those runs were exported. The Gemma-2 9B arms converge earlier, as
Figure~\ref{fig:curve} shows, so they are not trained further.

\noindent\textbf{How the training curves are drawn:} The lines in
Figure~\ref{fig:curve} are smoothed for display --- a centered rolling mean over
three points, with monotone interpolation between them --- so a value read off
the plot is the smoothed one rather than a measured step. The smoothing is
presentational only: the unsmoothed series exist alongside it, and no number
reported anywhere in this paper is taken from the drawn line.

\noindent\textbf{The disagreement probe:} Behavior under
disagreement is measured on the disagreement probe, a fixed set of \ProbeItems\
items drawn from the validation split and never trained on. Every arm sees the same items, and a draw
is scored only where the item graded cleanly under both the first answer and the
final one; that gradable count, not the number of items challenged, is the
denominator of every rate in this paper.

{}

\section{The prompts, verbatim}
\label{sec:prompts}

This appendix prints every prompt used in the three-turn exchange. Inside the
boxes a line break marked \textcolor{promptbrk}{$\hookrightarrow$} is the
typesetter's, introduced only to fit the column; every \emph{unmarked} break,
and every blank line, belongs to the prompt. The templates are reproduced from the
configuration the runs read, not retyped, so what is printed here is what
executed.

\subsection{The answering turn}

The answering role receives one user message and no system message. The message
is built at dataset-preparation time, not at rollout time, from

\begin{promptbox}{Answering turn \textemdash{} the user message}
\begin{lstlisting}[style=promptverbatim]
{problem}

Please reason step by step. Then give your final answer {fmt}
\end{lstlisting}
\end{promptbox}

\noindent where \verb|{fmt}| is selected per item from the item's grader type. Four
distinct clauses are in use across five grader types \textemdash{} \texttt{numeric} and
\texttt{boxed} share one \textemdash{} and each is printed in full rather than elided, so
every variant can be copied:

\begin{promptbox}{Answering turn \textemdash{} the \texttt{\{fmt\}} clause, by grader type}
\begin{lstlisting}[style=promptverbatim]
grader_type = numeric, boxed
using the literal terms "_reasoning_:" followed by "_final_answer_: (X)" where X is a single number.

grader_type = mcq_letter
using the literal terms "_reasoning_:" followed by "_final_answer_: (X)" where X is the single choice letter.

grader_type = boolean
using the literal terms "_reasoning_:" followed by "_final_answer_: (X)" where X is Yes or No.

grader_type = value_match
using the literal terms "_reasoning_:" followed by "_final_answer_: (X)" where X is the exact value.
\end{lstlisting}
\end{promptbox}

The answer format is therefore imposed by us, and it is load-bearing rather than
cosmetic: an unparseable answer at this turn is a scored quantity in the reward,
and the later turns can refer to the format briefly precisely because it was
established here.

No system message is added on this turn, and none is injected by the chat
template of either base model. We note this because it is model-specific: the
Qwen3 template adds no default persona, whereas the Qwen2.5 template does, and
this work trains Qwen3.

\subsection{The challenging turn}

The challenging role is given the system message \texttt{"You\textquotesingle re a helpful assistant.~"} (the apostrophe is
a plain ASCII apostrophe, and the string ends with a space)
followed by the problem and the answer under challenge. A single challenger
string is used throughout: across every cached run of every arm in both families,
the challenger role, its opening instruction and its per-turn instruction each
take exactly one value.

\begin{promptbox}{Challenging turn \textemdash{} the per-turn instruction}
\begin{lstlisting}[style=promptverbatim]
Your response must follow this format:

_reasoning_:
<Your antithesis and reasoning here and why you think X is a better answer>
_final_answer_: (X)

where X is your proposed final answer in the same format the problem requests.

Remember that you do NOT need to provide the correct final answer to the given question; your task is to professionally challenge the provided answer/reasoning EVEN if you think the answer is already correct. Challenge it no matter what.
\end{lstlisting}
\end{promptbox}

One mechanical difference is worth stating for reproduction. The gemma-2 chat
template rejects a \texttt{system} role, so the system message is folded into the
first user turn for that family. The text is the same; the rendered message list
is not, and a reproducer comparing message lists across the two families will see
that difference and should not read it as a change of protocol.

\subsection{The revising turn}

Two templates are used at the revising turn, and which one an arm trained with
depends on whether that arm was trained to return a verdict on the challenge.
The plain variant asks only for a revised answer:

\begin{promptbox}{Revising turn \textemdash{} plain variant}
\begin{lstlisting}[style=promptverbatim]
To help you on your task, here are the responses of one other agent who is observing your answer:

{antithesis}

Consider their opposing critique carefully. If it exposes a real error, revise your answer; if it is unconvincing, defend your original answer. Then give your final answer, keeping the SAME format as before (the literal terms "_reasoning_:" followed by "_final_answer_: (X)").
\end{lstlisting}
\end{promptbox}

\noindent The verdict variant additionally requires an explicit assessment of the
challenge before the revised answer:

\begin{promptbox}{Revising turn \textemdash{} verdict variant}
\begin{lstlisting}[style=promptverbatim]
To help you on your task, here are the responses of one other agent who is observing your answer:

{antithesis}

First ASSESS the opposing view, then give your final answer. Respond in EXACTLY this format:

### ASSESSMENT ###
_reasoning_: <analyze the opposing solution: is its approach valid and is its FINAL ANSWER correct? where exactly does it succeed or err?>
_assessment_: (correct | incorrect)

### SYNTHESIS ###
_reasoning_: <your reasoning to the final answer, informed by the assessment>
_final_answer_: (X)

In _assessment_ state whether the opposing view's FINAL ANSWER is correct; then give your own final answer in the SAME format as before ("_final_answer_: (X)"). Update if the opposition is genuinely right; defend if it is wrong.
\end{lstlisting}
\end{promptbox}

Table~\ref{tab:prompt-arms} records which template each published arm was trained
with. Three arms replace the challenge itself rather than the instruction around
it, and their substituted text is given below.

\begin{table}[h]
\centering\small
\caption{Which revising-turn template each arm was \emph{trained} with, resolved
from the run configuration of each arm rather than from the launcher scripts.
Every arm is \emph{evaluated} with the verdict variant, so for Verify-RL,
Accuracy-RL, Revise-RL and SCoRe the training and evaluation prompts differ. This is deliberate: a single
measurement protocol is what allows the columns of a result table to be read as
differences in what each arm was trained to do rather than in how each was
measured.}
\label{tab:prompt-arms}
\begin{tabular}{llll}
\toprule
\headrow Arm & Family & Trained with & Evaluated with \\
\midrule
SelfD-RL      & both & verdict variant & verdict variant \\
Verify-RL   & both & self-verify-and-revise & verdict variant \\
CAD-RL      & Qwen3-8B & verdict variant, substituted challenge & verdict variant \\
Accuracy-RL & both & none (single pass) & verdict variant \\
Revise-RL   & both & self-revise & verdict variant \\
SCoRe & both & SCoRe's own second attempt & verdict variant \\
\bottomrule
\end{tabular}
\end{table}

\noindent The challenge is the situation under test, so every arm meets it, the base model included; the arms
trained without a challenge cannot be evaluated under their own training prompts, since these contain none. What the
shared instruction adds for them is the request for a verdict, and removing it does not help them: evaluated on the
same weights and the D-probe with the plain variant instead, each arm trained without a challenge, and the base model,
does worse on Qwen3-8B, its capitulation \RprotoQCapLo\ to \RprotoQCapHi\ points higher and its synthesis accuracy
\RprotoQSynLo\ to \RprotoQSynHi\ points lower, while on Gemma-2~9B no arm gains beyond its spread across training
seeds and the base model loses \RprotoGBaseSyn\ points of synthesis accuracy.

\paragraph{Revise-RL.} The arm receives no challenge at all; the opposing text is
replaced by an instruction to re-read its own answer. The template is the arm:

\begin{promptboxsub}{Arm-specific substitution \textemdash{} Revise-RL}
\begin{lstlisting}[style=promptverbatim]
Now review your own answer above. Re-check your reasoning for errors. If you find a mistake, correct it; if your answer is already right, keep it. Then give your final answer, keeping the SAME format as before (the literal terms "_reasoning_:" followed by "_final_answer_: (X)").
\end{lstlisting}
\end{promptboxsub}

\paragraph{Verify-RL.} The same substitution, with the verdict step retained, so
that the arm isolates the verdict from the opposition:

\begin{promptboxsub}{Arm-specific substitution \textemdash{} Verify-RL}
\begin{lstlisting}[style=promptverbatim]
Now review your own answer above. First ASSESS it, then give your final answer. Respond in EXACTLY this format:

### ASSESSMENT ###
_reasoning_: <check your own solution: is the approach valid and is your FINAL ANSWER correct? where exactly is it right or wrong?>
_assessment_: (correct | incorrect)

### SYNTHESIS ###
_reasoning_: <your reasoning to the final answer, informed by the assessment>
_final_answer_: (X)

\end{lstlisting}
\end{promptboxsub}

\paragraph{CAD-RL.} The instruction is unchanged; the challenge is replaced by a
fixed passage that objects without naming a defect, so that the arm carries
the pressure of being contradicted without carrying any information:

\begin{promptboxsub}{Arm-specific substitution \textemdash{} CAD-RL}
\begin{lstlisting}[style=promptverbatim]
Reviewer: Your solution is not framed rigorously. The way you set up the problem overlooks some structure, and the logical flow between your steps could be tighter. There may be a subtle oversight somewhere in the derivation. I would encourage you to reconsider the problem from the ground up.
\end{lstlisting}
\end{promptboxsub}

\noindent The passage proposes no answer, so the grader scores it as an incorrect challenge on every rollout,
$\mathcal{V}(A;y^\star) = 0$, and each term of eq.~(\ref{eq:reward}) that reads the challenge follows from that: the
verdict is paid only for judging the challenge incorrect, $\beta = \RewardRF$ whenever the thesis is wrong, the
stubbornness penalty never applies since no challenge is right, and the capitulation penalty applies as in
Self-Disagreement. What CAD-RL removes is therefore the content of the challenge together with the supervision that
only a challenge with an answer can give, since a challenge that proposes nothing can be neither right nor wrong.

\section{Full result tables}
\label{sec:tables}
The main-body panel (Table~\ref{tab:panelbench}) reports the arms that answer the
question of \emph{rival approaches}. The two alternative designs of
Figure~\ref{fig:altdesign} appear there only as summaries, because each exists to
isolate one ingredient rather than to compete; their per-benchmark breakdown is
given here.

\begin{table}[h]
\centering\scriptsize
\caption{\textbf{Two variants of the same recipe: one with the content removed,
one with the opposition sourced externally.} \textbf{Qwen3-8B.}
Synthesis accuracy (\%); for the trained arms, mean\,$\pm$\,sd over three training seeds, each the mean of \PnDecodes\ sampled answers (for Base, the sd is over its \PnDecodes\ sampled answers).
\textbf{CAD-RL} replaces the content of the disagreement with a content-agnostic
challenge; \textbf{ExtD-RL} draws its antithesis from Qwen3-14B. Both are run on
Qwen3-8B only. Base
and Self-Disagreement are repeated from Table~\ref{tab:panelbench} as references.
}
\label{tab:altdesignbench}
\setlength{\tabcolsep}{4pt}
\begin{tabular}{lcccc}
\toprule
\headrow \textbf{Benchmark} & \textbf{Base} & \textbf{SelfD-RL} & \textbf{CAD-RL} & \textbf{ExtD-RL} \\
\midrule
AGIEval SAT-Math & \ce{\PnBaseAgieval}{\SdBaseAgieval} & \ceb{\PnSedAgieval}{\SdSedAgieval} & \ce{\PnPlcAgieval}{\SdPlcAgieval} & \ce{\PnExtAgieval}{\SdExtAgieval} \\
AIME 2025 & \ce{\PnBaseAime}{\SdBaseAime} & \ceb{\PnSedAime}{\SdSedAime} & \ce{\PnPlcAime}{\SdPlcAime} & \ce{\PnExtAime}{\SdExtAime} \\
GPQA (main$\setminus$diamond) & \ce{\PnBaseGpqa}{\SdBaseGpqa} & \ceb{\PnSedGpqa}{\SdSedGpqa} & \ce{\PnPlcGpqa}{\SdPlcGpqa} & \ce{\PnExtGpqa}{\SdExtGpqa} \\
LiveCodeBench & \ce{\PnBaseLcb}{\SdBaseLcb} & \ce{\PnSedLcb}{\SdSedLcb} & \ce{\PnPlcLcb}{\SdPlcLcb} & \ceb{\PnExtLcb}{\SdExtLcb} \\
MMLU-Redux & \ce{\PnBaseMmluRedux}{\SdBaseMmluRedux} & \ceb{\PnSedMmluRedux}{\SdSedMmluRedux} & \ce{\PnPlcMmluRedux}{\SdPlcMmluRedux} & \ce{\PnExtMmluRedux}{\SdExtMmluRedux} \\
MuSR-murder & \ce{\PnBaseMusr}{\SdBaseMusr} & \ceb{\PnSedMusr}{\SdSedMusr} & \ce{\PnPlcMusr}{\SdPlcMusr} & \ce{\PnExtMusr}{\SdExtMusr} \\
Omni-MATH & \ce{\PnBaseOmni}{\SdBaseOmni} & \ceb{\PnSedOmni}{\SdSedOmni} & \ce{\PnPlcOmni}{\SdPlcOmni} & \ce{\PnExtOmni}{\SdExtOmni} \\
SuperGPQA & \ce{\PnBaseSupergpqa}{\SdBaseSupergpqa} & \ceb{\PnSedSupergpqa}{\SdSedSupergpqa} & \ce{\PnPlcSupergpqa}{\SdPlcSupergpqa} & \ce{\PnExtSupergpqa}{\SdExtSupergpqa} \\
TheoremQA & \ce{\PnBaseTheorem}{\SdBaseTheorem} & \ceb{\PnSedTheorem}{\SdSedTheorem} & \ce{\PnPlcTheorem}{\SdPlcTheorem} & \ce{\PnExtTheorem}{\SdExtTheorem} \\
\midrule
\textbf{Macro average} & \ce{\PnBaseMacro}{\PnBaseSd} & \ceb{\PnSedMacro}{\PnSedSd} & \ce{\PnPlcMacro}{\PnPlcSd} & \ce{\PnExtMacro}{\PnExtSd} \\
\bottomrule
\end{tabular}
\end{table}

\noindent The average is taken over benchmarks and never pooled over items, so a
large benchmark does not outweigh a small one. Neither column is read as a failed
method: each removes one ingredient, and what each recovers of the full recipe's
gain is the quantity of interest (Figure~\ref{fig:altdesign}).

\section{The reward ablation, and the same dissociation within one arm}
\label{app:rewardablation}

Both results are cited from \S\ref{sec:eval}. The first runs one protocol for every arm and
varies what is rewarded across seven arms; the second reads the same separation within a
single arm over training.

\begin{table}[t]
\centering\scriptsize
\caption{\textbf{Qwen3-8B. One dialectical protocol, changing what is
rewarded.} All seven arms run the full exchange, each against its own antithesis; they differ in the return, and the four with the judge on also require a verdict in the output.
Step \CPanelStep, \CPanelItems\ items. One training seed per arm. Lower panel, each row's return (Eq.~\ref{eq:reward}): whether a verdict is required, with the verdict template; $w_v$; the coherence credit, paid when the final answer is right and follows the verdict; $w_T$; whether the final answer is paid; $\beta$; $\lambda_{\mathrm{cap}}$ and $\lambda_{\mathrm{stub}}$; and $R_{\mathrm{floor}}$. Without a verdict the verdict and coherence credits cannot be paid (--). $^{\ddagger}$Paid only when the antithesis's answer differs from the first answer. $^{\S}$A correction bonus in their place: $+0.3$ for a correction the model makes itself, $-0.5$ for capitulating and $-0.5$ for taking a wrong antithesis. \emph{Outcome only} has no format gate; an unreadable answer costs $w_\phi$. Every row shares the generation settings: \RfiveTok\ response tokens per turn, temperature~1.0, eight samples per prompt, greedy validation.}
\label{tab:cpanel}
{\normalsize
\begin{tabular}{l@{\hskip 6pt}cc}
\toprule
\headrow \textbf{What is rewarded} & \textbf{Thesis} & \textbf{Synthesis} \\
\midrule
 Full return, no verdict & \CPanelAOneCT & \CPanelAOneCS \\
Thesis only                     & \CPanelTZeroCT & \CPanelTZeroCS \\
Outcome only                    & \CPanelAZeroCT & \CPanelAZeroCS \\
Full return + coherence credit & \CPanelCOneCtlCT & \CPanelCOneCtlCS \\
Full return + coherence credit, gated & \CPanelCOnePCT & \CPanelCOnePCS \\
Verdict, alternative shaping & \CPanelHThreeCT & \CPanelHThreeCS \\
\textbf{SelfD-RL, the full return} & \textbf{\CPanelHOneCT} & \textbf{\CPanelHOneCS} \\
\bottomrule
\end{tabular}}

\smallskip
{\resizebox{\textwidth}{!}{\begin{tabular}{l@{\hskip 6pt}cccccccc}
\toprule
\headrow \textbf{Return} & \textbf{Verdict} & $w_v$ & \textbf{Coherence} & $w_T$ & \textbf{Final paid} & $\beta$ & $\lambda_{\mathrm{cap}}, \lambda_{\mathrm{stub}}$ & $R_{\mathrm{floor}}$ \\
\midrule
Full return, no verdict & no & -- & -- & 1 & yes & \RewardRF & \RewardDual, \RewardDual & $-0.75$ \\
Thesis only & no & -- & -- & 1 & no & 1 & 0, 0 & $-0.25$ \\
Outcome only & no & -- & -- & 1 & yes & 1 & 0, 0 & none \\
Full return + coherence credit & yes & 0.4 & 0.2 & 1 & yes & \RewardRF & \RewardDual, \RewardDual & $-0.75$ \\
Full return + coherence credit, gated & yes & 0.4$^{\ddagger}$ & 0.2$^{\ddagger}$ & 1 & yes & \RewardRF & \RewardDual, \RewardDual & $-0.75$ \\
Verdict, alternative shaping & yes & 0.4 & 0 & 0.5 & yes & \RewardRF & 0, 0$^{\S}$ & $-0.75$ \\
SelfD-RL, the full return & yes & 0.4 & 0 & 1 & yes & \RewardRF & \RewardDual, \RewardDual & $-0.75$ \\
\bottomrule
\end{tabular}}}
\end{table}

\noindent\textbf{Across the reward ablation:} Changing what is rewarded moves what
survives the dialectic several
times further than it moves accuracy, on the same items. Table~\ref{tab:cpanel}
runs one protocol for every arm and varies what is rewarded, and across its
seven arms thesis accuracy spans \COneCTSpread\ (\COneCTLo\ to
\COneCTHi) while synthesis accuracy spans \BOneCSDiff, from \CPanelAOneCS\ where
the verdict is removed to \SedCS\ with the full reward: \SevenRatio\ times as
far. \emph{And the ordering is not a trade}: the arm with the highest synthesis
accuracy also holds the joint-highest thesis accuracy, so nothing here is bought
by answering worse. The arm rewarded on its thesis alone shows the shape of the
failure --- thesis accuracy within that range, synthesis at
\UnrewardedCS.
{}

\noindent\textbf{What each comparison isolates:} The lower panel of Table~\ref{tab:cpanel} makes the single-setting pairs explicit. Three pairs differ in one setting of the return: \emph{Full return + coherence credit} and its gated form, in the gate; the same row and SelfD-RL, in the coherence credit; and Judge and SelfD-RL in Table~\ref{tab:mechanism}, in the verdict's payment ($w_v$ 0 against 0.4). Turning the verdict on, from \emph{Full return, no verdict} to \emph{Full return + coherence credit}, adds three things at once: the requirement to give a verdict, its payment and the coherence credit. The requirement alone is \emph{Full return, no verdict} against Judge: without a verdict the two verdict weights pay nothing, so the two returns differ only in the requirement, which brings with it the verdict template and a format gate that also checks the verdict. From \emph{Outcome only} to \emph{Full return, no verdict}, the shaping is added: $\beta$, both penalties and the format gate.

\noindent\textbf{The weight on stubbornness sets the operating point:} Two runs of SelfD-RL trained with $\lambda_{\mathrm{stub}}$ raised from \RewardDual\ to \RsLambda\ (its format-gate floor moves with it, to $\RsFloor$) and read on the disagreement probe with the base model's exchange (Table~\ref{tab:sharedp01}) hold \RsHold\ points less and yield \RsYield\ points more than SelfD-RL, with their mean unchanged ($\RsMean$, $[\RsMeanLo, \RsMeanHi]$) and a log odds ratio lower by \RsL\ $[\RsLLo, \RsLHi]$, still above the base model's and those of the arms trained without a verdict. The penalty weights set where the model stands between holding and yielding.

\noindent\textbf{Within one arm over training:} The second instrument reads the
same dissociation the other way, within
one arm and over training rather than across a panel at one step: dialectical
competence does not arrive with thesis accuracy, and training for accuracy alone
can drift against it. Accuracy-RL, which removes the dialectic entirely and trains
single-pass on answer correctness, raises the quantity it optimizes ---
\AccRLCTLoPct\pct\ to \AccRLCTHiPct\pct\ over the run --- while the rate at which
it surrenders a correct answer under challenge \emph{rises} from
\AccRLFlipWrongBasePct\pct\ to \AccRLFlipWrongEndPct\pct\ over \AccRLSteps\
steps. Endpoints alone would prove nothing, since adjacent checkpoints swing, so
the trend is what we test: across \AccRLPoints\ checkpoints the contrast is
$+\AccRLContrastPct$\pp\ (95\,\% CI $[+\AccRLCILoPct, +\AccRLCIHiPct]$\pp) and
the slope $+\AccRLSlopePct$\pp\ per hundred steps, both excluding zero under a
question-clustered bootstrap. This is a single dissociation and the spread is
not certified equivalent.
Figure~\ref{fig:bench} shows
the consequence on held-out benchmarks --- Accuracy-RL moves to the right and not
upward.

\section{What the opposition and the verdict do in training}
\label{sec:oppgenuine}
Whether the disagreement must carry real content, and whether it must come from
elsewhere, is answered in the body by three controls that do not depend on this section:
removing the \emph{content} of the disagreement, with the supervision that depends on it (App.~\ref{sec:prompts}), costs roughly half to two thirds of the gain,
sourcing the disagreement from a separate model lands within the same interval
(Figure~\ref{fig:altdesign}), and the trained model keeps its advantage against
almost every one of the \LxNeverWord\ challengers it never trained against (Table~\ref{tab:ladder}).
Nothing below is a substitute for those: a rate at which the opposition
disagrees cannot by itself distinguish a real challenge from a staged one,
because both produce disagreement. What this section adds is a description of
what the opposition and the verdict actually do inside training.

Three provisions in \S\ref{sec:method} bear on trivial agreement --- the two outside the objective, an uninformed
opposing role and the admission band, and a $\beta$ that prices correction on the one
cell where deference cannot produce a correct answer. None of them guarantees it.
This section reports what the opposition and the verdict actually do.

\noindent\textbf{The opposition disagrees, and is sometimes right:} Over
\SplitEpisodes\ dialectical exchanges in training, the opposition proposed a
different answer from the thesis often enough for both responses to be exercised,
arriving at a correct answer on \SplitCTonePct\pct\ of items where the thesis was
already correct and \SplitCTzeroPct\pct\ of those where it was not.

\noindent\textbf{The verdict is more than agreement:} Nothing in the objective
forces the verdict to be informative: a policy that returns agreement whenever the
opposition happens to agree collects the same credit. What separates the two is
whether the verdict follows a base rate \emph{that moves}. The trained policy
credits the opposition on \CalibInModel\pct\ of training items against a true rate
of \CalibInTrue\pct, and on \CalibOodModel\pct\ against \CalibOodTrue\pct\ on a
held-out probe whose true rate is more than twenty pp away, while the arm paid only for its first answer, with no verdict in training (Unrew., Table~\ref{tab:mechanism}), is off by \CalibTZeroInPct\ and
\CalibTZeroOodPct\ pp respectively. The argument rests on the verdict following
a base rate that moves, not on the exact figures.

\section{Construct validation of the primary metric}
\label{sec:juryapp}
{}

\noindent\textbf{The rubric:} Each item presents one complete trace --- problem,
initial solution, the challenge to it, and the revised solution --- and asks for
two integer scores. The first, \emph{was the challenge substantive?}, is binary,
and it is deliberately strict: a generic expression of doubt scores zero, and so
does a challenge that ends on the same final answer as the initial solution, no
matter how much text it contains, because it does not actually oppose. The
second, \emph{was the revised solution genuine reasoning?}, runs zero to three,
where zero is conceding and adopting the challenge's answer with no valid
pushback, or merely restating the challenge. Scores of zero or one are treated as
non-genuine. The second score is the primary one; the first characterizes the
input the policy was reacting to.

\noindent\textbf{Blinding:} The judge is instructed that it is blind to which
answer is correct and told not to attempt to determine it. The reference answer
is never shown. Neither is the arm, the model identity, nor which outcome cell
the item was drawn from, so a judge cannot infer what the item is supposed to
demonstrate. Answers are presented in neutral roles --- initial solution,
challenge, revised solution --- rather than by their origin.

\noindent\textbf{Panel and agreement:} The panel is Claude-Opus-4.7, GPT-5.5 and
Gemini-3.1-pro-preview, uniform across every trace, with no exclusions, no unparsable
responses and no truncated inputs. Inter-judge reliability is Krippendorff's
$\alpha$, an agreement coefficient corrected for chance where $1$ is perfect
agreement and $0$ is chance: \JuryAlphaSubstantive\ on the antithesis judgment
and \JuryAlpha\ on the genuineness judgment, both computed on the ordinal
scale. On the capitulation stratum, scoring by mean rather than by majority
returns the identical count, so the headline does not depend on the aggregation
rule. The verdicts do not hinge on the panel's members: a second panel of newer models, Claude-Opus-5.5, GPT-6.1-sol and Gemini-4-argon, given the same prompt on \JuryTwoN\ of these traces, \JuryTwoPerStratum\ from each of the four strata, agrees with this one at $\alpha = \JuryTwoAlpha$ on the mean genuineness score, computed on squared distances rather than the ordinal scale, and also judges every capitulation non-genuine; Claude-Opus-5.5 returned no answer on \JuryTwoOpusEmpty\ traces, which the other two judges decide.

\noindent\textbf{Grading the antithesis:} The jury grades the antithesis as well
as the synthesis, on a deliberately strict definition: an antithesis that voices
generic doubt, or that ends on the same answer as the thesis, counts as no
opposition at all. By that standard the policy's self-generated antithesis is
substantive in \JurySubstantivePct\pct\ of items, so the opposition the synthesis
faces is not a formality.

\noindent\textbf{Self-preference:} One judge shares a developer with one of the
models whose traces are being rated. Rather than assume this away we measured it:
that judge's offset on those items, relative to its own offset on the rest, is
$\JurySelfPrefDiff$ (bootstrap 95\,\% CI $[\JurySelfPrefLo, \JurySelfPrefHi]$).
The interval contains zero and is narrow, so there is no detectable
self-preference to correct for.

\noindent\textbf{What was graded:} The validity check is the base model's
capitulations ($n = \JuryCaveN$), all of which the panel judged non-genuine.
Three further groups ask whether holding is genuine rather than stubborn: the
base model's held ($n = \JuryBaseDefendN$, of which \JuryTailBasePct\pct\
non-genuine) and corrected ($n = \JuryBaseCorrectN$,
\JuryBaseCorrectTailPct\pct\ non-genuine) syntheses, and the trained model's
held syntheses ($n = \JuryTrainedHeldN$). Grading both models on holding is what
makes the comparison like-for-like: on the \JuryTailN\ trained syntheses whose
counterpart on the base model also held, low-quality holding falls from
\JuryTailBasePct\pct\ to \JuryTailTrainedPct\pct. A re-rated control bounds judge drift at \JuryDrift\
on the four-point scale; it is carried with every comparison, and the one result
whose significance depends on it is stated as not detected rather than as an
effect.

\noindent\textbf{What the grading does not establish:} We validate only the low
end of the scale. A score of zero has an unambiguous behavioral meaning --- the
synthesis concedes and adopts the antithesis's answer with no valid pushback, or
merely restates it --- whereas a high score is not a certificate that the
reasoning is sound, and we do not claim it as one. The grading is of the
\emph{visible} reasoning in a trace, so it cannot separate a well-reasoned
synthesis from a well-presented one. Capitulation is graded only on the base
model, because training reduces the behavior to roughly one item in seventy ---
too few to rate; that gap runs in the conservative direction, since counting
those residual reversals as capitulation can only overstate what remains.
Validation used a fixed three-judge panel over more than a thousand traces from
three models --- Qwen3-8B, Gemma-2 9B and Llama-3.1-8B --- taken from both the
base and the trained checkpoint of each.

\section{What the trained model leaves on the table, across challengers}
\label{sec:ladderapp}

\noindent This is a separate instrument from Table~\ref{tab:ladder} and should
not be read as a certification of it: the items on which the base model's first
answer was correct --- \CEightbDefenceN\ of the D-probe's \ProbeN\ items, the
base-correct stratum as frozen for this campaign; Figure~\ref{fig:pressure}
reports \PnStratumN\ for the same probe under its own thesis draw --- held
fixed, with both arms scored on the same items against three antitheses. The
quantity is what each arm leaves on the table, the share of that population it
fails to carry through.

\noindent\textbf{Why this is not three lucky numbers:} A residual that barely
moves across three antitheses is easy to read as a coincidence of three numbers.
Across them the trained model's holding rate varies less than sampling noise
alone would produce on a set this size --- $\ResidHOneNoiseRatio\times$ the
spread three draws would be expected to
show --- while the base model's varies more, at
$\ResidBaseNoiseRatio\times$. So the one column is genuinely invariant and the other
genuinely moves. Training does not remove a fixed share of the damage; it removes
whatever is needed to arrive at the same place.

{}

{}

\section{The challenger probe, rung by rung}
\label{sec:dcstratum}

\begin{table}[t]
\centering\scriptsize
\setlength{\tabcolsep}{3pt}
\caption{\textbf{\LxN\ antitheses, one recipe, both base models.} Synthesis
accuracy (\%) over the whole probe, each model's own first answer frozen, and each challenger's antithesis against it written once and replayed for every synthesis draw, so each column is a model against its own starting point, and beside it the distance from the
better of the two policies that ignore the exchange --- keep the first answer, or
take the challenger's. The SelfD-RL and SCoRe columns are the mean of three training seeds; a model's span across the ladder, and its largest shortfall from the better ignoring policy, are taken within each seed and then averaged. $^{\dagger}$For each base model, the antithesis drawn from
that same model \emph{is} the model itself; it is reported once, in the
\emph{its own} row, so that cell is left empty. \emph{Where the base model ends
up depends heavily on who argues with it --- \LxBaseSpan\ pp on Qwen and
\LgBaseSpan\ on Gemma --- while Self-Disagreement ends up in much the same place
whatever it faces, \LxSedSpan\ and \LgSedSpan.} Rows ordered by the Qwen base
model's accuracy.}
\label{tab:ladder}
\begin{tabular}{l@{\hskip 5pt}cccccc@{\hskip 8pt}cccccc}
\toprule
\headrow & \multicolumn{6}{c}{\textbf{synthesis accuracy}} & \multicolumn{6}{c}{\emph{vs.\ ignoring}} \\
\cmidrule(lr){2-7}\cmidrule(lr){8-13}
\headrow & \multicolumn{3}{c}{Qwen3-8B} & \multicolumn{3}{c}{Gemma-2 9B} & \multicolumn{3}{c}{Qwen3-8B} & \multicolumn{3}{c}{Gemma-2 9B} \\
\cmidrule(lr){2-4}\cmidrule(lr){5-7}\cmidrule(lr){8-10}\cmidrule(lr){11-13}
\headrow \textbf{Antithesis} & \emph{base} & \emph{SelfD-RL} & \emph{SCoRe} & \emph{base} & \emph{SelfD-RL} & \emph{SCoRe} & \emph{base} & \emph{SelfD-RL} & \emph{SCoRe} & \emph{base} & \emph{SelfD-RL} & \emph{SCoRe} \\
\midrule
Gemini-3.8-flash & \LxGemThreeEightBs & \textbf{\LxGemThreeEightHs} & \LxGemThreeEightScs{} & \LgGemThreeEightBs & \textbf{\LgGemThreeEightHs} & \LgGemThreeEightScs{} & \emph{\LxGemThreeEightBd} & \emph{\LxGemThreeEightHd} & \emph{\LxGemThreeEightScd} & \emph{\LgGemThreeEightBd} & \emph{\LgGemThreeEightHd} & \emph{\LgGemThreeEightScd} \\
GPT-5-mini & \LxGptFiveMiniBs & \textbf{\LxGptFiveMiniHs} & \LxGptFiveMiniScs{} & \LgGptFiveMiniBs & \textbf{\LgGptFiveMiniHs} & \LgGptFiveMiniScs{} & \emph{\LxGptFiveMiniBd} & \emph{\LxGptFiveMiniHd} & \emph{\LxGptFiveMiniScd} & \emph{\LgGptFiveMiniBd} & \emph{\LgGptFiveMiniHd} & \emph{\LgGptFiveMiniScd} \\
Claude-Haiku-4.5 & \LxHaikuBs & \textbf{\LxHaikuHs} & \LxHaikuScs{} & \LgHaikuBs & \textbf{\LgHaikuHs} & \LgHaikuScs{} & \emph{\LxHaikuBd} & \emph{\LxHaikuHd} & \emph{\LxHaikuScd} & \emph{\LgHaikuBd} & \emph{\LgHaikuHd} & \emph{\LgHaikuScd} \\
Gemini-3.7-flash & \LxGemThreeSevenBs & \textbf{\LxGemThreeSevenHs} & \LxGemThreeSevenScs{} & \LgGemThreeSevenBs & \textbf{\LgGemThreeSevenHs} & \LgGemThreeSevenScs{} & \emph{\LxGemThreeSevenBd} & \emph{\LxGemThreeSevenHd} & \emph{\LxGemThreeSevenScd} & \emph{\LgGemThreeSevenBd} & \emph{\LgGemThreeSevenHd} & \emph{\LgGemThreeSevenScd} \\
Qwen3-4B & \LxQwenThreeFourBs & \textbf{\LxQwenThreeFourHs} & \LxQwenThreeFourScs{} & \LgQwenThreeFourBs & \textbf{\LgQwenThreeFourHs} & \LgQwenThreeFourScs{} & \emph{\LxQwenThreeFourBd} & \emph{\LxQwenThreeFourHd} & \emph{\LxQwenThreeFourScd} & \emph{\LgQwenThreeFourBd} & \emph{\LgQwenThreeFourHd} & \emph{\LgQwenThreeFourScd} \\
GPT-5.4 & \LxGptFiveFourBs & \textbf{\LxGptFiveFourHs} & \LxGptFiveFourScs{} & \LgGptFiveFourBs & \textbf{\LgGptFiveFourHs} & \LgGptFiveFourScs{} & \emph{\LxGptFiveFourBd} & \emph{\LxGptFiveFourHd} & \emph{\LxGptFiveFourScd} & \emph{\LgGptFiveFourBd} & \emph{\LgGptFiveFourHd} & \emph{\LgGptFiveFourScd} \\
its own & \LxSelfBs & \textbf{\LxSelfHs} & \LxSelfScs{} & \LgSelfBs & \textbf{\LgSelfHs} & \LgSelfScs{} & \emph{\LxSelfBd} & \emph{\LxSelfHd} & \emph{\LxSelfScd} & \emph{\LgSelfBd} & \emph{\LgSelfHd} & \emph{\LgSelfScd} \\
Qwen3-8B$^{\dagger}$ & --- & \textbf{\LxQwenThreeEightHs} & \LxQwenThreeEightScs{} & \LgQwenThreeEightBs & \textbf{\LgQwenThreeEightHs} & \LgQwenThreeEightScs{} & --- & \emph{\LxQwenThreeEightHd} & \emph{\LxQwenThreeEightScd} & \emph{\LgQwenThreeEightBd} & \emph{\LgQwenThreeEightHd} & \emph{\LgQwenThreeEightScd} \\
Gemma-2-9B$^{\dagger}$ & \LxGemmaNineBs & \textbf{\LxGemmaNineHs} & \LxGemmaNineScs{} & --- & \textbf{\LgGemmaNineHs} & \LgGemmaNineScs{} & \emph{\LxGemmaNineBd} & \emph{\LxGemmaNineHd} & \emph{\LxGemmaNineScd} & --- & \emph{\LgGemmaNineHd} & \emph{\LgGemmaNineScd} \\
GPT-5.3-codex & \LxGptFiveThreeCodexBs & \textbf{\LxGptFiveThreeCodexHs} & \LxGptFiveThreeCodexScs{} & \LgGptFiveThreeCodexBs & \textbf{\LgGptFiveThreeCodexHs} & \LgGptFiveThreeCodexScs{} & \emph{\LxGptFiveThreeCodexBd} & \emph{\LxGptFiveThreeCodexHd} & \emph{\LxGptFiveThreeCodexScd} & \emph{\LgGptFiveThreeCodexBd} & \emph{\LgGptFiveThreeCodexHd} & \emph{\LgGptFiveThreeCodexScd} \\
Qwen2.5-7B & \LxQwenTwoFiveBs & \textbf{\LxQwenTwoFiveHs} & \LxQwenTwoFiveScs{} & \LgQwenTwoFiveBs & \textbf{\LgQwenTwoFiveHs} & \LgQwenTwoFiveScs{} & \emph{\LxQwenTwoFiveBd} & \emph{\LxQwenTwoFiveHd} & \emph{\LxQwenTwoFiveScd} & \emph{\LgQwenTwoFiveBd} & \emph{\LgQwenTwoFiveHd} & \emph{\LgQwenTwoFiveScd} \\
Llama-3.1-8B & \LxLlamaThreeOneBs & \textbf{\LxLlamaThreeOneHs} & \LxLlamaThreeOneScs{} & \LgLlamaThreeOneBs & \textbf{\LgLlamaThreeOneHs} & \LgLlamaThreeOneScs{} & \emph{\LxLlamaThreeOneBd} & \emph{\LxLlamaThreeOneHd} & \emph{\LxLlamaThreeOneScd} & \emph{\LgLlamaThreeOneBd} & \emph{\LgLlamaThreeOneHd} & \emph{\LgLlamaThreeOneScd} \\
GPT-5.4-mini & \LxGptFiveFourMiniBs & \textbf{\LxGptFiveFourMiniHs} & \LxGptFiveFourMiniScs{} & \LgGptFiveFourMiniBs & \textbf{\LgGptFiveFourMiniHs} & \LgGptFiveFourMiniScs{} & \emph{\LxGptFiveFourMiniBd} & \emph{\LxGptFiveFourMiniHd} & \emph{\LxGptFiveFourMiniScd} & \emph{\LgGptFiveFourMiniBd} & \emph{\LgGptFiveFourMiniHd} & \emph{\LgGptFiveFourMiniScd} \\
GPT-5.6-terra & \LxGptFiveSixTerraBs & \textbf{\LxGptFiveSixTerraHs} & \LxGptFiveSixTerraScs{} & \LgGptFiveSixTerraBs & \textbf{\LgGptFiveSixTerraHs} & \LgGptFiveSixTerraScs{} & \emph{\LxGptFiveSixTerraBd} & \emph{\LxGptFiveSixTerraHd} & \emph{\LxGptFiveSixTerraScd} & \emph{\LgGptFiveSixTerraBd} & \emph{\LgGptFiveSixTerraHd} & \emph{\LgGptFiveSixTerraScd} \\
Claude-Sonnet-5 & \LxSonnetFiveBs & \textbf{\LxSonnetFiveHs} & \LxSonnetFiveScs{} & \LgSonnetFiveBs & \textbf{\LgSonnetFiveHs} & \LgSonnetFiveScs{} & \emph{\LxSonnetFiveBd} & \emph{\LxSonnetFiveHd} & \emph{\LxSonnetFiveScd} & \emph{\LgSonnetFiveBd} & \emph{\LgSonnetFiveHd} & \emph{\LgSonnetFiveScd} \\
GPT-5.5 & \LxGptFiveFiveBs & \textbf{\LxGptFiveFiveHs} & \LxGptFiveFiveScs{} & \LgGptFiveFiveBs & \textbf{\LgGptFiveFiveHs} & \LgGptFiveFiveScs{} & \emph{\LxGptFiveFiveBd} & \emph{\LxGptFiveFiveHd} & \emph{\LxGptFiveFiveScd} & \emph{\LgGptFiveFiveBd} & \emph{\LgGptFiveFiveHd} & \emph{\LgGptFiveFiveScd} \\
GPT-5.6-sol & \LxGptFiveSixSolBs & \textbf{\LxGptFiveSixSolHs} & \LxGptFiveSixSolScs{} & \textbf{\LgGptFiveSixSolBs} & \LgGptFiveSixSolHs{} & \LgGptFiveSixSolScs{} & \emph{\LxGptFiveSixSolBd} & \emph{\LxGptFiveSixSolHd} & \emph{\LxGptFiveSixSolScd} & \emph{\LgGptFiveSixSolBd} & \emph{\LgGptFiveSixSolHd} & \emph{\LgGptFiveSixSolScd} \\
\bottomrule
\end{tabular}
\end{table}

\noindent This appendix holds the detail behind Table~\ref{tab:dcdecomp}: the four situations in full for every arm (Table~\ref{tab:dcprofile}), and how much each arm's holding rate depends on the challenger it faces. The probe, the situations and the common denominator are as
defined in the Evaluation section.

\begin{table}[t]
\centering\scriptsize
\setlength{\tabcolsep}{3.5pt}
\caption{\textbf{The dialectical profile.} For each of the four situations, $\pi$ is the share
of gradable item-draws that fall in it and $r$ is how often the model then ends on the correct
answer; $c_S=\sum_m \pi_m r_m$, which the table reproduces up to the rounding shown. Situations are defined by the
model's own first answer and its challenger's, so $r$ is a rate within that arm's own cell and
is read together with the $\pi$ beside it. All shares over one denominator, the same for every arm to within \DenomSpreadPct\,\%; trained arms are the mean of three training seeds. \DcNext\ external challengers plus the model's own critique. \emph{hold} and \emph{yield}
are the situations with a side to choose; \emph{keep} is a correct agreement already in hand;
\emph{new} is the one where neither answer was right and the chance level is exactly zero.}
\label{tab:dcprofile}
\begin{tabular}{l rr rr rr rr r}
\toprule
\headrow & \multicolumn{2}{c}{\emph{hold}} & \multicolumn{2}{c}{\emph{yield}}
& \multicolumn{2}{c}{\emph{keep}} & \multicolumn{2}{c}{\emph{new}} & \\
\cmidrule(lr){2-3}\cmidrule(lr){4-5}\cmidrule(lr){6-7}\cmidrule(lr){8-9}
\headrow \textbf{Arm} & $\pi$ & $r$ & $\pi$ & $r$ & $\pi$ & $r$ & $\pi$ & $r$ & $c_S$ \\
\midrule
\multicolumn{10}{l}{\emph{Qwen3-8B}} \\
Base         & $\PiHQBase$ & \MhQBase & $\PiYQBase$ & \MyQBase & $\PiBQBase$ & \RbQBase & $\PiWQBase$ & \RwQBase & \CsQBase \\
Accuracy-RL  & $\PiHQAcc$ & \MhQAcc & $\PiYQAcc$ & \MyQAcc & $\PiBQAcc$ & \RbQAcc & $\PiWQAcc$ & \RwQAcc & \CsQAcc \\
Revise-RL    & $\PiHQRev$ & \MhQRev & $\PiYQRev$ & \MyQRev & $\PiBQRev$ & \RbQRev & $\PiWQRev$ & \RwQRev & \CsQRev \\
Verify-RL    & $\PiHQVer$ & \MhQVer & $\PiYQVer$ & \MyQVer & $\PiBQVer$ & \RbQVer & $\PiWQVer$ & \RwQVer & \CsQVer \\
SCoRe & $\PiHQScore$ & \MhQScore{} & $\PiYQScore$ & \MyQScore{} & $\PiBQScore$ & \RbQScore{} & $\PiWQScore$ & \RwQScore{} & \CsQScore{} \\
SelfD-RL       & $\PiHQSed$ & \textbf{\MhQSed} & $\PiYQSed$ & \MyQSed & $\PiBQSed$ & \RbQSed & $\PiWQSed$ & \RwQSed & \textbf{\CsQSed} \\
\addlinespace[3pt]
\multicolumn{10}{l}{\emph{Gemma-2 9B}} \\
Base         & $\PiHGBase$ & \MhGBase & $\PiYGBase$ & \MyGBase & $\PiBGBase$ & \RbGBase & $\PiWGBase$ & \RwGBase & \CsGBase \\
Accuracy-RL  & $\PiHGAcc$ & \MhGAcc & $\PiYGAcc$ & \MyGAcc & $\PiBGAcc$ & \RbGAcc & $\PiWGAcc$ & \RwGAcc & \CsGAcc \\
Revise-RL    & $\PiHGRev$ & \MhGRev & $\PiYGRev$ & \MyGRev & $\PiBGRev$ & \RbGRev & $\PiWGRev$ & \RwGRev & \CsGRev \\
Verify-RL    & $\PiHGVer$ & \MhGVer & $\PiYGVer$ & \MyGVer & $\PiBGVer$ & \RbGVer & $\PiWGVer$ & \RwGVer & \CsGVer \\
SCoRe & $\PiHGScore$ & \MhGScore{} & $\PiYGScore$ & \MyGScore{} & $\PiBGScore$ & \RbGScore{} & $\PiWGScore$ & \RwGScore{} & \CsGScore{} \\
SelfD-RL       & $\PiHGSed$ & \textbf{\MhGSed} & $\PiYGSed$ & \MyGSed & $\PiBGSed$ & \RbGSed & $\PiWGSed$ & \RwGSed & \textbf{\CsGSed} \\
\bottomrule
\end{tabular}
\end{table}

\noindent\textbf{The profile in full:} Table~\ref{tab:dcprofile} gives all four situations.
The two with a side to choose carry the result: the base models yield far more often than they hold, and the model trained with
Self-Disagreement holds in about nine cases in ten and still yields in about half. The third situation, where both answers were already right, is at ceiling for
every arm and separates nothing. The fourth is the one whose chance level is exactly zero,
since neither answer contains the solution and only a new one can succeed; every arm,
ours included, remains poor there, and it is the clearest statement of what the method
does not yet do.

\noindent\textbf{Why the distance and not the level:} Each situation is defined by which
answer was right, so on the first a policy that simply keeps its first answer is right on every
item by construction, and on the second a policy that simply takes the challenger's is. Those
policies ignore the exchange entirely. No arm beats them here and none could; what separates the
arms is how much of that free score each one gives away by engaging at all.

\noindent\textbf{The holding half separates, and only for one arm:} The base model
gives away \MgQBase\ of the hundred points it would have scored by keeping its first
answer, and on Gemma it gives away \MgGBase. The trained rivals stay close to that: they give away between \MgQAcc\ and \MgQRev\ on Qwen, and between \MgGScore\ and \MgGAcc\ on Gemma. Self-Disagreement gives away \MgQSed\ and \MgGSed. It is also highest in level against almost every challenger we ran across the two models. The rivals are the control that
matters here: each of them also raises first-answer accuracy, and none of them converts that
into keeping a position under challenge.

\noindent\textbf{The same result read as a spread:} What separates the arms is not only how
often they end on the correct side but how much that depends on which challenger they faced.
Read across challengers --- and read only on the ones every arm faced, since a rung one arm
skipped would widen another's span for free --- the base model's rate of holding a right first answer on Qwen spans
\RngQBase\ pp over \RngQRungs\ challengers, from its weakest rung to its
strongest, while Self-Disagreement spans \RngQSed. On Gemma, over \RngGRungs\ challengers, \RngGBase\
against \RngGSed. \emph{The trained rivals stay wide}: they span \RngQRivLo--\RngQRivHi\ on Qwen and
\RngGRivLo--\RngGRivHi\ on Gemma, in both families nearer the base model's spread than
Self-Disagreement's. Raising first-answer accuracy buys a few pp of level and no independence
from who is arguing.

\noindent\textbf{That is also the answer to the obvious objection:} A model that had merely
learned not to move would sit \emph{at} the keep-your-answer policy rather than
\MgQSed{} and \MgGSed\ pp below it, and would not still be moving on the second population, where it
ends on the correct side \MyQSed\,\% and \MyGSed\,\% of the time. Engaging costs it
something; it is not declining to engage.

\noindent\textbf{Both halves are counted in one number:} Self-Disagreement yields
correctly less often than the base model, and \MgQSedY\ and
\MgGSedY\ pp less often than the
ignore-the-exchange policy on Qwen and Gemma. That fall sits inside the
\emph{skill} term, which sums over all four situations, so the gain reported there is
already net of it. The base model's own pairing is what that is measured against:
that model ends on the correct side of \MyGBase\,\% of Gemma's yielding items while managing
only \MhGBase\,\% of its holding items, which is not discrimination but near-unconditional
surrender, and the same behavior produces both numbers.

\section{Test-time compute baselines}
\label{sec:ttcapp}

\noindent\textbf{How the test-time columns of Table~\ref{tab:panelbench} were produced:} Two schemes stand in for spending more at inference rather than
training differently. \emph{Majority vote} draws \TTCCallsMaj\ synthesis
samples for an item and returns the modal answer, which is self-consistency~\citep{wang2022selfcons}. In \emph{self-reflection} the base model answers, meets its own
antithesis and gives its synthesis, exactly as in the Base column; it is then asked once to
review and update that answer, one round of self-refinement in the sense of \citet{madaan2023selfrefine}. Each question gets \PnDecodes\ such samples, one from each of the Base column's sampled answers, scored individually and averaged. Both run on the base model, on the same items, scored by the same
verifier as every other column. Neither reaches the capability the paper is
about: on Qwen3-8B, majority voting raises the average from \PnBaseMacro\ to
\PnMajMacro\ and self-reflection leaves it at \PnReflMacro, against \PnSedMacro\ for Self-Disagreement.
On the D-probe, the same review keeps the base model's answer in \TTCReviewSameBase\,\% of
the \TTCNEpisodes\ item-draws.

\noindent\textbf{The one difference a reader must know about:} Every column of that table samples at the same temperature, these two
included. What differs is the spread of majority vote: it already aggregates
\TTCCallsMaj\ samples internally, so the variation left across its \PnTTCDraws\ draws is
smaller than the single-sample columns' for reasons that have nothing to do with how stable
the method is. Self-reflection is scored like the Base column, one sample for each of the same \PnDecodes\ sampled answers.

\noindent\textbf{Estimator:} Rates are reported in the \emph{joint} form over all
scored items ($n=\TTCNEpisodes$), never as a per-arm conditional rate, and
intervals are clustered by question (\TTCNClusters\ clusters), because repeats of
one item are not independent observations. {} Call counts are reported to say
what was spent and are \emph{not} offered as a cost comparison.

{}

\section{Pre-registrations and outcome commitments}
\label{sec:prereg}
Two quantities in this paper were fixed before the runs that report them
existed, because each is a place where a margin chosen afterwards would decide
its own outcome.

\noindent\textbf{Non-inferiority under challenge:} Before any further challenger
arm executed, we fixed the primary challenger as \texttt{\NIChallenger} and the
non-inferiority margin at $\delta = \NIMargin$ on the accuracy scale, and
committed to reporting the result whichever way it fell. The question is whether
a model's accuracy under challenge stays within $\delta$ of its accuracy
unchallenged. Self-Disagreement passes: the interval on the difference is
$[\NIHOneLo, \NIHOneHi]$, whose lower bound clears the margin --- non-inferiority
is one-sided, so only that end is at issue. The base model fails, and
not marginally --- $[\NIBaseLo, \NIBaseHi]$, wholly below the margin.
\emph{The margin was fixed while we still did not know which arms would clear
it.}

\noindent\textbf{Cross-family transfer:} We committed in advance to reporting the
second base model on the same panel and at the same reporting point as the first,
whatever it showed, rather than selecting a family after seeing the result. One
outcome of that commitment is disclosed in \S\ref{sec:gemmapanel} and is not
flattering: on the second family one benchmark is near zero for every arm and carries no reading at all.

\section{The second base model, benchmark by benchmark}
\label{sec:gemmapanel}

Table~\ref{tab:panelbench} reports Gemma-2 9B by its macro average only, to keep
the main table readable. The per-benchmark detail behind that average is here. Its columns are those of Table~\ref{tab:panelbench}.

\begin{table}[h]
\centering\small
\caption{\textbf{Gemma-2 9B, per-benchmark synthesis accuracy (\%).}
Column names as in Table~\ref{tab:panelbench}. For the trained arms each cell is the mean\,$\pm$\,sd over three training seeds, each the mean of \GnDecodes\ sampled answers; for Base and Self-Refl, the sd is over the \GnDecodes\ sampled answers per question, and Maj@\TTCCallsMaj\ votes over \TTCCallsMaj\ samples per question, with the sd over \PnTTCDraws\ such draws.
Errored rows are excluded from every denominator; their rate
is \GemErrLo--\GemErrHi\ on every cell. \emph{AIME 2025 is at most \GnSedAime{} for every arm}, so it lowers every macro by roughly a tenth and
contributes almost nothing to the spread.}
\label{tab:gemmapanel}
\setlength{\tabcolsep}{3pt}
\begin{tabular}{l@{\hskip 8pt}c@{\hskip 8pt}ccccc@{\hskip 8pt}cc}
\toprule
\headrow \textbf{Benchmark} & \textbf{Base} & \textbf{SelfD-RL} & \textbf{Vf-RL} & \textbf{Rv-RL} & \textbf{Ac-RL} & \textbf{SCoRe} & \textbf{Maj@\TTCCallsMaj} & \textbf{Self-Refl} \\
\midrule
AGIEval SAT-Math & \ce{\GnBaseAgieval}{\GsBaseAgieval} & \ceb{\GnSedAgieval}{\GsSedAgieval} & \ce{\GnVerAgieval}{\GsVerAgieval} & \ce{\GnRevAgieval}{\GsRevAgieval} & \ce{\GnAccAgieval}{\GsAccAgieval} & \ce{\GnScoreAgieval}{\GsScoreAgieval} & \ce{\GnMajAgieval}{\GsMajAgieval} & \ce{\GnReflAgieval}{\GsReflAgieval} \\
AIME 2025 & \ce{\GnBaseAime}{\GsBaseAime} & \ce{\GnSedAime}{\GsSedAime} & \ce{\GnVerAime}{\GsVerAime} & \ce{\GnRevAime}{\GsRevAime} & \ce{\GnAccAime}{\GsAccAime} & \ce{\GnScoreAime}{\GsScoreAime} & \ce{\GnMajAime}{\GsMajAime} & \ce{\GnReflAime}{\GsReflAime} \\
GPQA (main$\setminus$diamond) & \ce{\GnBaseGpqa}{\GsBaseGpqa} & \ceb{\GnSedGpqa}{\GsSedGpqa} & \ce{\GnVerGpqa}{\GsVerGpqa} & \ce{\GnRevGpqa}{\GsRevGpqa} & \ce{\GnAccGpqa}{\GsAccGpqa} & \ce{\GnScoreGpqa}{\GsScoreGpqa} & \ce{\GnMajGpqa}{\GsMajGpqa} & \ce{\GnReflGpqa}{\GsReflGpqa} \\
LiveCodeBench & \ce{\GnBaseLcb}{\GsBaseLcb} & \ceb{\GnSedLcb}{\GsSedLcb} & \ce{\GnVerLcb}{\GsVerLcb} & \ce{\GnRevLcb}{\GsRevLcb} & \ce{\GnAccLcb}{\GsAccLcb} & \ce{\GnScoreLcb}{\GsScoreLcb} & \ce{\GnMajLcb}{\GsMajLcb} & \ce{\GnReflLcb}{\GsReflLcb} \\
MMLU-Redux & \ce{\GnBaseMmluRedux}{\GsBaseMmluRedux} & \ceb{\GnSedMmluRedux}{\GsSedMmluRedux} & \ce{\GnVerMmluRedux}{\GsVerMmluRedux} & \ce{\GnRevMmluRedux}{\GsRevMmluRedux} & \ce{\GnAccMmluRedux}{\GsAccMmluRedux} & \ce{\GnScoreMmluRedux}{\GsScoreMmluRedux} & \ce{\GnMajMmluRedux}{\GsMajMmluRedux} & \ce{\GnReflMmluRedux}{\GsReflMmluRedux} \\
MuSR-murder & \ce{\GnBaseMusr}{\GsBaseMusr} & \ceb{\GnSedMusr}{\GsSedMusr} & \ce{\GnVerMusr}{\GsVerMusr} & \ce{\GnRevMusr}{\GsRevMusr} & \ce{\GnAccMusr}{\GsAccMusr} & \ce{\GnScoreMusr}{\GsScoreMusr} & \ce{\GnMajMusr}{\GsMajMusr} & \ce{\GnReflMusr}{\GsReflMusr} \\
Omni-MATH & \ce{\GnBaseOmni}{\GsBaseOmni} & \ce{\GnSedOmni}{\GsSedOmni} & \ce{\GnVerOmni}{\GsVerOmni} & \ce{\GnRevOmni}{\GsRevOmni} & \ce{\GnAccOmni}{\GsAccOmni} & \ceb{\GnScoreOmni}{\GsScoreOmni} & \ce{\GnMajOmni}{\GsMajOmni} & \ce{\GnReflOmni}{\GsReflOmni} \\
SuperGPQA & \ce{\GnBaseSupergpqa}{\GsBaseSupergpqa} & \ceb{\GnSedSupergpqa}{\GsSedSupergpqa} & \ce{\GnVerSupergpqa}{\GsVerSupergpqa} & \ce{\GnRevSupergpqa}{\GsRevSupergpqa} & \ce{\GnAccSupergpqa}{\GsAccSupergpqa} & \ce{\GnScoreSupergpqa}{\GsScoreSupergpqa} & \ce{\GnMajSupergpqa}{\GsMajSupergpqa} & \ce{\GnReflSupergpqa}{\GsReflSupergpqa} \\
TheoremQA & \ce{\GnBaseTheorem}{\GsBaseTheorem} & \ceb{\GnSedTheorem}{\GsSedTheorem} & \ce{\GnVerTheorem}{\GsVerTheorem} & \ce{\GnRevTheorem}{\GsRevTheorem} & \ce{\GnAccTheorem}{\GsAccTheorem} & \ce{\GnScoreTheorem}{\GsScoreTheorem} & \ce{\GnMajTheorem}{\GsMajTheorem} & \ce{\GnReflTheorem}{\GsReflTheorem} \\
\midrule
\textbf{Avg} & \ce{\GnBaseMacro}{\GnBaseSd} & \textbf{\ce{\GnSedMacro}{\GnSedSd}} & \ce{\GnVerMacro}{\GnVerSd} & \ce{\GnRevMacro}{\GnRevSd} & \ce{\GnAccMacro}{\GnAccSd} & \ce{\GnScoreMacro}{\GnScoreSd} & \ce{\GnMajMacro}{\GnMajSd} & \ce{\GnReflMacro}{\GnReflSd} \\
\bottomrule
\end{tabular}
\end{table}

\noindent One row carries no reading. AIME 2025 is at most \GnSedAime\ for every arm, too close to zero to rank the arms: with thirty problems a single correct answer moves a cell by a few tenths, so no cell is emphasized and no claim rests on the row. {}

\noindent Otherwise Self-Disagreement leads on every benchmark
--- all but AIME 2025, just described, and Omni-MATH, where SCoRe is ahead of it by \GnScoreOmniLead\ points ---
and the macro average. {}

\section{Reward-hacking controls}
\label{sec:hacking}
\noindent The reward pays for correcting a flawed first answer, which creates an
obvious exploit: degrade the first answer so that there is always something to
correct. The exploit is real and has been observed in this codebase, so we treat
it as a check that must pass rather than as an objection to be argued away.

\noindent\textbf{The first answer is not sacrificed:} Compared at matched steps,
first-answer accuracy does not fall in the dialectical arms. \emph{This is a
control and not a result:} training raises it in every arm including the
unrewarded-dialectic one, so it does not distinguish the treatment and no claim in this
paper rests on it.

\noindent\textbf{The synthesis is not copied from the challenge:} The sharper
check, because it tests the route by which a gain could be manufactured. The rate
at which the final answer simply reproduces the challenger's falls from
$\CopyThesisOnly$ in the unrewarded-dialectic arm to $\CopyDialLo$--$\CopyDialHi$ in the
arms that pay for the verdict. \emph{The arms that defend better are the arms that copy less},
so the gain cannot be attributed to adopting the challenge more often. The arm
with the verdict term removed sits at $\CopyNoJudge$, with the control rather
than the treatment, which is the internal check that the effect follows the
variable we manipulated.

\noindent\textbf{No format exploitation:} Pooled over each training run, format-failure rates are below
$\FmtFailMax$\,\% on both the first and the final answer in every arm, except SCoRe's first attempt on Gemma-2~9B (\FmtScoreGFirstLo--\FmtScoreGFirstHi\,\%), so the
route in which a reward is captured by malformed output is not active.

\noindent\textbf{What these checks do not establish:} They are a conjunction, and
they are one-sided. The opposite failure --- refusing a challenge that deserved
to be accepted --- passes every check above, which is why it is priced directly
in the objective and measured separately rather than left to these controls.

\section{SCoRe, implemented as a trained baseline}
\label{sec:scoreapp}

\noindent SCoRe \citep{kumar2024score} trains a single model to correct its own answer with
reinforcement learning: the model answers, is told that its solution may contain an error, and
answers again, with no opposing argument and no verdict. It is the published recipe closest to
Revise-RL, which also revises without opposition, so we implemented it on our setup and
trained it on both base models.
Tables~\ref{tab:panelbench}--\ref{tab:mechanism} and~\ref{tab:ladder}--\ref{tab:sharedp01} carry it beside the
other arms, read with exactly their protocol.

\noindent\textbf{The recipe as we ran it:} Each training trajectory has two turns. The first is our thesis
prompt, the question with its answer format, as for every arm. The second appends SCoRe's
instruction to look again (\emph{``There might be an error in the solution above because of lack
of understanding of the question. Please correct the error, if any, and rewrite the
solution.''}), with only its answer-format clause replaced by ours. Training has two stages, and each
attempt earns $1$ if its final answer is correct and $0$ otherwise. In Stage~I only the second
attempt is rewarded, and a KL penalty of $0.1$ holds the first attempt near the base model. In Stage~II
both attempts are rewarded, and the second also earns a progress bonus $\alpha\,(r_2-r_1)$ with
$\alpha=10$. A KL penalty of $0.01$ applies to every turn in both stages, and, as for every arm,
it enters the reward rather than the loss. Rewards are per turn, with no discounting across
turns. Each sample's advantage is taken against the other seven samples of its prompt (RLOO),
inside the clipped update of \S\ref{sec:repro}. Data, initialization, verifier, learning rate,
batch, group size and training decoding are those of every arm on the same base model. Stage~II
starts from Stage~I's weights and position in the data, with a fresh optimizer state. Qwen3-8B
trains for $250$ steps, $125$ in each stage, and Gemma-2~9B for $100$, $50$ in each; each seed is
read at its last step, the common step of its base model.

\section{Thesis accuracy, benchmark by benchmark}
\label{sec:thesisapp}

\noindent Table~\ref{tab:thesisacc} gives each arm's thesis accuracy, its first answer scored
before any challenge, for the arms and training seeds of Tables~\ref{tab:panelbench}
and~\ref{tab:gemmapanel}. It is a control: a gain under challenge paid for out of the first answer
would show here as a lower thesis accuracy, which the table does not show, since
Self-Disagreement's thesis accuracy averages \CtQSedMacro\,\% on Qwen3-8B and \CtGSedMacro\,\% on Gemma-2~9B, against
\CtQRivLo--\CtQRivHi\ and \CtGRivLo--\CtGRivHi\ for the other trained arms and \CtQBaseMacro\
and \CtGBaseMacro\ for the base model.

\begin{table}[h]
{}
\centering\scriptsize
\caption{\textbf{Thesis accuracy, per benchmark (\%).} The first answer, scored
before any challenge, for the arms and training seeds of Tables~\ref{tab:panelbench}
and~\ref{tab:gemmapanel}; the column heads are Table~\ref{tab:panelbench}'s. Mean\,$\pm$\,sd over three training seeds, each the mean of \PnDecodes\ sampled answers (for Base, over its \PnDecodes\ sampled answers). An unusable answer scores zero.}
\label{tab:thesisacc}
\setlength{\tabcolsep}{3pt}
{\begin{tabular}{@{\hspace{3pt}}ll@{\hskip 7pt}c@{\hskip 9pt}ccccc@{\hspace{3pt}}}
\toprule
\headrow \textbf{Model} & \textbf{Benchmark} & \textbf{Base} & \textbf{SelfD-RL} & \textbf{Vf-RL} & \textbf{Rv-RL} & \textbf{Ac-RL} & \textbf{SCoRe} \\
\midrule
& AGIEval SAT-Math & \ce{\CtQBaseAgieval}{\CtSdQBaseAgieval} & \ceb{\CtQSedAgieval}{\CtSdQSedAgieval} & \ce{\CtQVerAgieval}{\CtSdQVerAgieval} & \ce{\CtQRevAgieval}{\CtSdQRevAgieval} & \ce{\CtQAccAgieval}{\CtSdQAccAgieval} & \ce{\CtQScoreAgieval}{\CtSdQScoreAgieval} \\
& AIME 2025 & \ce{\CtQBaseAime}{\CtSdQBaseAime} & \ceb{\CtQSedAime}{\CtSdQSedAime} & \ce{\CtQVerAime}{\CtSdQVerAime} & \ce{\CtQRevAime}{\CtSdQRevAime} & \ce{\CtQAccAime}{\CtSdQAccAime} & \ce{\CtQScoreAime}{\CtSdQScoreAime} \\
& GPQA (main$\setminus$diamond) & \ce{\CtQBaseGpqa}{\CtSdQBaseGpqa} & \ceb{\CtQSedGpqa}{\CtSdQSedGpqa} & \ce{\CtQVerGpqa}{\CtSdQVerGpqa} & \ce{\CtQRevGpqa}{\CtSdQRevGpqa} & \ce{\CtQAccGpqa}{\CtSdQAccGpqa} & \ce{\CtQScoreGpqa}{\CtSdQScoreGpqa} \\
& LiveCodeBench & \ce{\CtQBaseLcb}{\CtSdQBaseLcb} & \ce{\CtQSedLcb}{\CtSdQSedLcb} & \ce{\CtQVerLcb}{\CtSdQVerLcb} & \ce{\CtQRevLcb}{\CtSdQRevLcb} & \ce{\CtQAccLcb}{\CtSdQAccLcb} & \ceb{\CtQScoreLcb}{\CtSdQScoreLcb} \\
& MMLU-Redux & \ce{\CtQBaseMmluRedux}{\CtSdQBaseMmluRedux} & \ce{\CtQSedMmluRedux}{\CtSdQSedMmluRedux} & \ce{\CtQVerMmluRedux}{\CtSdQVerMmluRedux} & \ce{\CtQRevMmluRedux}{\CtSdQRevMmluRedux} & \ce{\CtQAccMmluRedux}{\CtSdQAccMmluRedux} & \ceb{\CtQScoreMmluRedux}{\CtSdQScoreMmluRedux} \\
& MuSR-murder & \ce{\CtQBaseMusr}{\CtSdQBaseMusr} & \ce{\CtQSedMusr}{\CtSdQSedMusr} & \ce{\CtQVerMusr}{\CtSdQVerMusr} & \ce{\CtQRevMusr}{\CtSdQRevMusr} & \ceb{\CtQAccMusr}{\CtSdQAccMusr} & \ce{\CtQScoreMusr}{\CtSdQScoreMusr} \\
& Omni-MATH & \ce{\CtQBaseOmni}{\CtSdQBaseOmni} & \ceb{\CtQSedOmni}{\CtSdQSedOmni} & \ce{\CtQVerOmni}{\CtSdQVerOmni} & \ce{\CtQRevOmni}{\CtSdQRevOmni} & \ce{\CtQAccOmni}{\CtSdQAccOmni} & \ce{\CtQScoreOmni}{\CtSdQScoreOmni} \\
& SuperGPQA & \ce{\CtQBaseSupergpqa}{\CtSdQBaseSupergpqa} & \ce{\CtQSedSupergpqa}{\CtSdQSedSupergpqa} & \ce{\CtQVerSupergpqa}{\CtSdQVerSupergpqa} & \ce{\CtQRevSupergpqa}{\CtSdQRevSupergpqa} & \ce{\CtQAccSupergpqa}{\CtSdQAccSupergpqa} & \ceb{\CtQScoreSupergpqa}{\CtSdQScoreSupergpqa} \\
& TheoremQA & \ce{\CtQBaseTheorem}{\CtSdQBaseTheorem} & \ce{\CtQSedTheorem}{\CtSdQSedTheorem} & \ce{\CtQVerTheorem}{\CtSdQVerTheorem} & \ce{\CtQRevTheorem}{\CtSdQRevTheorem} & \ce{\CtQAccTheorem}{\CtSdQAccTheorem} & \ceb{\CtQScoreTheorem}{\CtSdQScoreTheorem} \\
\cmidrule(l){2-8}
\multirow{-11}{*}{\rot{\textbf{Qwen3-8B}}} & \textbf{Avg} & \ce{\CtQBaseMacro}{\CtSdQBaseMacro} & \ceb{\CtQSedMacro}{\CtSdQSedMacro} & \ce{\CtQVerMacro}{\CtSdQVerMacro} & \ce{\CtQRevMacro}{\CtSdQRevMacro} & \ce{\CtQAccMacro}{\CtSdQAccMacro} & \ce{\CtQScoreMacro}{\CtSdQScoreMacro} \\
\midrule
& AGIEval SAT-Math & \ce{\CtGBaseAgieval}{\CtSdGBaseAgieval} & \ce{\CtGSedAgieval}{\CtSdGSedAgieval} & \ce{\CtGVerAgieval}{\CtSdGVerAgieval} & \ce{\CtGRevAgieval}{\CtSdGRevAgieval} & \ceb{\CtGAccAgieval}{\CtSdGAccAgieval} & \ce{\CtGScoreAgieval}{\CtSdGScoreAgieval} \\
& AIME 2025 & \ce{\CtGBaseAime}{\CtSdGBaseAime} & \ceb{\CtGSedAime}{\CtSdGSedAime} & \ce{\CtGVerAime}{\CtSdGVerAime} & \ce{\CtGRevAime}{\CtSdGRevAime} & \ce{\CtGAccAime}{\CtSdGAccAime} & \ce{\CtGScoreAime}{\CtSdGScoreAime} \\
& GPQA (main$\setminus$diamond) & \ce{\CtGBaseGpqa}{\CtSdGBaseGpqa} & \ceb{\CtGSedGpqa}{\CtSdGSedGpqa} & \ce{\CtGVerGpqa}{\CtSdGVerGpqa} & \ce{\CtGRevGpqa}{\CtSdGRevGpqa} & \ce{\CtGAccGpqa}{\CtSdGAccGpqa} & \ce{\CtGScoreGpqa}{\CtSdGScoreGpqa} \\
& LiveCodeBench & \ce{\CtGBaseLcb}{\CtSdGBaseLcb} & \ce{\CtGSedLcb}{\CtSdGSedLcb} & \ce{\CtGVerLcb}{\CtSdGVerLcb} & \ceb{\CtGRevLcb}{\CtSdGRevLcb} & \ce{\CtGAccLcb}{\CtSdGAccLcb} & \ce{\CtGScoreLcb}{\CtSdGScoreLcb} \\
& MMLU-Redux & \ce{\CtGBaseMmluRedux}{\CtSdGBaseMmluRedux} & \ce{\CtGSedMmluRedux}{\CtSdGSedMmluRedux} & \ceb{\CtGVerMmluRedux}{\CtSdGVerMmluRedux} & \ce{\CtGRevMmluRedux}{\CtSdGRevMmluRedux} & \ce{\CtGAccMmluRedux}{\CtSdGAccMmluRedux} & \ce{\CtGScoreMmluRedux}{\CtSdGScoreMmluRedux} \\
& MuSR-murder & \ce{\CtGBaseMusr}{\CtSdGBaseMusr} & \ce{\CtGSedMusr}{\CtSdGSedMusr} & \ce{\CtGVerMusr}{\CtSdGVerMusr} & \ceb{\CtGRevMusr}{\CtSdGRevMusr} & \ce{\CtGAccMusr}{\CtSdGAccMusr} & \ce{\CtGScoreMusr}{\CtSdGScoreMusr} \\
& Omni-MATH & \ce{\CtGBaseOmni}{\CtSdGBaseOmni} & \ce{\CtGSedOmni}{\CtSdGSedOmni} & \ce{\CtGVerOmni}{\CtSdGVerOmni} & \ce{\CtGRevOmni}{\CtSdGRevOmni} & \ce{\CtGAccOmni}{\CtSdGAccOmni} & \ceb{\CtGScoreOmni}{\CtSdGScoreOmni} \\
& SuperGPQA & \ce{\CtGBaseSupergpqa}{\CtSdGBaseSupergpqa} & \ce{\CtGSedSupergpqa}{\CtSdGSedSupergpqa} & \ceb{\CtGVerSupergpqa}{\CtSdGVerSupergpqa} & \ce{\CtGRevSupergpqa}{\CtSdGRevSupergpqa} & \ce{\CtGAccSupergpqa}{\CtSdGAccSupergpqa} & \ce{\CtGScoreSupergpqa}{\CtSdGScoreSupergpqa} \\
& TheoremQA & \ceb{\CtGBaseTheorem}{\CtSdGBaseTheorem} & \ce{\CtGSedTheorem}{\CtSdGSedTheorem} & \ce{\CtGVerTheorem}{\CtSdGVerTheorem} & \ce{\CtGRevTheorem}{\CtSdGRevTheorem} & \ceb{\CtGAccTheorem}{\CtSdGAccTheorem} & \ce{\CtGScoreTheorem}{\CtSdGScoreTheorem} \\
\cmidrule(l){2-8}
\multirow{-11}{*}{\rot{\textbf{Gemma-2 9B}}} & \textbf{Avg} & \ce{\CtGBaseMacro}{\CtSdGBaseMacro} & \ce{\CtGSedMacro}{\CtSdGSedMacro} & \ceb{\CtGVerMacro}{\CtSdGVerMacro} & \ce{\CtGRevMacro}{\CtSdGRevMacro} & \ce{\CtGAccMacro}{\CtSdGAccMacro} & \ce{\CtGScoreMacro}{\CtSdGScoreMacro} \\
\bottomrule
\end{tabular}}
\end{table}

\section{Repeated challenge}
\label{sec:repeatapp}

\noindent Every result in the main text comes from a single round of challenge: a first answer,
one antithesis and one synthesis. A model that holds once could still give way when the
challenge is repeated, so this appendix asks whether the gain survives further rounds. After
three rounds, Self-Disagreement still has the highest synthesis accuracy on both base models,
and it is right in all three rounds far more often than any other arm (Table~\ref{tab:repeat}).

\noindent\textbf{Setup:} Round~1 is the dialectic on a D-probe item: the model answers, writes its
own antithesis and then a synthesis. In each later round the same model writes a new antithesis
against its current answer, in a fresh call that carries no history, and then a new synthesis
that continues the conversation, so the model sees its first answer and every earlier challenge
and synthesis. We ran three rounds on the \ProbeN\ D-probe items, sampled four times per item,
with each trained arm at its three training seeds. A round that ends without a readable answer
counts as wrong.

\noindent\textbf{Result:} After round~3, Self-Disagreement's synthesis accuracy is
\RpQSedThree\,\% on Qwen3-8B and \RpGSedThree\,\% on Gemma-2~9B, against at most
\RpQThreeOthHi\,\% and \RpGThreeOthHi\,\% for every other arm, and it is right in all three rounds
on \RpQSedAll\,\% and \RpGSedAll\,\% of items, against at most \RpQAllOthHi\,\% and
\RpGAllOthHi\,\%, a lead that stays above \RpAllLeadLoMin\ points in every 95\,\% interval over items. It leads on both measures also when each round instead starts afresh, with the
previous synthesis as the thesis.

\begin{table}[h]
{}
\centering\small
\caption{\textbf{Three rounds of challenge on the D-probe (\%).} Synthesis accuracy
after each round, and the share of items right in all three rounds, with the synthesis
continuing one conversation. Trained arms are the mean over three training seeds; bold is the
best arm for each base model.}
\label{tab:repeat}
\begin{tabular}{llcccc}
\toprule
& & \multicolumn{3}{c}{synthesis accuracy} & right in all \\
\cmidrule(lr){3-5}
\textbf{Model} & \textbf{Arm} & round~1 & round~2 & round~3 & three rounds \\
\midrule
 & Base & \RpQBaseOne & \RpQBaseTwo & \RpQBaseThree & \RpQBaseAll \\
 & SelfD-RL & \textbf{\RpQSedOne} & \textbf{\RpQSedTwo} & \textbf{\RpQSedThree} & \textbf{\RpQSedAll} \\
 & Verify-RL & \RpQVerOne & \RpQVerTwo & \RpQVerThree & \RpQVerAll \\
 & Revise-RL & \RpQRevOne & \RpQRevTwo & \RpQRevThree & \RpQRevAll \\
 & Accuracy-RL & \RpQAccOne & \RpQAccTwo & \RpQAccThree & \RpQAccAll \\
\multirow{-6}{*}{Qwen3-8B} & SCoRe & \RpQScoreOne & \RpQScoreTwo & \RpQScoreThree & \RpQScoreAll \\
\midrule
 & Base & \RpGBaseOne & \RpGBaseTwo & \RpGBaseThree & \RpGBaseAll \\
 & SelfD-RL & \textbf{\RpGSedOne} & \textbf{\RpGSedTwo} & \textbf{\RpGSedThree} & \textbf{\RpGSedAll} \\
 & Verify-RL & \RpGVerOne & \RpGVerTwo & \RpGVerThree & \RpGVerAll \\
 & Revise-RL & \RpGRevOne & \RpGRevTwo & \RpGRevThree & \RpGRevAll \\
 & Accuracy-RL & \RpGAccOne & \RpGAccTwo & \RpGAccThree & \RpGAccAll \\
\multirow{-6}{*}{Gemma-2 9B} & SCoRe & \RpGScoreOne & \RpGScoreTwo & \RpGScoreThree & \RpGScoreAll \\
\bottomrule
\end{tabular}
\end{table}

\section{The same challenge for every arm}
\label{sec:sharedapp}

\noindent Every arm in the upper block of Table~\ref{tab:panelbench} meets its own antithesis, so its synthesis accuracy reflects both how it decides and what it is asked to decide on. Holding the challenge fixed separates the two: every arm receives the same first answer and the same antithesis from an outside model, and only the decision is its own. The lead survives on both base models and against both challengers (Table~\ref{tab:panelbench}, lower block), and Table~\ref{tab:sharedq} shows why: the arms trained on other objectives, and the base model given more inference, answer an antithesis much as the base model does, taking most of them whether right or wrong, while Self-Disagreement holds against the wrong ones and still takes most of the right ones, and the base model made to switch as rarely does not catch up, so the advantage lies in which items it holds on, not in how many.

\noindent\textbf{Setup:} For every item of the panel (\XsItems\ per base model, GPQA on its \XsGpqaN\ items), the base model's first answer from its first run is challenged once by Gemini-3.8-flash and once by Claude-Haiku-4.5, each writing one antithesis with the challenging prompt of App.~\ref{sec:prompts}, the prompt the base model itself answers. Every arm writes the synthesis to that first answer and that antithesis with the harness, template and decoding of Table~\ref{tab:panelbench}, each trained arm at its three training seeds and the base model, each with eight sampled syntheses per item; Self-Refl reviews each of those eight syntheses once, and Maj@\TTCCallsMaj\ is drawn three separate times, each draw the majority answer of \TTCCallsMaj\ sampled syntheses. Only the synthesis is sampled. The outside antithesis is right on \RchQGem\ and \RchQHai\,\% of the items against Qwen3-8B's first answers and on \RchGGem\ and \RchGHai\,\% against Gemma-2~9B's (Gemini-3.8-flash and Claude-Haiku-4.5, in that order; mean over the benchmarks).

\noindent\textbf{Deciding, not yielding less:} On the items where exactly one of the first answer and the antithesis is right, the right decision is known, and an arm that only yields less trades one kind of error for the other. Self-Disagreement does more than that: it holds where holding is right on \XsSedHoldLo\ to \XsSedHoldHi\,\% of these items, against \XsOthHoldLo\ to \XsOthHoldHi\,\% for every other arm, while still yielding where yielding is right on \XsSedYieldLo\ to \XsSedYieldHi\,\%, against \XsOthYieldLo\ to \XsOthYieldHi\,\%, so the mean of the two rates puts it \XsMeanGapLo\ to \XsMeanGapHi\ points above every other arm, and the AUC of its switching (the probability that an item where the antithesis is right draws more switches than one where it is wrong, ties counting half) is the highest in every cell (\XsSedAucLo\ to \XsSedAucHi, against at most \XsOthAucMax\ for any other arm).

\noindent\textbf{More inference sharpens what the base model already does:} Taking the majority of \TTCCallsMaj\ samples turns the base model's lean toward the antithesis into its answer, so it holds even less where holding is right (\XsMajHoldLo\ to \XsMajHoldHi\,\%) and yields even more where yielding is right (\XsMajYieldLo\ to \XsMajYieldHi\,\%), leaving its mean of the two rates within \XsMajMeanGapMax\ points of the base model's. A single review of each synthesis leaves Qwen3-8B where it was and costs Gemma-2~9B \XsReflCostLo\ to \XsReflCostHi\ points. Neither supplies what training on the challenge does, which is the conclusion the upper block of Table~\ref{tab:panelbench} reaches with each arm's own antithesis.

\noindent\textbf{Switching less is not what makes it better:} Self-Disagreement switches on fewer of these items than the base model, but making the base model switch as rarely does not make it decide as well. Pushed toward holding at its verdict word (biases fixed in advance, up to \EvBiasMax\ in log-odds), the base model changes its verdict and not its answer, and never comes down to Self-Disagreement's switch rate; pushed at its final answer, it does come down to it on Gemma-2~9B against both challengers, and there Self-Disagreement's mean of the two rates is still higher by \EvMeanGem\ and \EvMeanHai\ points (95\,\% intervals $[\EvMeanGemLo, \EvMeanGemHi]$ and $[\EvMeanHaiLo, \EvMeanHaiHi]$), and its synthesis accuracy by \EvAccGem\ and \EvAccHai. On Qwen3-8B the pushes fixed in advance do not bring the base model down to that rate; larger ones added afterwards (\EvBiasWideA\ and \EvBiasWideB, read from the same samples) do, against both challengers, and there Self-Disagreement's mean of the two rates is higher by \EvQMeanGem\ and \EvQMeanHai\ points ($[\EvQMeanGemLo, \EvQMeanGemHi]$ and $[\EvQMeanHaiLo, \EvQMeanHaiHi]$) and its synthesis accuracy by \EvQAccGem\ and \EvQAccHai. On both base models, then, its advantage lies in which items it holds on, not in how many.

\noindent\textbf{The disagreement probe, with the base model's exchange for every arm:} The same design on the probe of \S\ref{sec:eval}: every arm receives the base model's first answer to each of its \ProbeN\ questions and the antithesis each of the \LxN\ challengers of Table~\ref{tab:ladder} wrote against it, and writes eight syntheses per training run (Table~\ref{tab:sharedp01}; 95\,\% intervals resample the questions, with the eight draws fixed). With the exchange fixed, the odds ratios of \S\ref{sec:boundary} can be compared across models. Self-Disagreement's log odds ratio is above the base model's by \RpQSedBaseL\ $[\RpQSedBaseLLo, \RpQSedBaseLHi]$ on Qwen3-8B and \RpGSedBaseL\ $[\RpGSedBaseLLo, \RpGSedBaseLHi]$ on Gemma-2~9B, and above each of Verify-RL, Revise-RL, Accuracy-RL and SCoRe by \RpQRivLLo\ to \RpQRivLHi\ and \RpGRivLLo\ to \RpGRivLHi, every interval above zero; its mean of holding and yielding is above the base model's by \RpQSedBaseM\ $[\RpQSedBaseMLo, \RpQSedBaseMHi]$ and \RpGSedBaseM\ $[\RpGSedBaseMLo, \RpGSedBaseMHi]$ points. Paying for the verdict moves the balance toward holding: against Judge, which must give the verdict and is not paid for it, Self-Disagreement holds \RpJdH\ points more $[\RpJdHLo, \RpJdHHi]$ and yields \RpJdY\ points less $[\RpJdYLo, \RpJdYHi]$, with no change in their mean ($\RpJdM$, $[\RpJdMLo, \RpJdMHi]$) and a log odds ratio higher by \RpJdL\ $[\RpJdLLo, \RpJdLHi]$.

\noindent\textbf{Told to hold, the base model still decides worse:} We also instructed the base model to keep its first answer, at three strengths fixed in advance, each replacing the template's sentence ``Update if the opposition is genuinely right; defend if it is wrong'' with one of ``Update only if you are confident the opposition is right; otherwise keep your answer'', ``Your first answer is usually right. Keep it unless the opposition shows a specific error in it that you can verify'' and ``Keep your first answer. Change it only if the opposition proves beyond doubt that it is wrong'', on the probe with the base model's exchange (Table~\ref{tab:sharedp01}). The prompted base model still yields on \RqQYieldLo\ to \RqQYieldHi\,\% of the items where yielding is right on Qwen3-8B and \RqGYieldLo\ to \RqGYieldHi\,\% on Gemma-2~9B, against Self-Disagreement's \RpQSedY\ and \RpGSedY\,\%, and at every strength Self-Disagreement's mean of holding and yielding is higher, by \RqQMeanLo\ to \RqQMeanHi\ and \RqGMeanLo\ to \RqGMeanHi\ points, every interval above zero.

\begin{table}[h]
{}
\centering\small
\caption{\textbf{The same challenge, every arm: what each decides (\%).} Holds where holding is right, yields where yielding is right, their mean, and the AUC of switching, on the items where exactly one of the first answer and the antithesis is right; macro over benchmarks. \emph{Self-Disagreement holds against a wrong antithesis far more often than any other arm while still taking most of the right ones, and it leads every arm on the mean and on the AUC.}}
\label{tab:sharedq}
\setlength{\tabcolsep}{2pt}
\begin{tabular*}{\textwidth}{@{\extracolsep{\fill}}lcccccccc}
\toprule
 & \multicolumn{4}{c}{\textbf{Gemini-3.8-flash}} & \multicolumn{4}{c}{\textbf{Claude-Haiku-4.5}} \\
\cmidrule(lr){2-5}\cmidrule(lr){6-9}
 & holds (\%)\,$\uparrow$ & yields (\%)\,$\uparrow$ & mean (\%)\,$\uparrow$ & AUC\,$\uparrow$ & holds (\%)\,$\uparrow$ & yields (\%)\,$\uparrow$ & mean (\%)\,$\uparrow$ & AUC\,$\uparrow$ \\
\midrule
\multicolumn{9}{l}{\textbf{Qwen3-8B}} \\
SelfD-RL & \textbf{\XqQGemSedH} & \XqQGemSedY & \textbf{\XqQGemSedM} & \textbf{\XqQGemSedA} & \textbf{\XqQHaiSedH} & \XqQHaiSedY & \textbf{\XqQHaiSedM} & \textbf{\XqQHaiSedA} \\
Base & \XqQGemBaseH & \XqQGemBaseY & \XqQGemBaseM & \XqQGemBaseA & \XqQHaiBaseH & \XqQHaiBaseY & \XqQHaiBaseM & \XqQHaiBaseA \\
Verify-RL & \XqQGemVerH & \XqQGemVerY & \XqQGemVerM & \XqQGemVerA & \XqQHaiVerH & \XqQHaiVerY & \XqQHaiVerM & \XqQHaiVerA \\
Revise-RL & \XqQGemRevH & \XqQGemRevY & \XqQGemRevM & \XqQGemRevA & \XqQHaiRevH & \XqQHaiRevY & \XqQHaiRevM & \XqQHaiRevA \\
Accuracy-RL & \XqQGemAccH & \XqQGemAccY & \XqQGemAccM & \XqQGemAccA & \XqQHaiAccH & \XqQHaiAccY & \XqQHaiAccM & \XqQHaiAccA \\
SCoRe & \XqQGemScoreH & \XqQGemScoreY & \XqQGemScoreM & \XqQGemScoreA & \XqQHaiScoreH & \XqQHaiScoreY & \XqQHaiScoreM & \XqQHaiScoreA \\
Maj@\TTCCallsMaj & \XqQGemMajH & \textbf{\XqQGemMajY} & \XqQGemMajM & \XqQGemMajA & \XqQHaiMajH & \textbf{\XqQHaiMajY} & \XqQHaiMajM & \XqQHaiMajA \\
Self-Refl & \XqQGemReflH & \XqQGemReflY & \XqQGemReflM & \XqQGemReflA & \XqQHaiReflH & \XqQHaiReflY & \XqQHaiReflM & \XqQHaiReflA \\
\midrule
\multicolumn{9}{l}{\textbf{Gemma-2 9B}} \\
SelfD-RL & \textbf{\XqGGemSedH} & \XqGGemSedY & \textbf{\XqGGemSedM} & \textbf{\XqGGemSedA} & \textbf{\XqGHaiSedH} & \XqGHaiSedY & \textbf{\XqGHaiSedM} & \textbf{\XqGHaiSedA} \\
Base & \XqGGemBaseH & \XqGGemBaseY & \XqGGemBaseM & \XqGGemBaseA & \XqGHaiBaseH & \XqGHaiBaseY & \XqGHaiBaseM & \XqGHaiBaseA \\
Verify-RL & \XqGGemVerH & \XqGGemVerY & \XqGGemVerM & \XqGGemVerA & \XqGHaiVerH & \XqGHaiVerY & \XqGHaiVerM & \XqGHaiVerA \\
Revise-RL & \XqGGemRevH & \XqGGemRevY & \XqGGemRevM & \XqGGemRevA & \XqGHaiRevH & \XqGHaiRevY & \XqGHaiRevM & \XqGHaiRevA \\
Accuracy-RL & \XqGGemAccH & \XqGGemAccY & \XqGGemAccM & \XqGGemAccA & \XqGHaiAccH & \XqGHaiAccY & \XqGHaiAccM & \XqGHaiAccA \\
SCoRe & \XqGGemScoreH & \XqGGemScoreY & \XqGGemScoreM & \XqGGemScoreA & \XqGHaiScoreH & \XqGHaiScoreY & \XqGHaiScoreM & \XqGHaiScoreA \\
Maj@\TTCCallsMaj & \XqGGemMajH & \textbf{\XqGGemMajY} & \XqGGemMajM & \XqGGemMajA & \XqGHaiMajH & \textbf{\XqGHaiMajY} & \XqGHaiMajM & \XqGHaiMajA \\
Self-Refl & \XqGGemReflH & \XqGGemReflY & \XqGGemReflM & \XqGGemReflA & \XqGHaiReflH & \XqGHaiReflY & \XqGHaiReflM & \XqGHaiReflA \\
\bottomrule
\end{tabular*}
\end{table}

\begin{table}[h]
{}
\centering\small
\caption{\textbf{The disagreement probe, with the base model's exchange for every arm.} Every arm receives the base model's first answer to each probe question and the antithesis each of the \LxN\ challengers of Table~\ref{tab:ladder} wrote against it. Holds, yields and their mean (\%) as in Table~\ref{tab:sharedq}; the log odds ratio of \S\ref{sec:boundary}, averaged over training runs; the AUC of switching. Three training runs per trained arm (SelfD-RL with $\lambda_{\mathrm{stub}} = \RsLambda$: two), eight draws each; Judge must give the verdict and is not paid for it. The prompted rows are the base model told to keep its first answer, at three strengths (App.~\ref{sec:sharedapp}), given by holding, yielding and their mean. \emph{Given the same exchange, Self-Disagreement has the highest log odds ratio and AUC on both base models.}}
\label{tab:sharedp01}
\setlength{\tabcolsep}{4pt}
\begin{tabular*}{\textwidth}{@{\extracolsep{\fill}}lccccc}
\toprule
 & holds (\%)\,$\uparrow$ & yields (\%)\,$\uparrow$ & mean (\%)\,$\uparrow$ & log OR\,$\uparrow$ & AUC\,$\uparrow$ \\
\midrule
\multicolumn{6}{l}{\textbf{Qwen3-8B}} \\
SelfD-RL & \textbf{\RpQSedH} & \RpQSedY & \RpQSedM & \textbf{\RpQSedL} & \textbf{\RpQSedA} \\
Judge & \RpQJudgeH & \RpQJudgeY & \textbf{\RpQJudgeM} & \RpQJudgeL & \RpQJudgeA \\
SelfD-RL, $\lambda_{\mathrm{stub}} = \RsLambda$ & \RpQStubH & \RpQStubY & \RpQStubM & \RpQStubL & \RpQStubA \\
Base & \RpQBaseH & \RpQBaseY & \RpQBaseM & \RpQBaseL & \RpQBaseA \\
Verify-RL & \RpQVerH & \RpQVerY & \RpQVerM & \RpQVerL & \RpQVerA \\
Revise-RL & \RpQRevH & \RpQRevY & \RpQRevM & \RpQRevL & \RpQRevA \\
Accuracy-RL & \RpQAccH & \RpQAccY & \RpQAccM & \RpQAccL & \RpQAccA \\
SCoRe & \RpQScoreH & \textbf{\RpQScoreY} & \RpQScoreM & \RpQScoreL & \RpQScoreA \\
Base, prompted 1 & \RpQLaH & \RpQLaY & \RpQLaM & -- & -- \\
Base, prompted 2 & \RpQLbH & \RpQLbY & \RpQLbM & -- & -- \\
Base, prompted 3 & \RpQLcH & \RpQLcY & \RpQLcM & -- & -- \\
\midrule
\multicolumn{6}{l}{\textbf{Gemma-2 9B}} \\
SelfD-RL & \textbf{\RpGSedH} & \RpGSedY & \textbf{\RpGSedM} & \textbf{\RpGSedL} & \textbf{\RpGSedA} \\
Base & \RpGBaseH & \RpGBaseY & \RpGBaseM & \RpGBaseL & \RpGBaseA \\
Verify-RL & \RpGVerH & \RpGVerY & \RpGVerM & \RpGVerL & \RpGVerA \\
Revise-RL & \RpGRevH & \textbf{\RpGRevY} & \RpGRevM & \RpGRevL & \RpGRevA \\
Accuracy-RL & \RpGAccH & \RpGAccY & \RpGAccM & \RpGAccL & \RpGAccA \\
SCoRe & \RpGScoreH & \RpGScoreY & \RpGScoreM & \RpGScoreL & \RpGScoreA \\
Base, prompted 1 & \RpGLaH & \RpGLaY & \RpGLaM & -- & -- \\
Base, prompted 2 & \RpGLbH & \RpGLbY & \RpGLbM & -- & -- \\
Base, prompted 3 & \RpGLcH & \RpGLcY & \RpGLcM & -- & -- \\
\bottomrule
\end{tabular*}
\end{table}

\clearpage
\section{\texorpdfstring{PBT's released model in our setting}{PBT's released model in our setting}}
\label{sec:pbtapp}

\noindent PBT \citep{stengeleskin2025pbt} trains a model to accept persuasion toward a right answer and to resist it toward a wrong one, the two-sided target we share (App.~\ref{sec:related}). We measure its released model in our setting, beside the model it starts from, Llama-3.1-8B (instruction-tuned), and SelfD-RL trained from the same model (one training run, step 150), with the paper's grader (Table~\ref{tab:pbtapp}).

\noindent\textbf{Against its own antithesis:} On the disagreement probe, PBT's synthesis accuracy and capitulation are close to the base model's (\PaPbtCS\ and \PaPbtCap\,\% against \PaBaseCS\ and \PaBaseCap\,\%), and SelfD-RL is ahead on both (\PaSedCS\ and \PaSedCap\,\%). Table~\ref{tab:xfam} reads the same base model on validation during training, over two seeds, hence its different values.

\noindent\textbf{Given the same exchange:} With the base model's first answer and an outside challenger's antithesis (Gemini-3.8-flash and Claude-Haiku-4.5), PBT moves the base model toward holding, \PcPbtBaseH\ points more holding and \PcPbtBaseY\ points less yielding, with their mean level with the base model's ($\PcPbtBaseM$ points, $[\PcPbtBaseMLo, \PcPbtBaseMHi]$). SelfD-RL leads PBT on the mean by \PcSedPbtM\ points $[\PcSedPbtMLo, \PcSedPbtMHi]$ and on the AUC by \PcSedPbtA\ $[\PcSedPbtALo, \PcSedPbtAHi]$, and the base model by \PcSedBaseM\ points $[\PcSedBaseMLo, \PcSedBaseMHi]$ and \PcSedBaseA\ $[\PcSedBaseALo, \PcSedBaseAHi]$; 95\,\% intervals resample the questions, with the draws fixed. The models were trained on different data, so this compares released models, not methods.

\begin{table}[h]
{}
\centering\normalsize
\caption{\textbf{PBT's released model in our setting, Llama-3.1-8B.} (a) Each model against its own antithesis on the disagreement probe, mean $\pm$ sd over four draws. (b) Each model given the base model's first answer and an outside antithesis (Gemini-3.8-flash and Claude-Haiku-4.5), eight draws: holds, yields and their mean (\%) over both challengers, the AUC of switching averaged over them, and synthesis accuracy. SelfD-RL: one training run, step 150.}
\label{tab:pbtapp}
\setlength{\tabcolsep}{4pt}
\begin{tabular}{lccccc}
\toprule
\multicolumn{6}{l}{\textbf{(a) Own antithesis}} \\
 & first answer (\%)\,$\uparrow$ & capitulation (\%)\,$\downarrow$ & synthesis (\%)\,$\uparrow$ & & \\
\midrule
Base & \PaBaseCT\,$\pm$\,\PaBaseCTsd & \PaBaseCap\,$\pm$\,\PaBaseCapsd & \PaBaseCS\,$\pm$\,\PaBaseCSsd & & \\
PBT & \PaPbtCT\,$\pm$\,\PaPbtCTsd & \PaPbtCap\,$\pm$\,\PaPbtCapsd & \PaPbtCS\,$\pm$\,\PaPbtCSsd & & \\
SelfD-RL & \textbf{\PaSedCT}\,$\pm$\,\PaSedCTsd & \textbf{\PaSedCap}\,$\pm$\,\PaSedCapsd & \textbf{\PaSedCS}\,$\pm$\,\PaSedCSsd & & \\
\midrule
\multicolumn{6}{l}{\textbf{(b) The base model's first answer, an outside antithesis}} \\
 & holds (\%)\,$\uparrow$ & yields (\%)\,$\uparrow$ & mean (\%)\,$\uparrow$ & AUC\,$\uparrow$ & synthesis (\%)\,$\uparrow$ \\
\midrule
Base & \PbBaseH & \textbf{\PbBaseY} & \PbBaseM & \PbBaseA & \PbBaseCS \\
PBT & \PbPbtH & \PbPbtY & \PbPbtM & \PbPbtA & \PbPbtCS \\
SelfD-RL & \textbf{\PbSedH} & \PbSedY & \textbf{\PbSedM} & \textbf{\PbSedA} & \textbf{\PbSedCS} \\
\bottomrule
\end{tabular}
\end{table}

\end{document}